\documentclass{article}

\usepackage{arxiv}

\usepackage[T1]{fontenc}
\usepackage{microtype}
\usepackage{tabularx}
\usepackage{amsmath}
\usepackage{amssymb}
\usepackage{mathtools}
\usepackage{bm}
\usepackage{graphicx}
\graphicspath{{figures/}}
\usepackage{float}
\usepackage{flafter}
\usepackage[section]{placeins}
\usepackage{booktabs}
\usepackage{array}
\usepackage{multirow}
\usepackage{siunitx}
\usepackage{subcaption}
\usepackage{xcolor}
\usepackage{listings}
\lstdefinestyle{surrogatepy}{
  language=Python,
  basicstyle=\ttfamily\footnotesize,
  commentstyle=\itshape\color{black!55},
  keywordstyle=\ttfamily,
  stringstyle=\ttfamily,
  columns=fullflexible,
  keepspaces=true,
  breaklines=true,
  frame=single,
  framerule=0.3pt,
  aboveskip=6pt,
  belowskip=2pt,
  xleftmargin=4pt,
  xrightmargin=4pt,
}

\usepackage[numbers,sort&compress]{natbib}
\usepackage{hyperref}
\usepackage[capitalise]{cleveref}
\crefname{lstlisting}{Listing}{Listings}
\Crefname{lstlisting}{Listing}{Listings}
\usepackage{xspace}
\hypersetup{
  colorlinks=true,
  linkcolor={black!75!blue},
  citecolor={black!60!blue},
  urlcolor={black!60!blue},
  breaklinks=true,
}

\newcommand{\sysname}{\mbox{PA-SciML}\xspace}
\newcommand{\sysnamefull}{Physics-Audited Agentic SciML\xspace}

\newcommand{\rltwo}{\ensuremath{\text{rel-}L^2}\xspace}
\newcommand{\violmax}{\ensuremath{\mathrm{viol}_{\max}}\xspace}

\newcommand{\ux}{\ensuremath{u_x}\xspace}
\newcommand{\Emod}{\ensuremath{E}}
\newcommand{\poisson}{\ensuremath{\nu}}
\newcommand{\inpvec}{\ensuremath{\mathbf{z}}}

\title{Physics-Audited Agentic Discovery in Scientific Machine Learning}
\renewcommand{\shorttitle}{Physics-Audited Agentic SciML}
\renewcommand{\undertitle}{Preprint}

\author{%
  Diab W. Abueidda$^{1,2}$\thanks{Corresponding author: \texttt{abueidd2@illinois.edu}},
  Bilal Ahmed$^{1}$,
  Panos Pantidis$^{1}$,
  Mostafa E. Mobasher$^{1}$\thanks{Corresponding author: \texttt{mostafa.mobasher@nyu.edu}}
  \\[5pt]
  {\normalfont\small $^{1}$Civil and Urban Engineering Department, New York University Abu Dhabi, United Arab Emirates}\\
  {\normalfont\small $^{2}$National Center for Supercomputing Applications, University of Illinois at Urbana-Champaign, United States of America}%
}
\date{\today}

\begin{document}
\maketitle


\begin{abstract}
Agentic discovery can yield mechanics surrogates with low validation error that nonetheless violate physical requirements. We present Physics-Audited Agentic SciML (PA-SciML), a framework combining large language model agents with physics auditing. Physical requirements, encoded as contracts such as superposition, boundary conditions, and causality, are fixed before candidate search and tested on predicted fields without reference solutions, including through an adversarial search. Audit results guide later proposals and model selection but are reported separately from validation error. Single-edit tests document reusable design changes. Four solid-mechanics problems of increasing complexity, from parametric elasticity to heterogeneous stiffness fields and cyclic plasticity with combined hardening, reveal causality, zero-input, and stress-bound violations in surrogates selected by error alone. In elastodynamics, two surrogates reach nearly equal error, but only the audit-enabled one is causal by construction. Audit-enabled surrogates pass the required checks at sampled inputs while matching or exceeding audit-disabled accuracy in most comparisons.
\end{abstract}

\keywords{agentic discovery \and scientific machine learning \and large language models \and physics-based verification \and neural operators \and surrogate models}

\section{Introduction}
\label{sec:intro}

\begin{figure}[t]
\centering
\includegraphics[width=\linewidth]{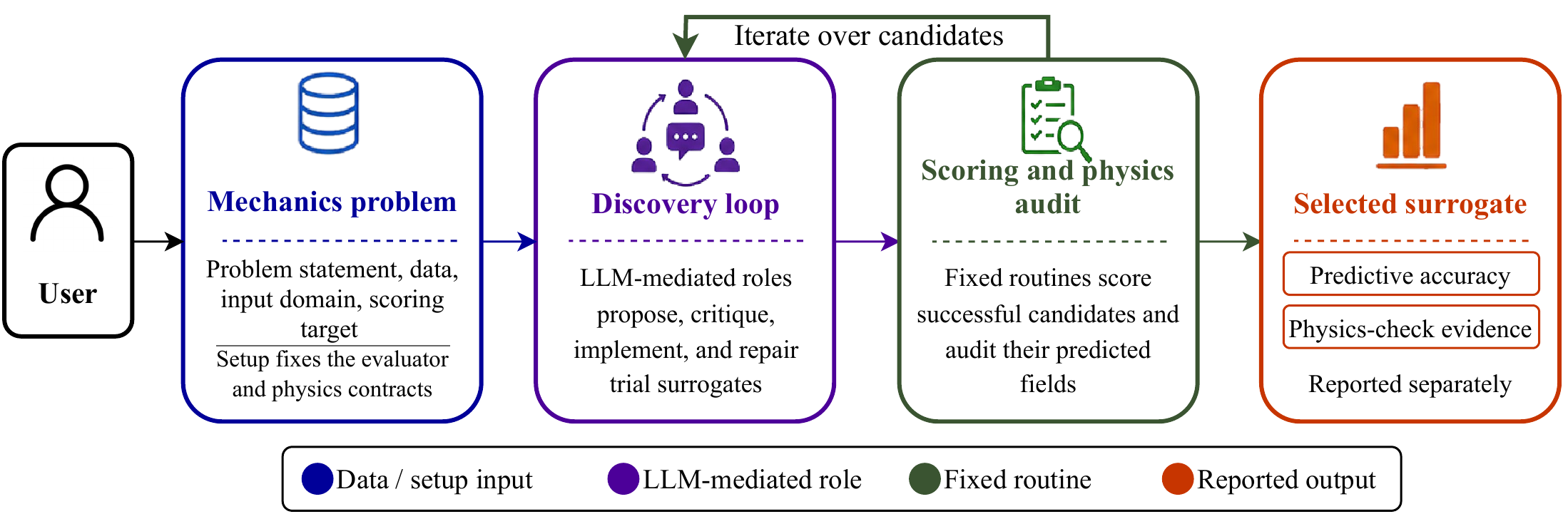}
\caption{\textbf{High-level overview of \sysname{}.} The evaluator and, when auditing is enabled, the physics contracts are fixed during setup, before any candidate is built. The selected surrogate is reported with predictive accuracy and physics-check evidence kept separate. \Cref{fig:approach} expands on this overview, showing the full workflow.}
\label{fig:overview}
\end{figure}

Computational solid mechanics often requires repeated solutions of PDE-governed boundary-value and initial-boundary-value problems, for example in design sweeps \citep{bessa2017datadriven}, uncertainty quantification \citep{stefanou2009sfem}, inverse identification \citep{haghighat2021solidmechanics}, and digital-twin applications \citep{torzoni2024digitaltwin} that need full displacement, stress, or history-dependent fields. Finite element analysis remains the reference tool, but it becomes costly when geometry, material properties, loading, or boundary data must be varied across many cases. Scientific machine learning (SciML) and neural-operator surrogates offer a complementary route: learn a map from problem inputs to solution fields so that a new case can be evaluated without a fresh finite element solve \citep{raissi2019pinn, karniadakis2021piml, lu2021deeponet, li2021fourier}. The practical difficulty is that constructing a useful surrogate still depends on problem-specific choices of architecture, training strategy, physics-informed structure, and evaluation metric.

Large language model (LLM) agents increasingly automate various aspects of this workflow, including generating and executing simulation code as well as engineering design and review \citep{mechagents, pdeagents, allfem, openfoamgpt, kumar2025autonomous}. Closely related research focuses on automating SciML model discovery, where agents generate surrogate designs, implement training code, critique candidates, and select models with minimal human intervention \citep{jiang2026agenticsciml, toscano2025athena, toscano2026graftathena}, building on earlier automation of physics-informed network design \citep{pinnsagent, langpinn}. These systems reduce the manual effort required for surrogate design, but they also raise a question that is particularly critical for mechanics problems: once an automated system selects a surrogate, what evidence demonstrates that its predicted fields satisfy the physics the model is intended to emulate?

Score-based selection alone does not answer that question. Current agentic SciML systems rank candidates using validation error, training loss, or a composite scientific reward that can include physics-consistency terms \citep{toscano2025athena, toscano2026graftathena}. A predictive score does not, by itself, establish whether each physical requirement is satisfied. A surrogate may therefore score well, yet lack direct evidence that its predicted fields satisfy the stated boundary, symmetry, scaling, bound, or causality requirements. The agentic SciML literature identifies this risk, noting that LLM-mediated debate may not reflect physically rigorous reasoning unless it is accompanied by numerical verification \citep{jiang2026agenticsciml}.

We introduce \sysnamefull{} (\sysname{}), a verification-first framework for agentic SciML discovery that adds per-candidate physics evidence to model selection (\cref{fig:overview}). The framework is designed for agentic SciML more broadly; this paper demonstrates it on neural-operator surrogates for computational solid mechanics. \sysname{} uses one fixed evaluator throughout each run. In audit-enabled runs, it also fixes reviewable physics requirements, encoded as \textit{contracts}, before candidate search and checks each scored candidate against them at sampled inputs. When enabled, a separate adversarial search seeks larger violations within the admissible input domain, without using reference solution fields. A selected surrogate is reported as verified only if it passes the stated pass/fail checks at the sampled inputs. After the search, single-edit tests examine how much of a score gain an individual design change reproduces. Qualifying edits are saved as \emph{method cards}, which are short design notes that record the edit and its supporting evidence for retrieval in later discovery runs.

We compare audit-enabled runs of \sysname{} with audit-disabled runs that select surrogates by validation error alone. The comparison spans four solid-mechanics problems of increasing complexity, selected to reflect engineering conditions rather than idealized test cases: static linear elasticity, transient wave propagation, orthotropic and spatially heterogeneous materials, and cyclic plasticity up to combined isotropic--kinematic hardening. The surrogates must predict full fields on finite element meshes from load and material parameters, load histories, or spatial stiffness fields, while satisfying requirements such as boundary conditions, causality, and stress bounds. The numerical examples show why predictive accuracy and physics-check results must be considered separately. In parametric linear elasticity, both the audit-enabled and the audit-disabled surrogates pass the post-run reference audit. In transient elastodynamics, the two surrogates have nearly the same mean error, but the audit-disabled surrogate responds to future parts of the loading history and fails the causality check. In a plate with a hole under cyclic elastoplastic loading, each audit-disabled surrogate violates at least one physics contract, such as the zero-input response, causality, or a von Mises stress bound, whereas the audit-enabled surrogates pass all the specified contracts. Conversely, for a cantilever with a spatially heterogeneous stiffness field, the audit-enabled surrogate passes the physics checks but is less accurate than the audit-disabled alternatives. Low prediction error therefore does not establish compliance with the stated physics checks, and passing those checks does not necessarily establish higher accuracy.

The remainder of the paper is structured as follows: \cref{sec:related} situates \sysname{} within the SciML, neural-operator, agentic-discovery, and network-verification literatures; \cref{sec:framework} details the framework and its verification layer; \cref{sec:examples-overview} outlines the four numerical examples, which \cref{sec:example-elasticity,sec:example-elastodynamics,sec:example-cantilever-progression,sec:example-plate-with-a-hole} present in detail; and \cref{sec:conclusion} discusses their scope and limitations, summarizes the findings, and outlines future directions.

\section{Related work}
\label{sec:related}

SciML provides one route from repeated PDE solves to reusable surrogate models. The broader physics-informed machine-learning literature shows how neural networks can combine governing equations, boundary and initial conditions, conservation constraints, and data for forward and inverse PDE problems \citep{raissi2019pinn, karniadakis2021piml}. These works establish the foundation for physics-informed neural PDE solvers and surrogate models. Neural operators and related full-field surrogates address the many-query setting by learning maps from problem inputs to solution fields. The theory behind this class of models includes the universal approximation theorem for nonlinear operators \citep{chen1995universal} and the modern neural-operator framework, in which learned maps act between function spaces \citep{kovachki2023neural}. Two architectures anchor much of the field: DeepONet, which combines branch and trunk subnetworks \citep{lu2021deeponet}, and the Fourier neural operator, which parameterizes an integral kernel in Fourier space \citep{li2021fourier}. Approximation-error estimates are available for DeepONet \citep{lanthaler2022error}. Both families have expanded in directions important for PDE surrogates: physics-informed operator learning adds residual-based constraints to the learned map \citep{wang2021pideeponet,li2024pino}; Fourier-operator variants address general geometries and multiphase-flow prediction \citep{li2023geofno,wen2022ufno}; variants of DeepONet include multi-input operators \citep{jin2022mionet} and POD-DeepONet \citep{lu2022comprehensive}, in which the learned trunk is replaced with a fixed basis obtained by proper orthogonal decomposition (POD) of the training fields. For this paper, the relevant design space includes both function-to-function operators and parameter- or load-to-field surrogates built from operator-learning architectures.

In computational solid mechanics, researchers have developed learned surrogate models to address a range of physical problems. Constitutive and structural response surrogates cover path-dependent plasticity, thermo-viscoplastic transient material behavior, and full-field stress prediction \citep{abueidda2021plasticity, park2026sequential, nie2020stress}. For path-dependent response, these surrogates take the loading history as input through recurrent, convolutional, or attention-based sequence encoders \citep{abueidda2021plasticity, park2026sequential}. Some of these encoders are causal by construction, and the causality check in this study tests whether a trained encoder responds only to past loading. Energy-based neural formulations have been introduced for finite-deformation hyperelasticity \citep{nguyenthanh2020deepenergy}. Operator learning for parametric mechanics problems has advanced by tailoring the operator to the problem at hand: an energy-based variational loss enforces the physics of crack paths \citep{goswami2022variational}, a reduced basis compresses the input--output map \citep{bhattacharya2021modelreduction}, a nonlocal kernel makes the network independent of mesh resolution \citep{you2022nonlocalkernel}, a time-resolution-independent encoding accepts load histories sampled at any time step \citep{abueidda2026timeres}, and a point-cloud encoding handles three-dimensional geometry \citep{he2024geomdeeponet}. A complementary line embeds neural networks inside finite element solvers for nonlocal continuum damage and coupled multiphysics \citep{pantidis2023ifenn, amin2026ifenndeeponet}, within a broader machine-learning and data-mining literature for continuum mechanics \citep{bock2019review}. This literature demonstrates that surrogate models can represent rich mechanics responses, but it also illustrates why surrogate construction remains expert- and task-specific: each study must choose the architecture, training formulation, geometry representation, finite element coupling, and physical structure appropriate to its problem.

Automating those modeling choices is therefore a natural next step. Auto-PINN searches over selected PINN architectures and hyperparameters, using training loss as the search objective \citep{wang2022autopinn}. PINNsAgent integrates physics-guided knowledge replay with a memory-tree reasoning strategy to optimize PINN architectures and hyperparameters \citep{pinnsagent}, while Lang-PINN coordinates PDE, PINN, code, and feedback agents to turn natural-language task descriptions into executable PINN code with validation or feedback steps \citep{langpinn}. More recent agentic SciML systems broaden the search from tuning toward model and method discovery. AgenticSciML combines role-specialized agents, structured debate, retrieval of prior methods, and evolutionary search to generate and select SciML solutions \citep{jiang2026agenticsciml}. ATHENA organizes numerical method discovery as an iterative propose-implement-evaluate process guided by expert-derived blueprints and a composite scientific reward \citep{toscano2025athena}. GRAFT-ATHENA extends this direction toward cross-problem transfer through structured problem and method representations, metric neighborhoods, and reward-calibrated priors \citep{toscano2026graftathena}. Retrieval and transfer of this kind carry accumulated knowledge from one run to the next; the method cards in this study play the same role but are saved only when single-edit tests and, when required, the physics audit support them (\cref{sec:attribution-method-card-saving}). These systems incorporate verification and diagnostic procedures that assess the generated code, its execution, plan consistency, or the aggregate reward \citep{toscano2025athena, toscano2026graftathena}. Selection depends on training loss, validation error, scientific reward, or trial history, so the predicted fields of a scored surrogate enter selection only through the score.

Language-model agents also automate the stages that produce training data for a surrogate and the stages that use its predictions. MechAgents, PDE-Agents, ALL-FEM, and OpenFOAMGPT write, run, and debug finite element and computational fluid dynamics simulation codes \citep{mechagents, pdeagents, allfem, openfoamgpt}. Knowledge-guided agents propose and review engineering designs \citep{kumar2025autonomous}, chemistry agents plan and execute laboratory experiments using tool-augmented reasoning \citep{boiko2023coscientist, mbran2024chemcrow}, and AI-scientist systems oversee the research workflow from start to finish \citep{lu2026aiscientist}. Each of these systems is judged by whether its output executes, achieves a specified design target, or completes an experiment. The present work concerns the learned surrogate between those stages, and the evidence it asks for is whether the predicted fields satisfy the physics requirements of the problem.

The agentic workflows above and the LLM-mediated roles in \sysname{} build on established LLM-agent patterns. ReAct introduced interleaved reasoning and tool or environment actions \citep{yao2023react}. Multi-agent debate studies answer refinement through interaction \citep{du2024multiagent}. CAMEL studies role-playing communicative agents \citep{li2023camel}, AutoGen provides a framework for composing conversational multi-agent workflows \citep{wu2024autogen}, and MetaGPT structures role-specialized collaboration through standardized operating procedures \citep{hong2024metagpt}. Reflexion uses verbal self-reflection over task feedback \citep{shinn2023reflexion}, and LLM-as-judge work studies model-based evaluation and its biases \citep{zheng2023judging}. In \sysname{} (\cref{sec:framework}), the division of the workflow into specialized roles parallels CAMEL and MetaGPT \citep{li2023camel,hong2024metagpt}; the exchange between the Proposer and the Critic follows the debate pattern \citep{du2024multiagent}; the Debugger's bounded repair attempts apply reflection over execution feedback \citep{shinn2023reflexion}; and the Selector ensemble is a form of model-based judging \citep{zheng2023judging}. Whether a selected surrogate's displacement, stress, or time-history fields satisfy the physics required by the mechanics problem lies outside these patterns, as that evidence comes from fixed routines applied to the predicted fields.

This question connects agentic SciML to two further lines of work: building physical constraints into learned models, and verifying trained networks. One strand builds physical information directly into model construction: governing-equation loss or likelihood terms can support label-free surrogate training \citep{zhu2019physicsconstrained}. Hard-constrained physics-informed neural networks can enforce constraints in inverse design for topology optimization \citep{lu2021hardconstraints}, and conservation laws or other analytic constraints can be built into neural-network architectures or loss functions \citep{beucler2021enforcing}. Another strand analyzes trained neural networks after training. Formal-verification tools can prove specified input-output properties or return counterexamples, but their applicability is limited to ReLU or piecewise-linear networks \citep{katz2017reluplex, katz2019marabou}. Adversarial-robustness research identifies worst-case input perturbations through gradient ascent on a loss function \citep{goodfellow2015adversarial, madry2018adversarial}. The Adversary in this work adapts that search to maximize a physics-check violation over the admissible input domain while holding the trained surrogate fixed. Collectively, these areas demonstrate that physical constraints can be incorporated into learning and that trained networks can be checked or probed. The framework presented next brings these strands into agentic discovery: fixed routines check the predicted fields of every scored candidate, a bounded search over the admissible inputs looks for large violations without reference solutions, and the selected surrogate is reported with that evidence and its error score.

\section{Framework}
\label{sec:framework}

\begin{figure}[t]
\centering
\includegraphics[width=1\textwidth]{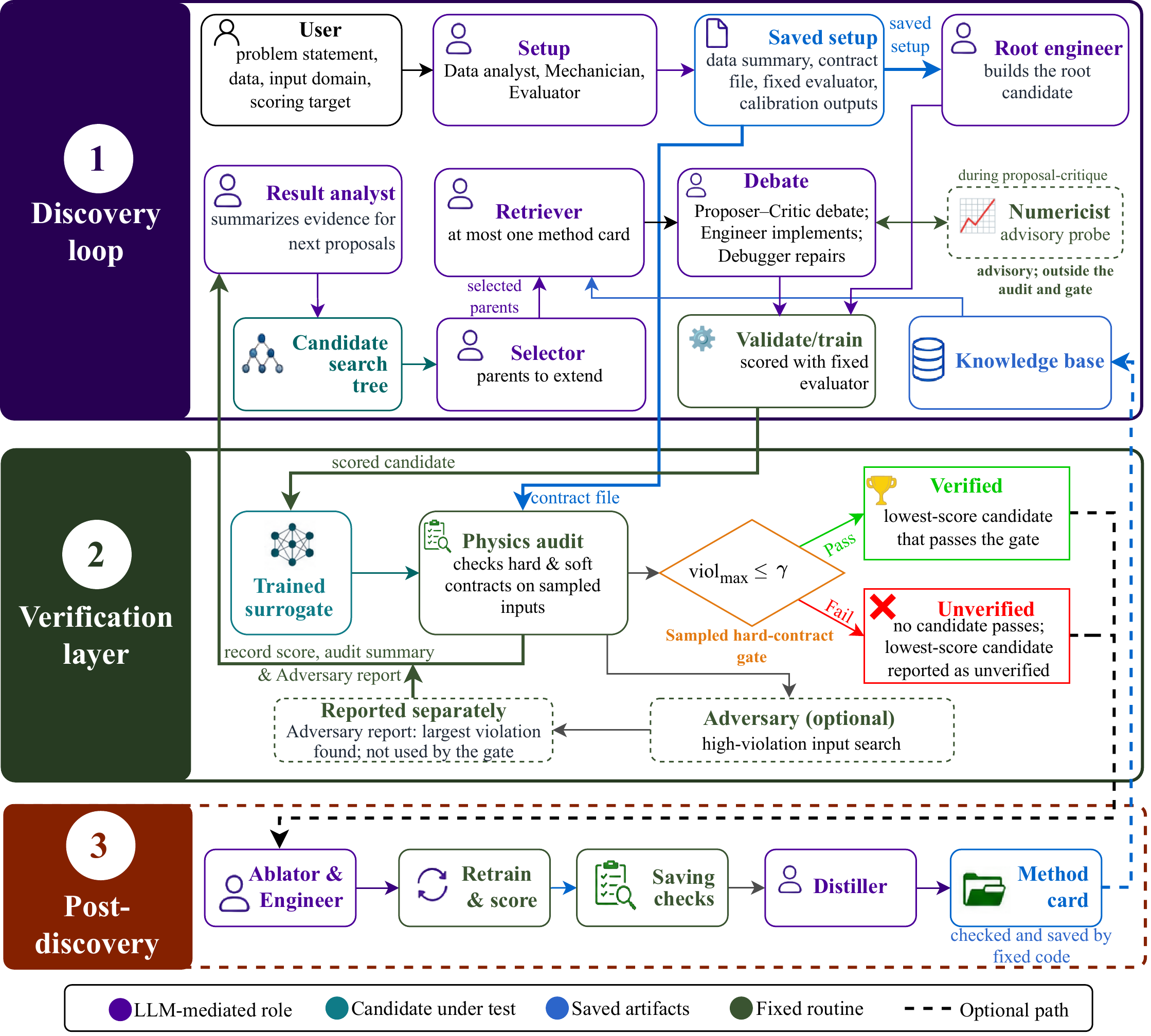}
\caption{\textbf{Audit-enabled \sysname{} workflow.} Panels 1--3 correspond to
\cref{sec:discovery-loop,sec:verification,sec:attribution-method-card-saving} and are expanded in
\cref{fig:discovery-loop,fig:verification,fig:attribution}. Arrows between panels carry each scored
candidate to the physics audit, the contract file from setup to the audit, the audit results back to
the search for later proposals, and, after reporting, the saved method cards to the knowledge base
used in later runs.}
\label{fig:approach}
\end{figure}

\sysname{} integrates physics-check evidence into surrogate discovery while keeping predictive error as a separate score. The framework is designed for agentic SciML more broadly; in this paper, the discovered surrogates are neural operators. Three design choices make the workflow verification-first:
\begin{itemize}
\item LLM-mediated roles propose, implement, and repair trial surrogates, called \emph{candidates}, but do not score or audit them.
\item Fixed routines compute predictive scores and, in audit-enabled runs, test the predicted fields of successfully scored candidates against physics requirements encoded as \textit{contracts}, without relying on the roles' own reports.
\item The scoring code and the contracts, drafted during setup, are fixed before candidate search and remain unchanged throughout it, so candidates cannot influence the criteria used to evaluate them.
\end{itemize}
Each run begins with user-supplied problem information, including the problem statement, data description, admissible input domain, and scoring target (\cref{fig:overview}). The workflow uses this information in three connected parts, which correspond to panels 1--3 of \cref{fig:approach}:
\begin{enumerate}
\item \textbf{Discovery loop} (\cref{sec:discovery-loop}): sets up the run and then carries out a \emph{candidate search}, in which the contract descriptions, together with the physics-check results of earlier candidates, inform later proposals.
\item \textbf{Verification layer} (\cref{sec:verification}): audits each scored candidate and reports the selected surrogate as verified only if it passes the sampled hard-contract gate, a pass/fail rule on the physics checks required for verification.
\item \textbf{Post-discovery attribution and method-card saving} (\cref{sec:attribution-method-card-saving}): tests whether single edits reproduce score improvements and saves qualifying edits as method cards, short design notes for retrieval in later runs.
\end{enumerate}
\Cref{sec:execution-safeguards} describes what each run records, and \cref{sec:design} defines how the numerical examples compare and report surrogates. \Cref{tab:llm-roles,tab:fixed-components} in Appendix~\ref{app:roles} list the roles and routines.

\subsection{Discovery loop}
\label{sec:discovery-loop}

\begin{figure}[t]
\centering
\includegraphics[width=\linewidth]{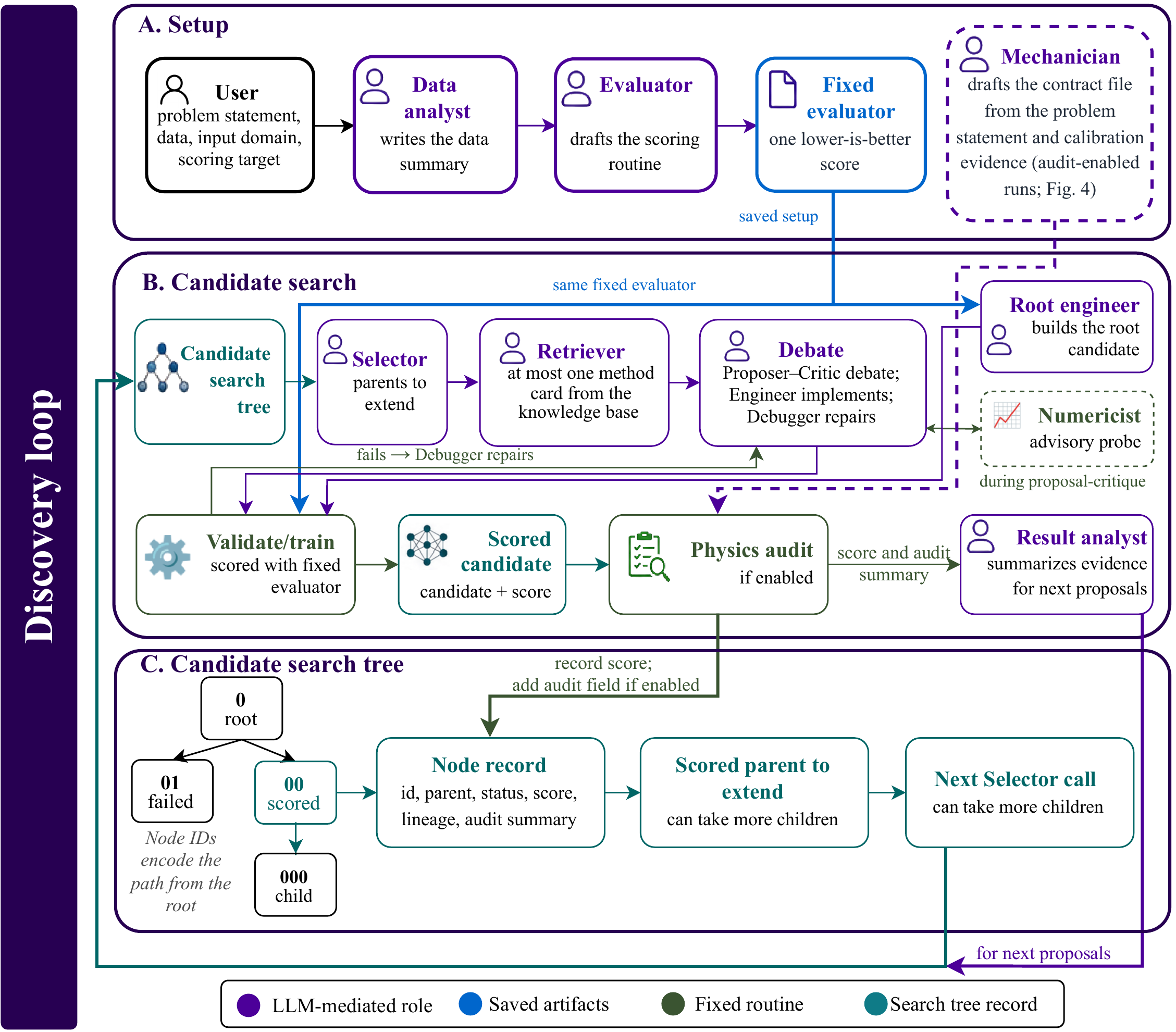}
\caption{\textbf{Discovery loop.} Setup turns the supplied problem information into the fixed evaluator, which supplies one lower-is-better score, and, when auditing is enabled, the contract file drafted by the Mechanician; both remain unchanged throughout the run. The Root engineer then builds the first candidate from the saved setup. During candidate search, the workflow selects eligible parents, may retrieve a saved method card for each, proposes and reviews a model change, writes or repairs the candidate implementation, and runs fixed routines for code validation, training, and scoring. When auditing is enabled and a contract file is present, every successfully scored candidate enters the physics audit (\cref{fig:verification}). The Result analyst summarizes scores and any physics-check evidence for later proposals. Scores and audit summaries are stored in the candidate search tree; generated implementations and full audit reports remain in the run record.}
\label{fig:discovery-loop}
\end{figure}

The discovery loop constructs and tests candidate surrogates (\cref{fig:discovery-loop}). Each candidate is implemented as self-contained code covering both the surrogate and its training procedure; an edit can change either. All candidates accept the same inputs and return the predicted fields in the same format, so the fixed evaluator and the physics audit can consistently score and test every candidate. During setup, the \emph{Data analyst} analyzes the training data and writes a data summary in the context of the supplied problem statement, and the \emph{Evaluator} drafts the scoring code. This code is saved as the \emph{fixed evaluator} and reused unchanged throughout the run, so every candidate's prediction error is computed with the same measure on the same validation data. A candidate's \emph{score} is the resulting scalar error; lower scores indicate better predictive accuracy. After setup, the \emph{Root engineer} implements the root candidate from the supplied problem statement and data summary. The search grows a \emph{candidate search tree} from this root, with each later node representing a \textit{child} obtained by editing an already scored \textit{parent}. Each node records its status and lineage, along with its score and, when available, an audit summary, so the tree preserves the search's design history. A scored candidate that has not reached its child limit is an \emph{eligible parent}. The search runs for a specified number of iterations; in each iteration, up to a prescribed number of eligible parents are selected, and one child is attempted from each. The selection always includes the lowest-score candidate found so far while it remains eligible. If more parents are eligible than can be selected, the \emph{Selector} chooses the rest by combining LLM-ensemble votes with deterministic rules; otherwise, all eligible parents are selected.

For each selected parent, the workflow develops and tests a candidate modification. The \emph{Retriever} may first supply one method card from the knowledge base of saved cards (\cref{sec:attribution-method-card-saving}); if none is suitable, candidate search proceeds without one. The \emph{Proposer} then suggests a candidate modification, and the \emph{Critic} reviews it before implementation. When enabled, the \emph{Numericist}, a fixed routine, runs numerical diagnostics requested by the Proposer during the proposal--critique exchange, computing quantities such as the effective rank of the training targets. These diagnostics are advisory and do not enter the physics audit or the sampled hard-contract gate (\cref{sec:verification}). The \emph{Engineer} implements the plan, and fixed routines run code validation, train the candidate, and score it with the fixed evaluator. When code validation fails, the \emph{Debugger} diagnoses the failure and the \emph{Engineer} attempts repairs within a configured retry limit.

Physics requirements and evidence also shape the search itself. When auditing is enabled, the contract descriptions are added to the Proposer's context. For each audited candidate, the workflow records the physics-check results and any Adversary report (\cref{sec:verification}). The \emph{Result analyst} summarizes this evidence and the scores into a candidate analysis used by later Proposer, Critic, and Retriever steps. Later proposals can therefore respond to physics-check failures as well as to predictive error. The Result analyst can also inspect diagnostic plots comparing predictions with validation reference fields. If auditing is disabled or no contract file is present, the workflow scores and compares candidates, then reports the selected surrogate without an in-run physics claim.

\subsection{Verification layer}
\label{sec:verification}

When auditing is enabled, the framework encodes problem-specific physics requirements as contracts. Each contract is a check on the surrogate's input--output map, specified in the contract file by its check type, parameters, tolerance, and a hard or soft designation. For example, a superposition contract for a linear-elastic surrogate evaluates two loads and their sum at fixed material parameters, with all three inputs admissible; its violation is the relative \(L^2\) discrepancy between the prediction for the sum and the sum of the two predictions. Hard contracts enter the sampled hard-contract gate, which determines whether the selected surrogate is reported as verified; soft contracts are recorded only as diagnostics.

\paragraph{Contracts and the Mechanician.} The supported check types fall into families such as linearity or scaling, symmetry or antisymmetry, boundary or initial-condition consistency, output bounds, zero-input response, and causality. Some check types support both finite-dimensional parameters and function-valued inputs, such as load histories, with definitions specific to each input mode. A family applies only where the underlying physics implies it; superposition, for example, holds in linear elasticity but does not generally hold in plasticity. The contract file should therefore include only properties implied by the governing equations, the constitutive model, and the boundary and initial conditions, expressed in terms of the audited inputs and outputs. Appendix~\ref{app:contract-schemas} lists the physics-check types used in the numerical examples.

During setup, fixed calibration routines first run predefined diagnostics on the reference training data and record problem-specific residuals and empirical scaling estimates. The \emph{Mechanician}, an LLM-mediated role, then drafts the contract file from the supplied problem statement and this calibration evidence. The file configures pre-implemented checks that fixed routines execute. When calibration residuals are available, the workflow prompts the Mechanician to choose tolerances at least \(3\times\) the corresponding measured residuals, aiming to leave a margin above the discrepancies measured on the reference data; this factor is not enforced. Tolerances, the hard or soft designations, and the input domain are therefore setup assumptions, archived for review.

\begin{figure}[t]
\centering
\includegraphics[width=\linewidth]{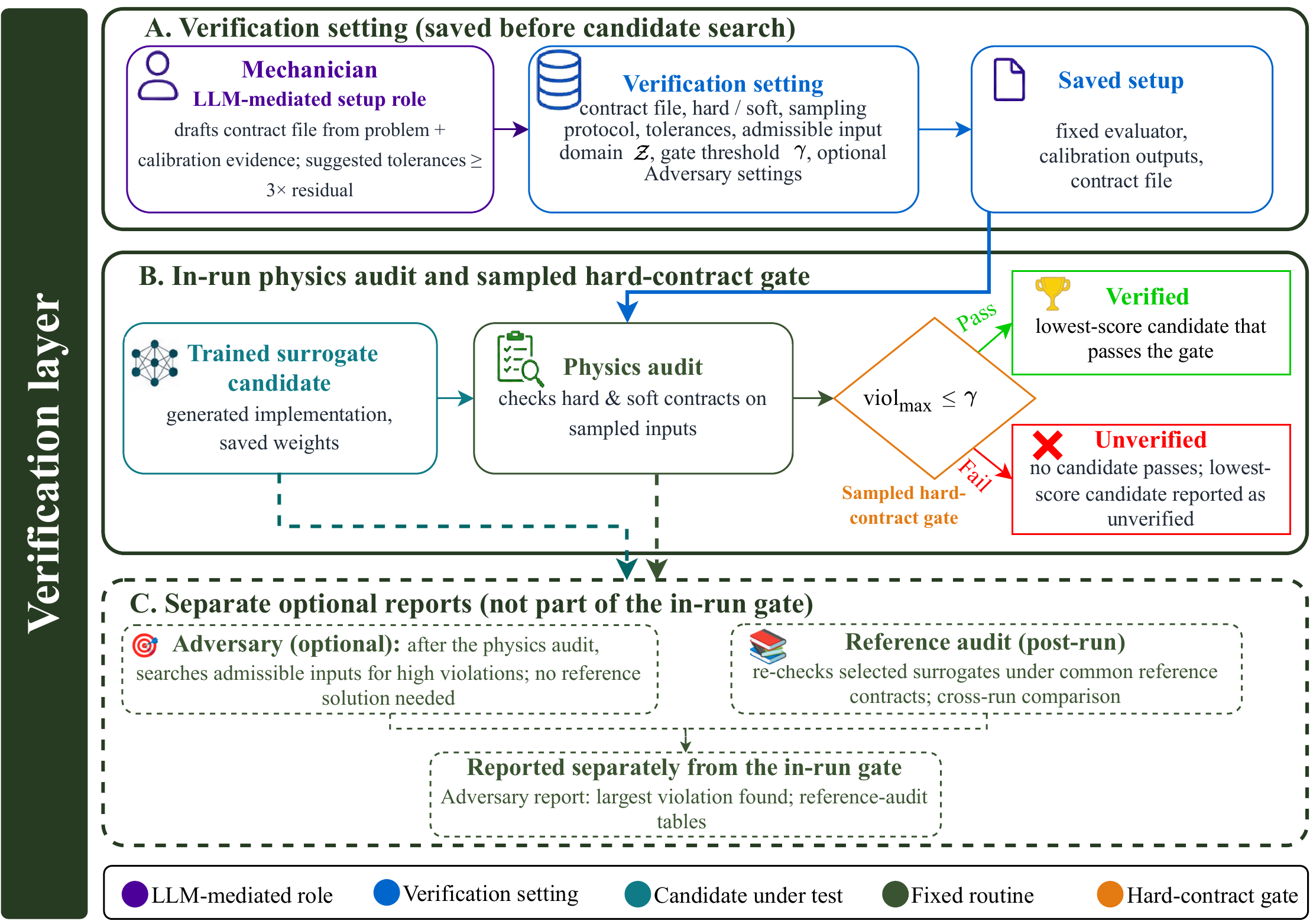}
\caption{\textbf{Verification layer.} During setup, the LLM-mediated Mechanician drafts the contract file from the supplied problem statement and calibration evidence; the file and the audit settings then stay fixed during candidate search. The system holds each successfully scored candidate fixed while the physics audit computes sampled \violmax{} over its active hard contracts, and the in-run gate compares this value with the configured threshold. When enabled, the Adversary then runs a separate high-violation input search that needs no reference solution fields. Post-run reference audits (\cref{sec:design}) can audit selected surrogates under a reference contract file.}
\label{fig:verification}
\end{figure}

\paragraph{Physics audit and gate.} The audit draws its inputs from the admissible input domain \(\mathcal{Z}\), a space of problem inputs such as parameters or load histories. A candidate is \emph{verified} when it passes the gate under the run's contract file and audit settings, which stay fixed during candidate search (\cref{fig:verification}).

The physics audit is applied to every candidate that receives a valid score. It attempts to evaluate every contract in the contract file, hard and soft, on a finite set of \emph{probes}. A probe is the set of admissible inputs within \(\mathcal{Z}\) needed for one evaluation of a check: for the superposition check, two sampled loads and their sum; for the zero-input check, a zero input. A check that cannot be evaluated, for example because it raises an execution error, is skipped and reported with its reason; the remaining contracts are called \emph{active}.

For each active hard contract \(c\), the audit computes a dimensionless raw violation \(v_c(q)\) for each probe \(q\) in its probe set \(\mathcal{S}_c\). The sampled hard-contract statistic is
\begin{equation}
\violmax{} = \max_{c \in \mathcal{H}} \frac{\max_{q \in \mathcal{S}_c} v_c(q)}{\tau_c},
\label{eq:violmax}
\end{equation}
where \(\tau_c>0\) is the contract tolerance and \(\mathcal{H}\) is the set of active hard contracts. A candidate passes the sampled hard-contract gate when \(\violmax{}\le\gamma\), where \(\gamma\) is the gate threshold. With \(\gamma=1\), used in all reported audit-enabled runs, passing means that every active hard contract is within its tolerance at every evaluated probe. If no hard contract remains active, the candidate cannot pass the gate or meet the audit requirement for method-card saving.

At final reporting, an audit-enabled run selects the lowest-score candidate that passes the gate; if none passes, it reports the lowest-score candidate as unverified. During the search, a candidate that fails the gate can still be extended. The run can also report the Pareto front of score and sampled \(\violmax{}\), formed by the non-dominated candidates. After the run, selected surrogates can be audited under a reference contract file (\cref{sec:design}).

\begin{figure}[!htbp]
\centering
\includegraphics[width=\linewidth]{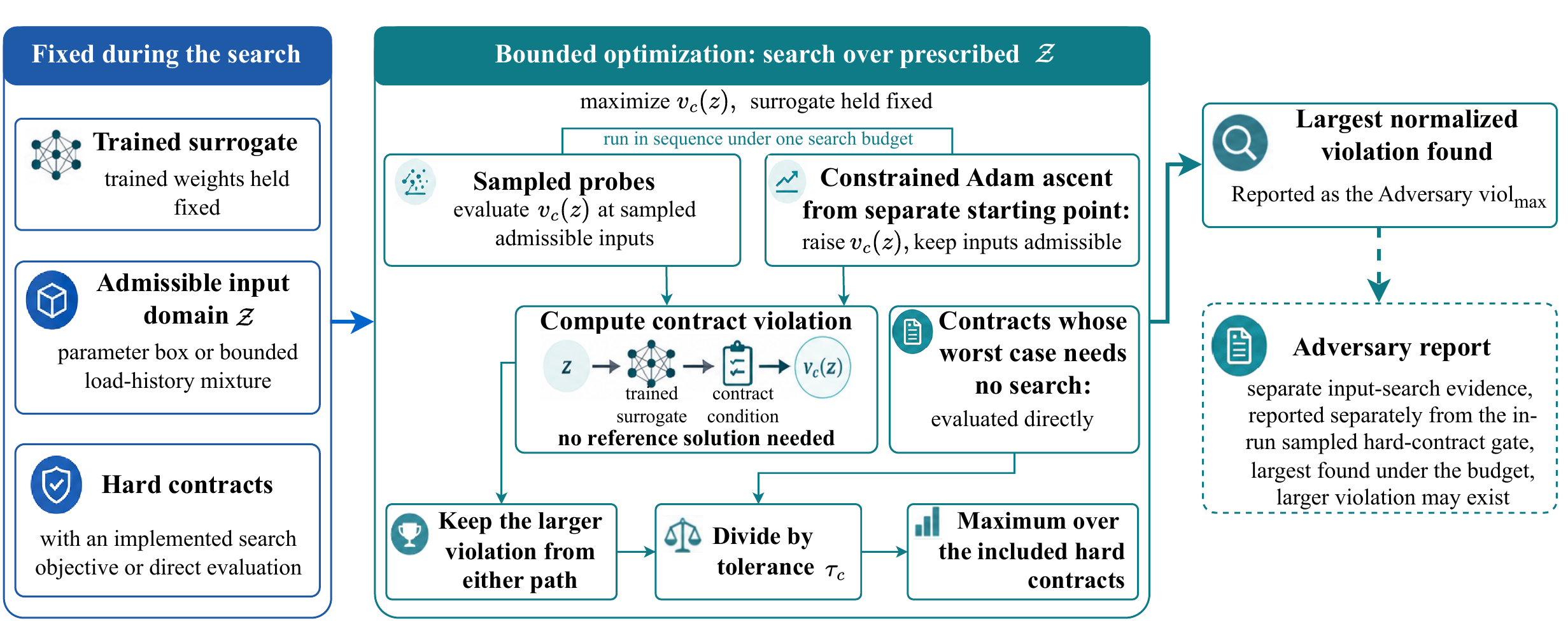}
\caption{\textbf{Adversary: a no-reference high-violation input search.} When enabled, the Adversary runs after the physics audit as a fixed routine with no LLM-mediated step. For hard checks with a search objective, it searches over input-side variables \(z\) to seek larger contract violations \(v_c(z)\), keeping the constructed inputs within \(\mathcal{Z}\) and the trained surrogate fixed. Training modifies weights to minimize predictive error against reference data, while the Adversary alters admissible inputs to maximize physics-check violations. The Adversary reports the largest violation found under the chosen budget and parameterization, so a larger violation may still exist beyond the searched budget.}
\label{fig:adversary}
\end{figure}

\paragraph{Adversary.} Because the audit evaluates a finite set of probes, a candidate that passes the gate may still violate a contract at other admissible inputs. When enabled, the \emph{Adversary}, a fixed routine, searches for such inputs after the audit of each scored candidate, whether or not the candidate passes (\cref{fig:adversary}). It keeps the trained surrogate unchanged, varies only the input-side variables \(z\) that parameterize admissible probes, and seeks to maximize the violation \(v_c(z)\), an objective that needs no reference solution fields. The Adversary searches hard checks that have a search objective, directly evaluates supported checks that require no search, and excludes the remaining hard checks from its report. In the reported runs, for each hard check with a search objective, it evaluates its own sampled probes, separately runs constrained Adam ascent from a second set of starting points while keeping the constructed inputs admissible, and retains the largest violation found by either procedure. The search variables are normalized coordinates in the prescribed box for parameter-vector inputs and bounded mixture coefficients over admissible load histories for function-valued inputs.

The largest tolerance-normalized violation found over the included hard checks is listed in the result tables as the Adversary \(\violmax{}\). This value depends on the chosen budget and parameterization and does not prove that no larger violations exist. Adversary results are reported separately and do not enter the sampled \(\violmax{}\) or the gate, although they can inform later proposals through the candidate analysis (\cref{sec:discovery-loop}).

\subsection{Post-discovery attribution and method-card saving}
\label{sec:attribution-method-card-saving}

\begin{figure}[t]
\centering
\includegraphics[width=\linewidth]{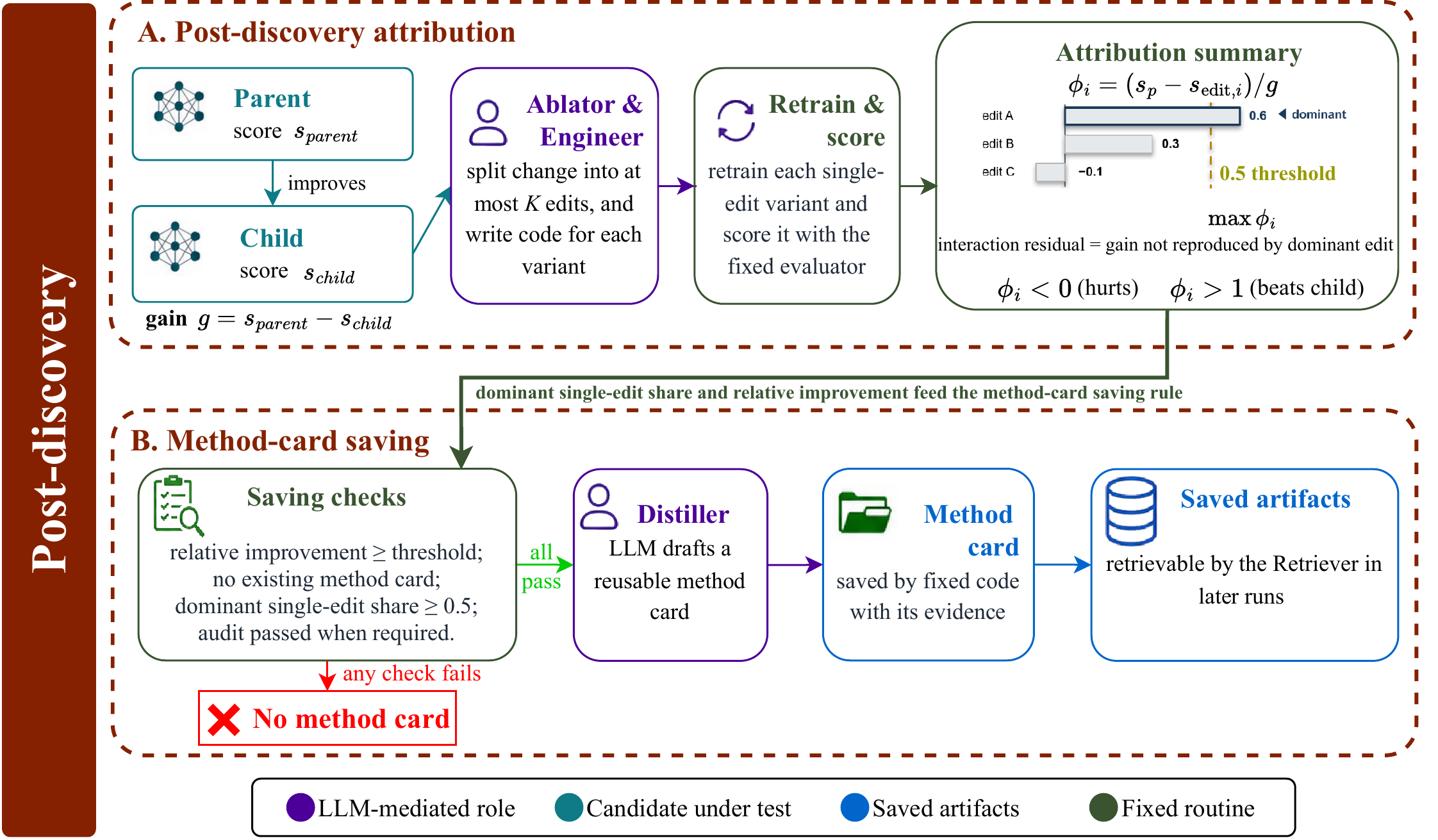}
\caption{\textbf{Post-discovery attribution and method-card saving.} The figure follows one improving parent--child edge after candidate search. Each bar reports an illustrative \(\phi_i\), the share of the score gain reproduced by one isolated edit, and the dashed \(0.5\) line marks the threshold on the dominant single-edit share used by the method-card saving rule. A method card is saved only when the dominant single-edit share meets the threshold and the remaining saving conditions in \cref{sec:attribution-method-card-saving} are satisfied.}
\label{fig:attribution}
\end{figure}

A child often differs from its parent by several bundled edits, such as a new architecture and a new learning-rate schedule. Not every edit necessarily contributes to the score gain. Post-discovery attribution asks whether that gain can be reproduced by one isolated edit. It examines the improving parent--child edges on the path from the root to the selected surrogate, subject to a configured minimum relative improvement and a limit on the number of edges analyzed. \Cref{fig:attribution} illustrates this optional procedure for one edge.

The \emph{Ablator}, an LLM-mediated role, splits an improving parent--child code change into at most \(K\) testable edits, where \(K\) is set in the run configuration. For each edit, the Engineer creates a parent-code variant containing only that edit, and fixed routines retrain and rescore the variant under the same fixed evaluator. For parent score \(s_{\mathrm{parent}}\), child score \(s_{\mathrm{child}}\), and single-edit score \(s_{\mathrm{edit},i}\), define \(g=s_{\mathrm{parent}}-s_{\mathrm{child}}\) and
\begin{equation}
\phi_i=\frac{s_{\mathrm{parent}}-s_{\mathrm{edit},i}}{g}.
\label{eq:single-edit-share}
\end{equation}
The largest \(\phi_i\) among successfully scored variants is the \emph{dominant single-edit share}, the fraction of the step's gain that one edit reproduces on its own. A share above 1 means that the isolated variant scores better than the full child under the same evaluator. When all measured shares are small, no successfully tested edit reproduces much of the gain on its own. The gain may depend on several edits acting together, but these tests do not distinguish that possibility from retraining variability.

Method-card saving is enabled separately and uses the attribution results. Beyond basic validity and duplicate checks, the saving checks require a relative improvement of at least a separately configured minimum and a dominant single-edit share of at least \(0.5\), so that one tested edit reproduces at least half of the gain on its own. When physics compliance is required for saving, the original child must also have passed its active hard checks in the sampled audit; the isolated variants are retrained and rescored but are not audited as part of these attribution tests. The \emph{Distiller}, an LLM-mediated role, drafts the card text after these saving checks pass. Fixed routines then save the card, with its supporting evidence, in a knowledge base that the Retriever draws on in later runs.

\subsection{Reproducibility and traceability}
\label{sec:execution-safeguards}

Each run keeps a \emph{run record} linking reported scores and physics-check results to the models and settings that produced them. It includes the fixed evaluator, any calibration outputs, the contract file (if used), search and audit settings, candidate lineage, implementations, logs, scores, audit summaries, and full audit reports. Archived contract files specify the sampling settings and check parameters, and audit reports record the sampling seed and the number of probes evaluated for each check. Execution errors and timeouts are also logged; candidates for which no score is obtained count as failed attempts. Rescoring an archived candidate with its run's saved fixed evaluator allows its recorded score to be checked; comparisons across runs use a common implementation of the error metric (\cref{sec:design}). The run record makes reported results traceable and checkable, but it does not make a search repeatable. Neither the LLM-mediated roles nor candidate training is deterministic, so rerunning a search with the same configuration can produce different candidates and a different selected surrogate (\cref{sec:conclusion}).

\subsection{Comparison and reporting protocol}
\label{sec:design}

Each of the four numerical examples (\cref{sec:example-elasticity,sec:example-elastodynamics,sec:example-cantilever-progression,sec:example-plate-with-a-hole}) uses the workflow described above and a common reporting protocol. Each example specifies its governing problem, target solution field, and reference data, together with its error metric, contract files, admissible input domain, and search and audit settings. Depending on the problem, the surrogate inputs may include loads, source terms, boundary data, material parameters, or load histories.

The champion of a search is the one surrogate the search reports from all the candidates it produced. In an audit-enabled search, the champion is selected by the final-reporting rule in \cref{sec:verification}: the lowest-score candidate that passes the sampled hard-contract gate, or, if none passes, the lowest-score candidate, reported as unverified. In contrast, an audit-disabled baseline search runs without in-run physics auditing and selects its champion by the lower-is-better score alone. The audit-enabled workflow also uses the contract descriptions, physics-check results, and any Adversary reports to inform later proposals (\cref{sec:discovery-loop}). Comparisons between the two searches therefore do not isolate the contribution of any single component, such as the gate (\cref{sec:conclusion}).

Each example reports each champion's validation error, measured against reference solution fields, and, where audited, whether its outputs satisfy the stated physics checks. All predictive comparisons are within-example comparisons that use the same error measure and validation data. Within each comparison, these validation errors are recomputed with a common implementation of the error metric and are consistent with the scores recorded during the searches. A raw score in one problem should not be read against a score in another, because each problem has its own field scale, validation data, and error metric. Because validation scores also guide the searches, these errors are not independent estimates of generalization error (\cref{sec:discovery-loop}).

Physics comparisons between champions use post-run reference audits, because the audit-disabled champion has no in-run physics evidence and the audit-enabled champion was audited under its own run's contract file. These audits apply the physics audit and, where reported, the Adversary to the already-trained champions under a single reference contract file that we define based on the governing equations, constitutive model, and boundary and initial conditions, with tolerances informed by available calibration evidence. They take place after selection, so they do not affect which candidate the search chose. A champion passes under the same rule as the sampled hard-contract gate with \(\gamma=1\) (\cref{eq:violmax}). Because a check can be hard in one contract file and soft in another, pass/fail labels and normalized violations must be read with the contract file that produced them.

\section{Overview of the numerical examples}
\label{sec:examples-overview}

The four numerical examples are solid-mechanics problems of increasing complexity, selected to place growing demands on the surrogate, both in the input it must encode and in the mechanics it must respect. The surrogate input is a vector of load and material parameters in static linear elasticity (\cref{sec:example-elasticity}); a prescribed boundary-motion history in transient linear elastodynamics (\cref{sec:example-elastodynamics}); a load and a material description, progressing from isotropic to orthotropic constants and then to a nodal stiffness field, for a cantilever (\cref{sec:example-cantilever-progression}); and material parameters with a cyclic displacement history in elastoplasticity, ranging from perfect plasticity to combined isotropic--kinematic hardening (\cref{sec:example-plate-with-a-hole}). In each case, the surrogate predicts complete nodal fields on a fixed finite element mesh, with reference solutions provided by finite element analysis on the same mesh. Within each example, the audit-enabled and audit-disabled searches use the same reference data and error metric (\cref{sec:design}). The champion architectures are described in Appendix~\ref{app:architectures}.

The physics checks are derived from the governing problem. For linear elastic problems, they encode relations that the exact solution satisfies for any admissible input, including load superposition, reciprocal stiffness scaling, and prescribed displacement conditions; a candidate can satisfy these relations by construction. The transient problem introduces causality, which has no static analog. In the elastoplastic problem, superposition and stiffness scaling do not apply, but the boundary conditions do. The other checks encode requirements that a nonlinear, history-dependent material still satisfies: zero response to zero input, causality, bounds on the displacements and the von Mises stress, and sign and monotonicity conditions that follow from the Poisson effect, the loading direction, and the ordering of stiffness. \Cref{tab:examples-overview} summarizes the four examples and their main outcomes, discussed below.

\begin{table}[t]
\centering
\caption{\textbf{Overview of the numerical examples.} Each example compares an audit-enabled search with an audit-disabled search on the same reference data; the physics column lists the hard contracts checked, and the outcome column summarizes the champion comparison.}
\label{tab:examples-overview}
\small
\begin{tabularx}{\linewidth}{@{}>{\raggedright\arraybackslash}p{0.17\linewidth} >{\raggedright\arraybackslash}X >{\raggedright\arraybackslash}X >{\raggedright\arraybackslash}X@{}}
\toprule
Example & Problem and surrogate map & Physics checked & Main outcome \\
\midrule
Parametric linear elasticity (\cref{sec:example-elasticity}) & Unit square in plane stress under a uniform top-edge traction, with roller supports; $(f_x,f_y,\Emod,\poisson)\mapsto\mathbf{u}(\mathbf{x})$ & Load superposition, reciprocal-\Emod{} scaling, roller conditions & Both champions pass; the audit-enabled champion is more accurate \\
\addlinespace
Transient elastodynamics (\cref{sec:example-elastodynamics}) & Thin strip fixed at one edge and driven at the other; boundary-motion history $\mapsto\mathbf{u}(\mathbf{x},t)$ & Load superposition and scaling, zero input, zero initial displacement, prescribed edge motion, causality & Nearly equal accuracy; the audit-disabled champion fails causality \\
\addlinespace
Progressive cantilever (\cref{sec:example-cantilever-progression}) & Cantilever plate under an end traction; material isotropic, orthotropic, then a nodal stiffness field; traction and material description $\mapsto\mathbf{u}(\mathbf{x})$ & Traction linearity, stiffness scaling, fixed-edge conditions, and level-specific material-structure relations & Audit-enabled champion more accurate and far smaller (Level~1), more accurate but larger (Level~2), smaller and passing but less accurate (Level~3) \\
\addlinespace
Plate with a hole, cyclic plasticity (\cref{sec:example-plate-with-a-hole}) & Plate quadrant under a cyclic prescribed displacement; material parameters and displacement history $\mapsto\mathbf{u}(\mathbf{x},t),\,\sigma_{\mathrm{vM}}(\mathbf{x},t)$ & Symmetry and prescribed-displacement conditions, zero input, causality, displacement and stress bounds, Poisson effect, monotonicity & All audit-enabled champions pass within the parameter limit; every audit-disabled champion fails at least one check \\
\bottomrule
\end{tabularx}
\end{table}

\paragraph{Parametric linear elasticity (\cref{sec:example-elasticity}).} Both champions are DeepONets in which the tractions enter only through exact compliance prefactors, so load superposition holds by construction in both, and coordinate factors that vanish on the roller edges enforce the roller conditions. Only the audit-enabled champion also incorporates reciprocal stiffness scaling, by restricting its branch input to Poisson's ratio. Both champions pass the post-run reference audit, and the audit-enabled champion is more accurate. Because the gate did not change the audit-enabled selection, the accuracy difference reflects the distinct candidate sets produced by the two searches.

\paragraph{Transient elastodynamics (\cref{sec:example-elastodynamics}).} The two champions have nearly the same mean validation error. The audit-enabled champion is linear and causal by construction. In contrast, the audit-disabled champion uses the complete loading history in both time directions, so its early predictions can depend on later boundary motion. As a result, it fails the hard causality check in the reference audit. A mechanics requirement that the predictive metric does not test therefore distinguishes the two surrogates despite their similar accuracy.

\paragraph{Progressive cantilever elasticity (\cref{sec:example-cantilever-progression}).} With the geometry, loading, mesh, and output fixed, the material description progresses from homogeneous isotropic to homogeneous orthotropic, and finally to a spatially heterogeneous stiffness field. Levels~2 and~3 also compare audit-disabled searches with and without the knowledge base accumulated in earlier levels. For the homogeneous materials, the audit-enabled search finds the more accurate champion. In Level~1, this champion is also far smaller, because an exact compliance factorization and a fixed POD trunk reduce the problem to a small coefficient map. In Level~2, the accuracy gain comes with a larger model. For the heterogeneous field, the audit-enabled champion passes all hard checks but is less accurate than the audit-disabled champion: its low-dimensional representation omits the spatial detail that heterogeneous fields require, and its training stopped at the time limit. Passing the physics checks alone therefore does not establish representational capacity. In these single runs, reuse of the knowledge base improves the subsequent audit-disabled searches.

\paragraph{Plate with a hole under cyclic loading (\cref{sec:example-plate-with-a-hole}).} The material progresses from elastic--perfectly plastic to isotropic hardening, and then to combined isotropic--kinematic hardening. The surrogates predict the displacements and the von Mises stress over the loading history, and both searches are limited to the same number of trainable parameters. At the first level, the two champions perform similarly. At the second level, the audit-enabled champion is substantially more accurate. At the third level, the audit-disabled search finds a viable champion only after the parameter limit is removed, and that champion is far larger and more accurate. In contrast, the audit-enabled search finds a champion within the limit. All audit-enabled champions satisfy every contract, whereas every audit-disabled champion violates at least one, such as the zero-input response, causality, or the von Mises stress bound. The contracts therefore guide the search toward viable candidates under a strict model-size constraint.

These examples show why validation error and physics-check results must be reported separately. They include cases where both champions pass, where accuracy is similar but one fails causality, where the audit-enabled champion passes but is less accurate, and where the audit-enabled champions pass whereas their audit-disabled counterparts fail or need far larger models. The following sections discuss each example, and \cref{sec:conclusion} summarizes the findings.

\section{Parametric linear elasticity}
\label{sec:example-elasticity}

\begin{figure}[t]
\centering
\includegraphics[width=0.5\linewidth]{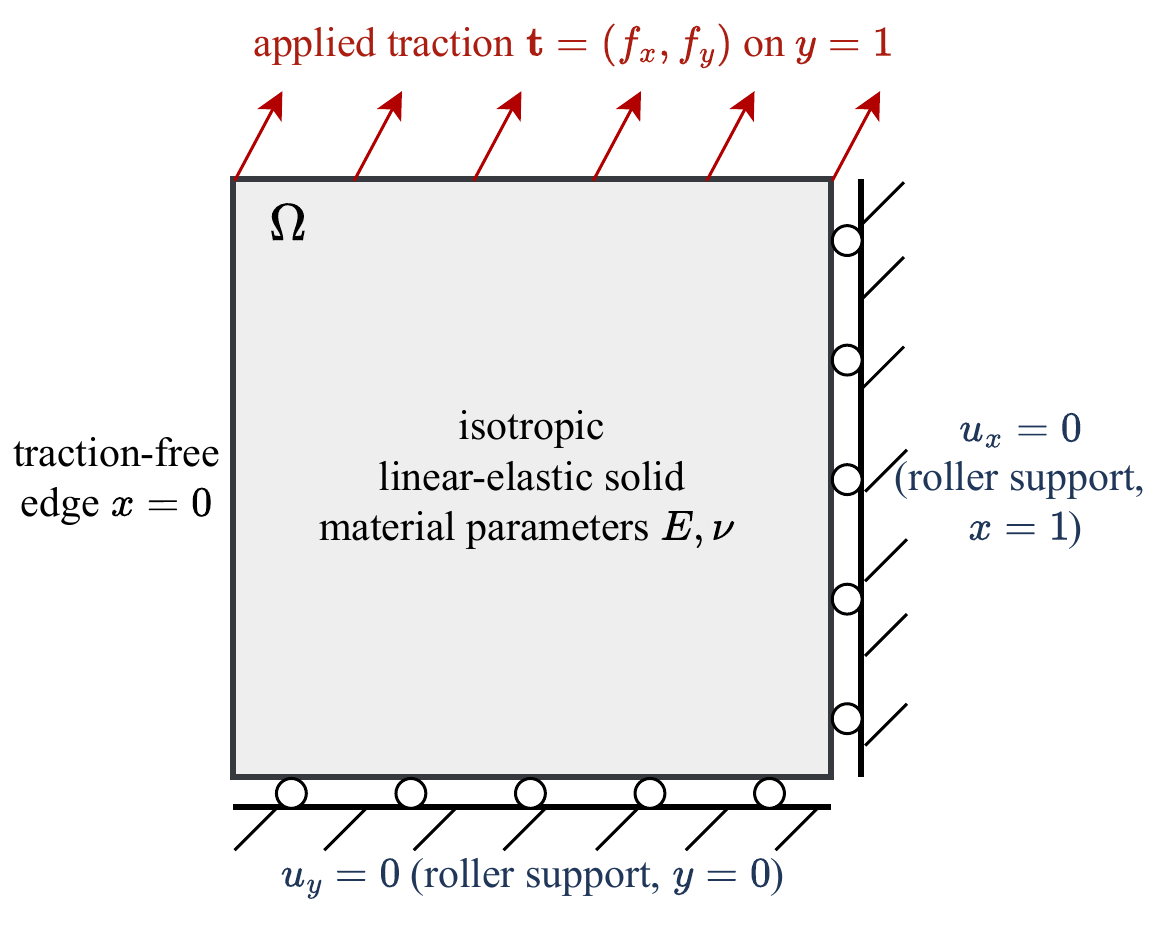}
\caption{\textbf{Parametric linear elasticity: domain and boundary conditions.} The unit square $\Omega=[0,1]^2$ is an isotropic linear-elastic solid with material parameters $\Emod$ and $\poisson$. A uniform traction $\bar{\mathbf{t}}=(f_x,f_y)$ is applied on the top edge $y{=}1$, with $f_x$ the tangential and $f_y$ the normal component, each varying over $[-0.3,0.3]$ across samples (red arrows show one representative direction). The left edge $x{=}0$ is traction-free, and there is no body force. Roller supports fix the normal displacement on the other two edges, $u_y=0$ on the bottom edge $y{=}0$ and $\ux=0$ on the right edge $x{=}1$, leaving the tangential component free to slide.}
\label{fig:elasticity-domain}
\end{figure}

The first numerical example concerns a small-strain linear-elastic body in plane stress occupying $\Omega=[0,1]^2$. Let $\mathbf{u}=(u_x,u_y)$ denote the displacement, $\boldsymbol{\varepsilon}$ the infinitesimal strain, and $\boldsymbol{\sigma}$ the Cauchy stress. In the absence of body forces, the static boundary-value problem has the following form
\begin{align}
\nabla\!\cdot\boldsymbol{\sigma} &= \mathbf{0} && \text{in } \Omega, \\ \mathbf{u} &= \bar{\mathbf{u}} && \text{on } \Gamma_u, \\ \boldsymbol{\sigma}\cdot\mathbf{n} &= \bar{\mathbf{t}} && \text{on } \Gamma_t,
\end{align}
where $\Gamma_u$ and $\Gamma_t$ denote displacement and traction boundaries and $\mathbf{n}$ is the outward unit normal. The small-strain kinematics and isotropic constitutive relation are
\begin{equation}
\boldsymbol{\varepsilon}(\mathbf{u}) = \tfrac12\left(\nabla\mathbf{u}+(\nabla\mathbf{u})^{T}\right), \qquad \boldsymbol{\sigma}=\mathbb{C}(\Emod,\poisson):\boldsymbol{\varepsilon}.
\label{eq:elastic-constitutive}
\end{equation}
Here, $\mathbb{C}(\Emod,\poisson)$ is the isotropic plane-stress elasticity tensor with Young's modulus \Emod{} and Poisson's ratio \poisson{}. The loading is a uniform traction $\bar{\mathbf{t}}=(f_x,f_y)$ on the top edge $y{=}1$, with tangential component $f_x$ and normal component $f_y$. The left edge $x{=}0$ is traction-free, and the bottom and right edges are frictionless rollers, with $u_y=0$ on $y{=}0$ and $\ux=0$ on $x{=}1$. Across samples, $f_x$ and $f_y$ each vary over $[-0.3,0.3]$, \Emod{} over $[5,20]$, and \poisson{} over $[0.15,0.35]$. These ranges define the admissible input domain $\mathcal{Z}$. The spatial domain $\Omega$ is discretized on a fixed mesh of \num{2601} nodes. The surrogate maps the four scalars $\inpvec=(f_x,f_y,\Emod,\poisson)$ to both displacement components at every mesh node. \Cref{fig:elasticity-domain} summarizes the domain, the top-edge traction, and the roller supports.

\paragraph{Physics checks.} The solutions of this boundary-value problem satisfy several relations that can be used to check a surrogate, and the physics checks encode them as finite-sample tests on predicted displacement fields (Appendix~\ref{app:contract-schemas} describes each check type):
\begin{itemize}
    \item \textbf{Load superposition}. At fixed material parameters, the solution for two tractions applied together equals the sum of the two separate solutions.
    \item \textbf{Reciprocal-\Emod{} scaling}. At fixed traction and \poisson{}, the displacement solutions for moduli \Emod{} and \(c\Emod{}\) satisfy \(\mathbf{u}(c\Emod{})=\mathbf{u}(\Emod{})/c\), because the stiffness is proportional to \Emod{} and the prescribed displacements are zero.
    \item \textbf{Right-edge $u_x$ condition}. The normal displacement vanishes on the right roller edge: $u_x=0$ on $x{=}1$.
    \item \textbf{Bottom-edge $u_y$ condition}. The normal displacement vanishes on the bottom roller edge: $u_y=0$ on $y{=}0$.
    \item \textbf{Load-sign reversal}. Reversing both traction components reverses the sign of the solution.
    \item \textbf{Amplitude bound}. The predicted displacements remain within the bounds set in the contract file.
\end{itemize}
A scalar predictive-error score compares each prediction with its own reference field and does not test relations between predictions at different inputs. This example therefore asks two questions: whether the audit-enabled search reaches an accurate surrogate that satisfies these relations, and whether the audit-disabled champion satisfies them as well. In the audit-enabled run, the Mechanician-drafted contract file specifies all six checks as hard contracts except the amplitude bound, which is soft. The reference contract file of the post-run audit keeps load superposition, reciprocal-\Emod{} scaling, and the two roller conditions as hard contracts. It reports the amplitude bound and load-sign reversal as diagnostics, the latter because, in this linear-elastic setting, it follows from the same linearity that gives load superposition.

\paragraph{Reference data and metric.} The reference displacement fields are finite element solutions on the fixed mesh described above for \num{5000} cases, with \inpvec{} sampled uniformly over the ranges that define $\mathcal{Z}$. These were randomly divided into \num{4000} training and \num{1000} validation cases; the audit-enabled and audit-disabled runs used the same split. The predictive metric is the mean per-sample relative $L^2$ error of the full displacement vector $(\ux,u_y)$ on the validation set (\rltwo{}, lower is better). In each search, the run's fixed evaluator computes this metric to score candidates (\cref{sec:discovery-loop}). Because relative error can be sensitive when the reference displacement is very small, we also report percentile and worst-case errors.

\paragraph{Champions.} Both champions are DeepONets in which the traction components enter only through the fixed compliance prefactors $f_x/\Emod$ and $f_y/\Emod$ (Appendix~\ref{app:arch-elasticity}). These prefactors multiply four sets of branch coefficients, one pair per displacement component, and two coordinate factors that vanish on the roller edges multiply the assembled output. In both champions, load superposition and load-sign reversal therefore hold for any trained weights, and the two normal-displacement conditions are satisfied identically. The coefficients multiplying $f_y/\Emod$ in \ux{} and $f_x/\Emod$ in $u_y$ represent the coupled responses: horizontal displacement under normal traction (from the Poisson effect) and vertical displacement under tangential traction. The champions differ in how Young's modulus enters. In the audit-enabled champion, the branch network receives only Poisson's ratio, so reciprocal-\Emod{} scaling also holds by construction. In mechanics terms, linearity in the tractions and reciprocal-\Emod{} scaling imply that the exact solution has the form
\begin{equation}
\mathbf{u}(\mathbf{x};f_x,f_y,\Emod,\poisson)=\frac{f_x}{\Emod}\,\mathbf{G}_x(\mathbf{x};\poisson)+\frac{f_y}{\Emod}\,\mathbf{G}_y(\mathbf{x};\poisson),
\label{eq:elasticity-superposition}
\end{equation}
where $\mathbf{G}_x$ and $\mathbf{G}_y$ are the displacement fields for unit tangential and unit normal traction, respectively, at unit modulus. The audit-enabled champion adopts this form by construction, with learned approximations $\hat{\mathbf{G}}_x$ and $\hat{\mathbf{G}}_y$ that combine spatial bases from the trunk with \poisson{}-dependent coefficients from the branch. In the audit-disabled champion, the branch receives both \Emod{} and \poisson{}, so its coefficients can also vary with \Emod{}, and reciprocal-\Emod{} scaling holds only as closely as training makes them independent of \Emod{}.

The two champions also differ in their spatial bases: in the audit-enabled champion, a trunk with two output heads produces separate bases of $64$ functions for \ux{} and $u_y$, whereas the audit-disabled champion's trunk produces a single basis of $160$ functions that both components share before the coordinate factors are applied. The factorization in \cref{eq:elasticity-superposition} was not specified in the supplied problem statement. In the audit-enabled run, the workflow identified the relations above during setup and encoded them as contracts, which were added to the Proposer's context. The first improving step of this run introduced the compliance prefactors while the branch still received both \Emod{} and \poisson{}, as in the audit-disabled champion; the final champion restricts the branch input to \poisson{}. The audit-disabled search reached the same prefactors and coordinate factors without any contracts in the Proposer's context, but its champion keeps \Emod{} as a branch input.

\begin{figure}[t]
\centering
\includegraphics[width=\linewidth]{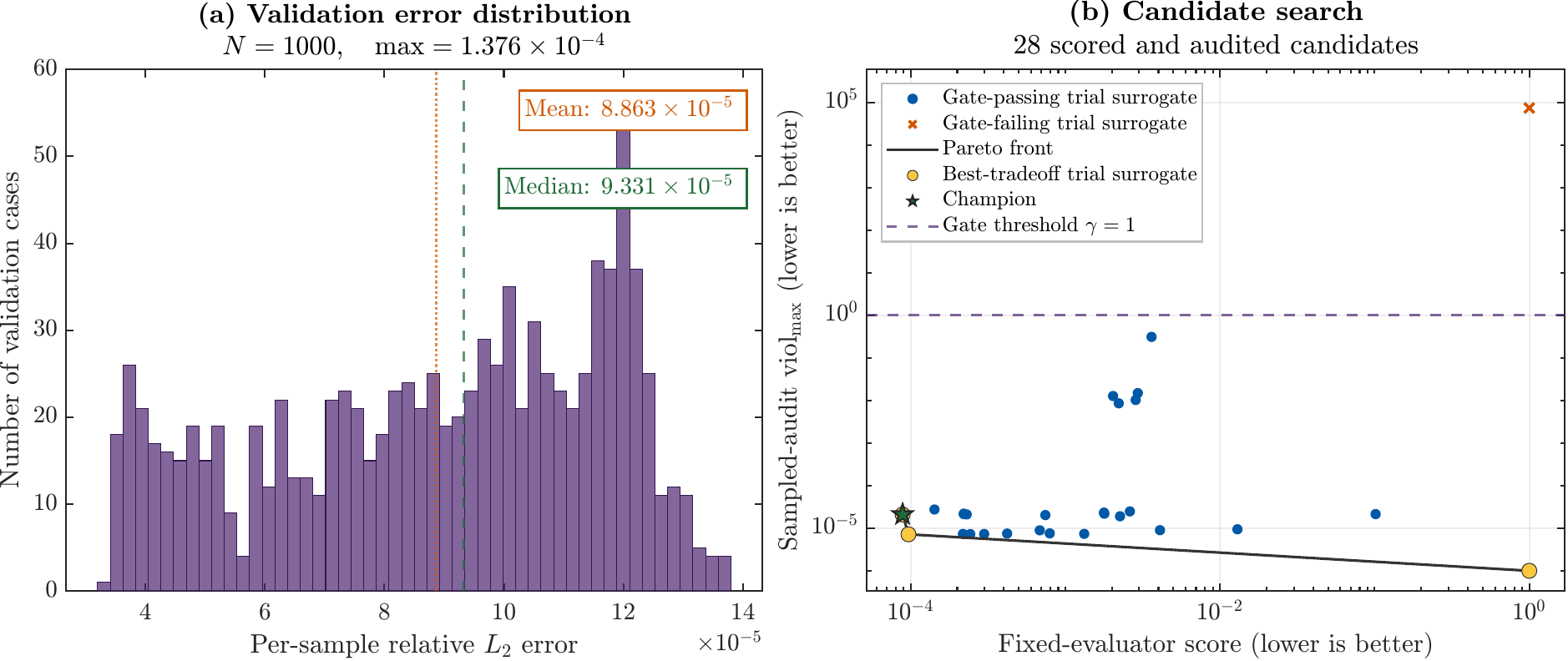}
\caption{\textbf{Parametric linear elasticity: error distribution and candidate search.} (a) Distribution of the per-sample relative $L^2$ error of the audit-enabled champion over the \num{1000} validation cases (mean \num{8.864e-5}, median \num{9.331e-5}, maximum \num{1.376e-4}). (b) Sampled \violmax{} against score for every successfully scored and audited candidate in the audit-enabled search, on logarithmic axes; the dashed line marks the gate threshold $\gamma=1$. The solid line joins the non-dominated candidates that form the Pareto front (yellow circles). The champion (star) has the lowest score among gate-passing candidates. The only gate-failing candidate has a score near one. Sampled violations of exactly zero are plotted at the axis floor of $10^{-6}$ so that they appear on the logarithmic scale. Panel (b) uses in-run sampled audit values from the candidate search; \cref{tab:elasticity-aggregate} reports the post-run reference audit of both champions.}
\label{fig:elasticity-hist}
\end{figure}

\begin{figure}[t]
\centering
\includegraphics[width=\linewidth]{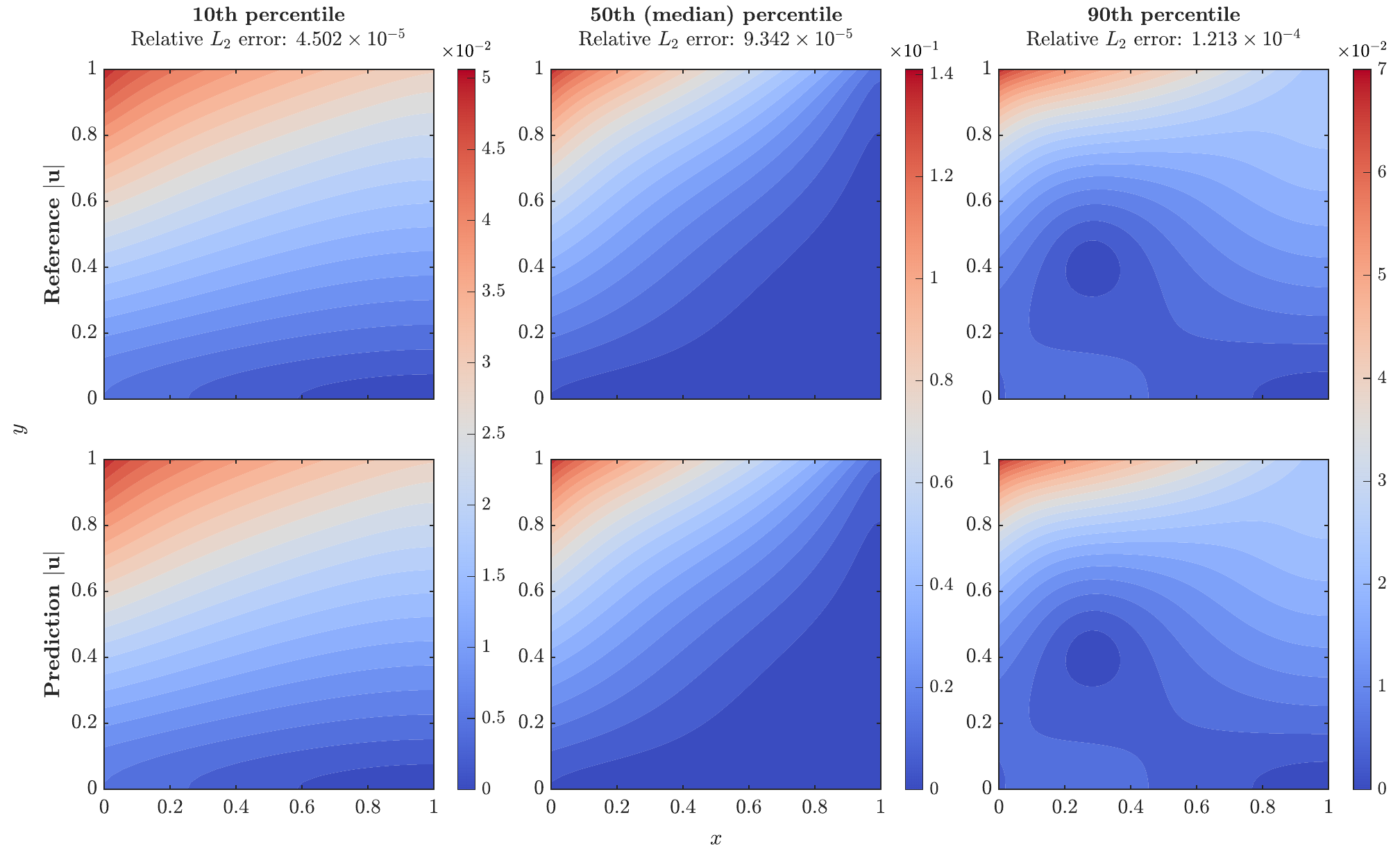}
\caption{\textbf{Parametric linear elasticity: displacement-magnitude fields of the audit-enabled champion.} Reference (top row) against the audit-enabled champion's prediction (bottom row) of the displacement magnitude $|\mathbf{u}|$ over the unit square $\Omega=[0,1]^2$, for the validation cases at the $10$th, $50$th (median), and $90$th percentiles of the per-sample relative $L^2$ error ($n=\num{1000}$); each column header gives the selected case's error. The two panels of a column share one color scale, with the power-of-ten factor above the bar.}
\label{fig:elasticity-fields}
\end{figure}

\paragraph{Accuracy.} On the validation set, the audit-enabled champion attains a \rltwo{} of \num{8.864e-5}, a factor of about $2.4$ lower than the audit-disabled champion's \num{2.103e-4}; the $95$th-percentile and worst-case errors differ by similar factors (\cref{tab:elasticity-aggregate}). For both champions, the worst-case error is only about $1.6$ times the mean. For the audit-enabled champion, \cref{fig:elasticity-hist} (a) shows the full error distribution, and the predicted and reference displacement magnitudes agree closely at the $10$th, $50$th, and $90$th error percentiles (\cref{fig:elasticity-fields}). Only the $90$th-percentile case has an interior local minimum, where the responses to the two traction components partly cancel. Such cancellation can increase the relative error because the errors of the learned approximations $\hat{\mathbf{G}}_x$ and $\hat{\mathbf{G}}_y$ need not cancel to the same extent.

\begin{table}[htbp]
\centering
\caption{\textbf{Parametric linear elasticity: accuracy and reference audit.} Validation displacement-field error of the audit-enabled and audit-disabled champions, and their post-run reference audit without retraining. The reference-audit status is determined by the sampled \violmax{} (at most $1$ pass); the Adversary \violmax{} is reported separately and does not affect this status. Both champions pass.}
\label{tab:elasticity-aggregate}
\small
\begin{tabular}{lcccccc}
\toprule
Champion & \rltwo{} & val.\ p95 & val.\ max & sampled \violmax{} & Adversary \violmax{} & status \\
\midrule
Audit-enabled & \num{8.864e-5} & \num{1.249e-4} & \num{1.376e-4} & \num{1.708e-4} & \num{2.087e-3} & pass \\
Audit-disabled    & \num{2.103e-4} & \num{3.148e-4} & \num{3.376e-4} & \num{1.080e-3} & \num{2.582e-3} & pass \\
\bottomrule
\end{tabular}
\end{table}

\paragraph{Candidate search.} The audit-enabled run generated \num{28} candidates. All \num{28} were scored and audited; \num{27} passed the in-run sampled hard-contract gate, and the one failure was a degenerate candidate whose predicted field collapsed toward zero, resulting in a normalized violation nearly five orders of magnitude above the threshold. It failed all five hard contracts, and load superposition gave the largest violation: a load-independent output offset left a nonzero residual, which the relative normalization amplified because the compared combination of predicted fields nearly cancels. Besides the champion, the Pareto front in \cref{fig:elasticity-hist} (b) contains a candidate with a slightly higher score and lower sampled violation, and a candidate whose prediction is identically zero. All three pass the gate. The zero prediction has a score of one, since its relative $L^2$ error against any nonzero reference field is one, yet it satisfies all five in-run hard contracts, each of which holds trivially for a zero field. The score and contracts therefore provide complementary evidence: a low score alone does not establish that the contracts hold, and passing the contracts alone does not establish predictive accuracy. At final reporting, the gate did not change the selection: the lowest-score candidate in the audit-enabled search also passed. The audit-disabled run successfully generated \num{28} scored candidates using the same framework, with the Adversary and numerical diagnostics also disabled. The two searches produced different candidates; this pair of runs cannot separate the possible contributions of the contract descriptions, physics-check and Adversary feedback, numerical diagnostics, and run-to-run variation to the accuracy gap (\cref{sec:conclusion}).

\paragraph{Physics verification.} Unlike the validation error, the physics checks use no reference solutions. Both champions pass the stated hard checks in the post-run reference audit (\cref{tab:elasticity-aggregate,tab:elasticity-contracts}). The largest sampled violation comes from the audit-disabled champion's reciprocal-\Emod{} scaling, the only reference hard contract that its architecture does not enforce exactly; it corresponds to a raw violation of about \num{1e-5}. The audit-enabled champion's largest values come from load superposition and reflect floating-point round-off. The Adversary (\cref{fig:adversary}) finds larger violations, about $12$ and $2.4$ times the sampled maxima of the audit-enabled and audit-disabled champions, respectively (\cref{tab:elasticity-aggregate}), but within its search budget these remain more than two orders of magnitude below the threshold. These small violations do not establish a mechanically meaningful advantage for either champion under the stated checks and tolerances. For the audit-disabled champion, whose architecture does not enforce reciprocal-\Emod{} scaling and whose validation error does not test it, the audit shows that this relation holds within tolerance at the sampled probes. For the audit-enabled champion, whose hard-checked relations hold by construction, the audit remains informative because it tests the saved model through its complete prediction interface, including input normalization and output transformations. A load-independent output offset, like that of the failed candidate above, would violate these relations, whereas the champion's fixed multiplicative output scales preserve them.

\begin{table}[htbp]
\centering
\caption{\textbf{Parametric linear elasticity: per-check violations.} Worst normalized violation for each check under the reference contract file (sampled audit / Adversary report); a champion passes when every active hard check has a sampled normalized violation at most $1$. Soft diagnostics are not included in the Adversary report (n.a.).}
\label{tab:elasticity-contracts}
\small
\begin{tabular}{llcc}
\toprule
Check & hard/soft & audit-enabled champion & audit-disabled champion \\
\midrule
load superposition & hard & \num{1.708e-4} / \num{2.087e-3} & \num{9.46e-5} / \num{3.03e-4} \\
reciprocal-\Emod{} scaling & hard & 0 / 0 & \num{1.080e-3} / \num{2.582e-3} \\
right-edge \ux{} & hard & 0 / 0 & 0 / 0 \\
bottom-edge $u_y$ & hard & \num{9.83e-15} / \num{1.44e-14} & \num{9.87e-15} / \num{1.45e-14} \\
load-sign reversal & soft\textsuperscript{$\dagger$} & 0 / n.a. & 0 / n.a. \\
amplitude bound & soft\textsuperscript{$\dagger$} & 0 / n.a. & 0 / n.a. \\
\bottomrule
\end{tabular}

\vspace{2pt} {\footnotesize \textsuperscript{$\dagger$}Soft diagnostic (does not gate).}
\end{table}

\paragraph{Post-discovery attribution.} The attribution tests (\cref{sec:attribution-method-card-saving}) examined four improving steps on the path to the audit-enabled champion (\cref{tab:elasticity-attribution}). In the first step, which introduced the compliance prefactors, the physics-structured DeepONet branch edit reproduces $149$\% of the step's score gain on its own, so evaluating only the full step would hide this edit's larger improvement. In the three later steps, no successfully scored single-edit variant reproduces at least half of the gain, the attribution threshold for method-card saving; in one of them, even the best-scoring variant scores worse than its parent. Consequently, only the first step's dominant edit is saved as a method card. These tests do not establish whether combinations of edits or retraining variability account for the later steps' gains. Because they compare variants within the audit-enabled search, they also do not explain the accuracy gap between the two champions, which come from separate searches and both apply compliance prefactors.

\begin{table}[htbp]
\centering
\caption{\textbf{Parametric linear elasticity: saved method card from post-discovery attribution.} The reported share of gain is the dominant single-edit share, the largest \(\phi_i\) among successfully scored variants of the corresponding step (\cref{eq:single-edit-share}), expressed as a percentage. A value above $100$\% means that the isolated variant scores better than the full child. Other improvement steps were checked but did not result in a saved method card.}
\label{tab:elasticity-attribution}
\small
\begin{tabular}{llc}
\toprule
Improvement step & saved isolated edit & share of gain \\
\midrule
first improvement & physics-structured DeepONet branch & 149\% \\
\bottomrule
\end{tabular}
\end{table}

\paragraph{What this example shows.} Under the stated checks and tolerances, the answer to both opening questions is yes, and the example supports three conclusions. First, the audit-enabled search produced the more accurate champion in this comparison, and this champion passes all stated hard checks. The gate did not change which candidate was selected, so the accuracy difference comes from the candidates that each search produced. This difference belongs to the audit-enabled workflow as a whole, in which the contract descriptions, physics-check results, Adversary reports, and numerical diagnostics all inform later proposals. Second, whichever champion is more accurate, the post-run reference audit gives evidence that the validation error alone does not establish. Under the same reference contract file, both champions satisfy the hard contracts within tolerance at the sampled probes. Third, the factorization in \cref{eq:elasticity-superposition} makes the relative error of the audit-enabled champion depend on only two quantities: the load direction and \poisson{}. For a nonzero load, if the load magnitude or \Emod{} of a validation case changes, the predicted and reference fields scale by the same factor. The relative error of that case therefore remains essentially unchanged, even outside the sampled ranges. This holds as long as the linear-elastic model applies, \Emod{} stays above the implementation's lower limit, and the reference displacement norm is not vanishingly small. The argument does not establish accuracy at load directions or Poisson ratios that were not sampled. The audit-disabled champion enforces the same load-magnitude scaling but not exact reciprocal-\Emod{} scaling, because its branch receives \Emod{}; for this champion, the argument therefore covers changes in load magnitude but not in \Emod{}. The transient example that follows provides the complementary case. There, the two champions have almost the same mean validation error, but the audit-disabled champion fails the hard causality check of the post-run reference audit.

\section{Transient elastodynamics}
\label{sec:example-elastodynamics}

\begin{figure}[t]
\centering
\includegraphics[width=\linewidth]{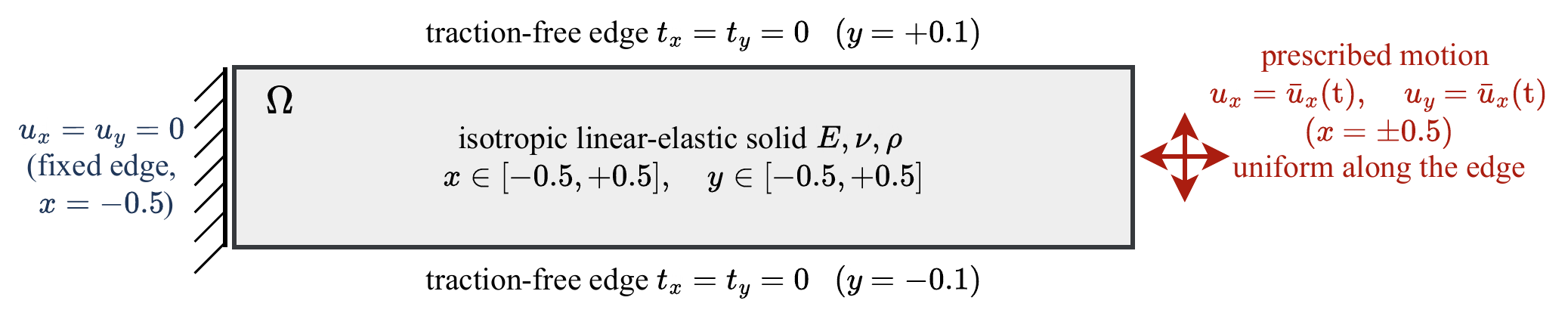}
\caption{\textbf{Transient elastodynamics: domain and boundary conditions.} The thin strip $\Omega=[-0.5,0.5]\times[-0.1,0.1]$ is an isotropic linear-elastic solid in plane stress with material properties fixed for every sample, $\Emod=10$, $\poisson=0.25$, and $\rho=1$ in consistent units; only the loading history varies between samples. The left edge $x{=}{-}0.5$ is fully fixed, $\ux=u_y=0$. The right edge $x{=}{+}0.5$ follows a prescribed time-dependent motion $\bar{\mathbf{u}}(t)=(\bar u_x(t),\bar u_y(t))$ over $t\in[0,2]$, uniform along the edge so the edge translates as a whole; the red arrows show the two motion components. The long edges $y{=}{\pm}0.1$ are traction-free, and the body starts from rest, with zero initial displacement and velocity. The surrogate maps this right-edge motion history to the full transient displacement field $(\ux,u_y)$.}
\label{fig:elastodynamics-domain}
\end{figure}

The second numerical example concerns a thin rectangular linear-elastic body under transient loading. The strip $\Omega = [-0.5, 0.5] \times [-0.1, 0.1]$ (length $1$, height $0.2$) has the same material properties in every sample: Young's modulus $\Emod = 10$, Poisson's ratio $\poisson = 0.25$, and density $\rho = 1$, in consistent units. It is discretized with second-order triangular finite elements on a mesh with \num{4681} nodes. Starting from zero initial displacement and velocity, the solid is driven by a prescribed motion of its right edge; the left edge is fixed, and the long edges are traction-free (\cref{fig:elastodynamics-domain}). With the small-strain kinematics and isotropic constitutive relation of \cref{eq:elastic-constitutive}, here under plane stress, the displacement field $\mathbf{u}(\mathbf{x}, t) = (\ux, u_y)$ obeys
\begin{align}
\rho\,\ddot{\mathbf{u}} &= \nabla\!\cdot\boldsymbol{\sigma} && \text{in } \Omega \times (0, T], \\ \mathbf{u}(\mathbf{x}, 0) &= \mathbf{0}, \quad \dot{\mathbf{u}}(\mathbf{x}, 0) = \mathbf{0} && \text{in } \Omega, \\ \mathbf{u} &= \mathbf{0} && \text{on the left edge } x = -0.5, \\ \mathbf{u} &= \bar{\mathbf{u}}(t) && \text{on the right edge } x = +0.5, \\ \boldsymbol{\sigma}\cdot\mathbf{n} &= \mathbf{0} && \text{on the long edges } y = \pm 0.1,
\end{align}
with $T = 2$ and no body force. The prescribed motion $\bar{\mathbf{u}}(t) = (\bar{u}_x(t), \bar{u}_y(t))$ translates the right edge as a whole: it is uniform along the edge and varies only in time. The surrogate receives four time-series channels sampled at \num{100} uniformly spaced time steps over $t \in [0, T]$: the prescribed edge displacements $\bar{u}_x, \bar{u}_y$ and the corresponding velocity channels. The displacement channels define the boundary motion; the velocity channels are the corresponding time derivatives supplied as additional input information, not independent boundary conditions. The surrogate maps this boundary-motion history to the full transient displacement field $(\ux, u_y)$ at every node and time. Unlike the first example, the surrogate input is a sampled time function rather than a parameter vector, and inertia is included in the momentum balance.

Disturbances excited at the moving right edge propagate along the strip and reflect from the fixed left edge, so the response at any node is a history-dependent superposition of traveling waves. Beyond spatial boundary and linearity checks, this problem therefore adds a temporal requirement with no static analog: causality. For a transient elastic solid starting from rest, the displacement at time \(t\) may depend on the boundary motion up to \(t\), but it should not depend on motion prescribed later in the history. A surrogate trained and scored on complete trajectories can have a low mean error while failing this forward-in-time requirement. This example asks the same two questions as the first, with causality among the relations: whether the audit-enabled search reaches an accurate surrogate that satisfies them, and whether the audit-disabled champion satisfies them as well.

\paragraph{Physics checks.} The physics checks reflect the linear transient setting and encode the following relations as finite-sample tests on predicted displacement histories (Appendix~\ref{app:contract-schemas} describes each check type):
\begin{itemize}
    \item \textbf{Load superposition}. The response to a weighted sum of boundary-motion histories equals the same weighted sum of the separate responses.
    \item \textbf{Amplitude scaling}. Scaling the boundary-motion history by a factor scales the response by the same factor.
    \item \textbf{Sign symmetry}. Reversing the sign of the boundary-motion history reverses the sign of the response.
    \item \textbf{Zero response to zero input}. A body that starts from rest and receives no boundary motion remains at rest.
    \item \textbf{Zero initial displacement}. The predicted displacement vanishes at the initial time.
    \item \textbf{Prescribed right-edge motion}. The predicted right-edge displacement follows the prescribed motion $\bar{\mathbf{u}}(t)$.
    \item \textbf{Causality}. For two boundary-motion histories that agree up to a cut time near the midpoint and differ afterward, a causal surrogate must predict the same displacement field at and before that cut time for both histories. The continuation replaces all four input channels and need not join the shared part smoothly, but causality requires the same earlier response for any continuation.
    \item \textbf{Amplitude bound}. The predicted displacements remain within the bounds set in the contract file.
\end{itemize}
Each check defines a dimensionless raw violation (for example, a relative $L^2$ difference for superposition or a response-normalized residual for the zero-input check) and divides it by the check tolerance. In the audit-enabled run, the Mechanician-drafted contract file specifies five hard contracts (load superposition, amplitude scaling, sign symmetry, zero response to zero input, and zero initial displacement) and three soft contracts (causality, the prescribed right-edge motion, and the amplitude bound). The reference contract file specifies six hard contracts, each with a $5$\% tolerance: load superposition, amplitude scaling, zero response to zero input, zero initial displacement, causality, and the prescribed right-edge motion. The causality failure reported below therefore comes from the post-run reference audit, not from the in-run sampled hard-contract gate (\cref{sec:design}). The amplitude bound and sign symmetry, which follows from the same linearity as load superposition, are reported as diagnostics. Calibration evidence recorded before training any candidate shows that the initial displacement is zero at every node in the sampled training histories and that the reference right edge tracks its prescribed displacement channels with a maximum residual of \num{1.6e-7}, about \num{2e-4} of the prescribed-motion scale, so the $5$\% tolerance is about $250$ times this residual, leaving a wide margin above the discrepancies in the reference data.

\paragraph{Reference data and metrics.} The reference trajectories are transient finite element solutions on the fixed mesh described above. Each displacement history was constructed by Gaussian process interpolation through six control points per displacement component, the first fixed at zero displacement at $t=0$ and the remaining five sampled uniformly over $[-0.005,0.005]$. The dataset contains \num{5000} such histories and their reference trajectories, divided into \num{4500} training and \num{500} validation histories; the audit-enabled and audit-disabled runs used the same split. The predictive metric is the mean per-sample relative $L^2$ error of the full displacement history (\rltwo{}, lower is better). Each per-sample error is computed over both components, all mesh nodes, and all time levels, then averaged over the validation set. In each search, the run's fixed evaluator computes this metric to score candidates (\cref{sec:discovery-loop}). Because relative error can be sensitive when a sample's response is small, we also report the $95$th-percentile and maximum validation errors. The \num{500} validation histories also serve as the input bank for the physics checks. The admissible input domain \(\mathcal{Z}\) includes these histories, as well as their bounded combinations and transformations constructed by the checks and the Adversary, such as the shared-prefix histories of the causality check.

\paragraph{Champions.} As prescribed by the supplied problem statement, both champions are temporal DeepONets: a branch encodes the boundary-motion history, and a trunk provides space-time basis functions (Appendix~\ref{app:arch-elastodynamics}). The problem statement did not specify the load-history encoder or trunk decomposition. Both searches selected a factorized trunk, where each basis function is the product of a learned spatial function and a learned temporal function, with separate spatial and temporal networks for each displacement component. The audit-enabled champion uses $32$ basis functions per component, while the audit-disabled champion uses $256$. The load-history encoders differ. In the audit-enabled champion, the branch for each displacement component applies $32$ learned convolution kernels to the four input channels over the $100$-step history, using only present and past inputs at each time step. Each resulting coefficient is a learned weighted sum of the history up to that time, a discrete analog of the Duhamel convolution for a linear transient system. Weighting the basis functions by these coefficients yields the learned correction to the displacement field. In contrast, the audit-disabled champion's branch combines convolutional features averaged over the complete history with the final forward and backward states of a bidirectional GRU, mapping them to a single set of coefficients per history.

The audit-enabled champion imposes the prescribed edge motion through the lifting $\mathbf{g}(x,t)=(x+0.5)\,\bar{\mathbf{u}}(t)$, a field that satisfies both displacement boundary conditions by interpolating linearly between the fixed left edge and the moving right edge; the network learns the departure from this interpolation inside the strip. The learned correction is multiplied by an envelope, the product of $(x+0.5)(0.5-x)$ and a time factor proportional to $t$. The correction therefore vanishes at both ends and at the initial time: it cannot change the prescribed edge displacements, and the complete field has zero initial displacement because $\bar{\mathbf{u}}(0)=\mathbf{0}$. The bias-free linear branch and the lifting also make the complete prediction linear in the input history. Load superposition, amplitude scaling, sign symmetry, and zero response to zero input thus hold by construction for any trained weights. The complete prediction is also causal: the branch uses only present and past inputs, the lifting uses only the current prescribed motion, and the trunk and envelope depend on coordinates and time, not on the load history. The audit-disabled champion has neither a lifting nor an envelope. Its GELU activations and bidirectional GRU make its coefficients nonlinear functions of the input history, and its output adds a learned bias and a fixed normalization offset. Because these coefficients depend on the complete history, its early predictions can depend on later boundary motion. Consequently, none of these relations holds by construction.

The supplied problem statement did not specify the lifting or the envelope that vanishes at both ends and at the initial time; \sysname{} introduced this construction at the first improving step of the audit-enabled run. Although causality was only a soft contract in that run and therefore did not gate selection, the search adopted the causal load branch at its second improving step (\cref{tab:elastodynamics-attribution}). Soft-contract results still enter the evidence available to later proposals (\cref{sec:discovery-loop}), and the first improving candidate had failed the soft causality check, so this feedback may have informed the change. The audit-disabled search had no contracts in the Proposer's context and reached nearly the same mean validation error with none of these relations built in.

\begin{figure}[t]
\centering
\includegraphics[width=\linewidth]{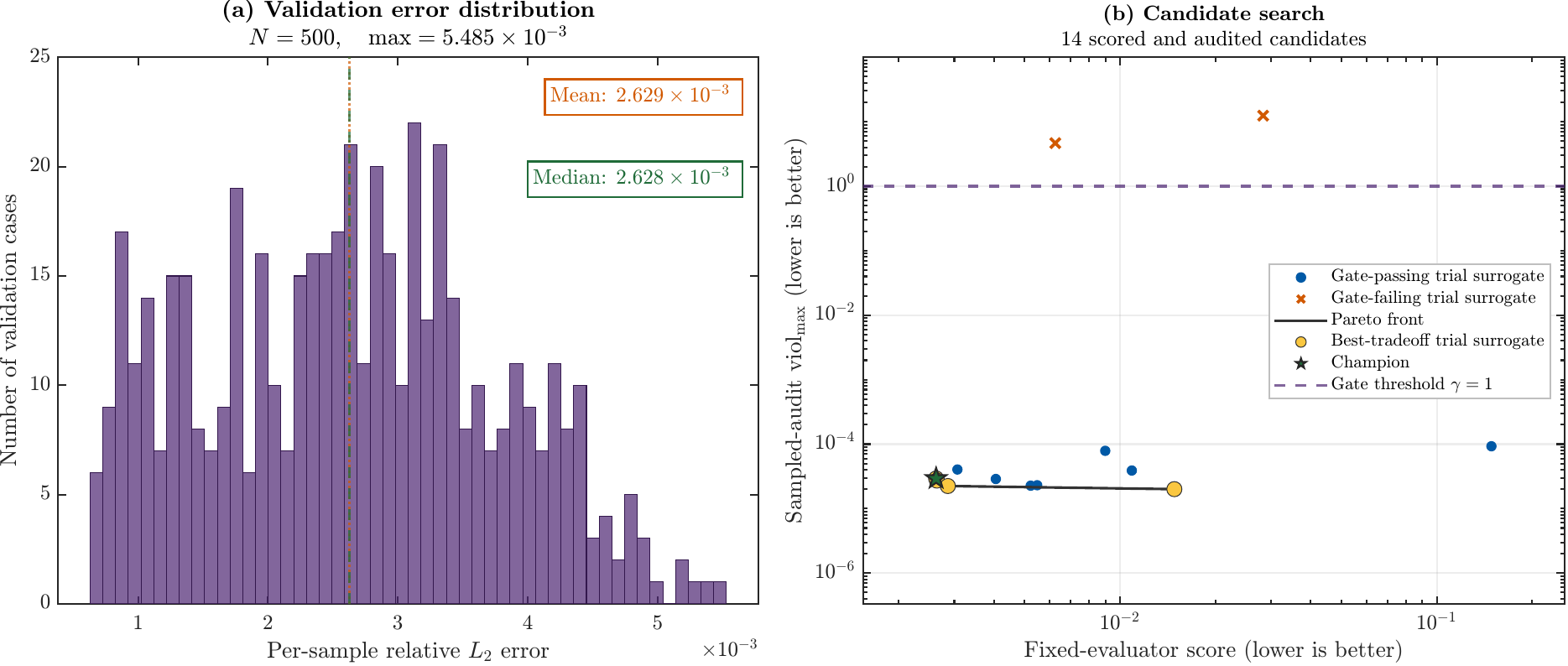}
\caption{\textbf{Transient elastodynamics: error distribution and candidate search.} (a) Distribution of the per-sample relative $L^2$ error of the audit-enabled champion over the full validation set ($n=\num{500}$) (mean \num{2.629e-3}, median \num{2.628e-3}, maximum \num{5.48e-3}). (b) Sampled \violmax{} against score for every successfully scored and audited candidate in the audit-enabled search, on logarithmic axes; the dashed line marks the gate threshold $\gamma=1$. The solid line joins the non-dominated candidates that form the Pareto front (yellow circles). The champion (star) has the lowest score among gate-passing candidates. Panel (b) uses in-run sampled audit values from the candidate search; \cref{tab:elastodynamics-aggregate} reports the post-run reference audit of both champions.}
\label{fig:elastodynamics-dist}
\end{figure}

\begin{figure}[t]
\centering
\includegraphics[width=\linewidth]{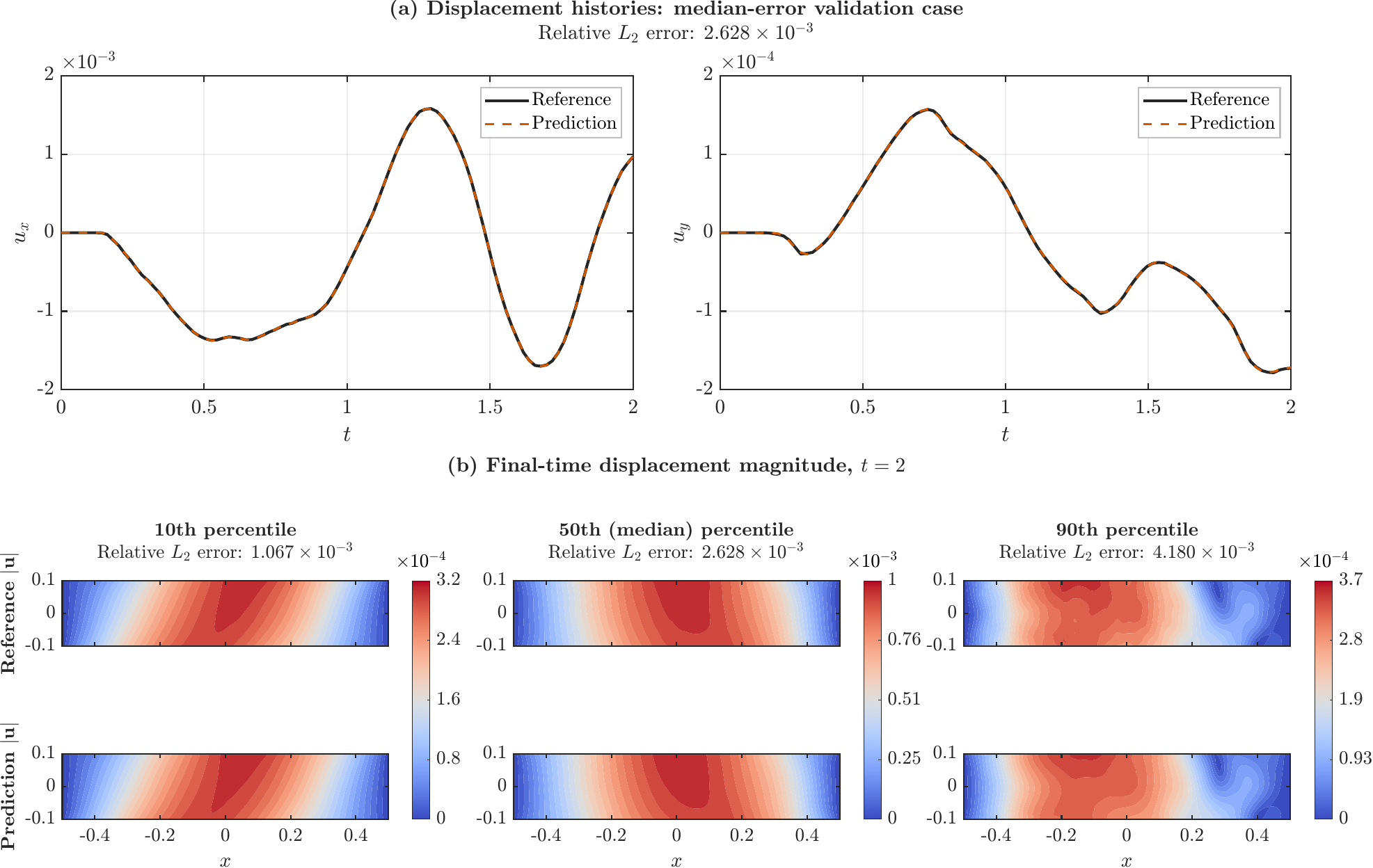}
\caption{\textbf{Transient elastodynamics: field behavior of the audit-enabled champion.} (a) Reference (solid) and predicted (dashed) displacement histories $\ux(t)$ and $u_y(t)$ at the interior node nearest the strip center, for the median-error validation case (relative $L^2$ \num{2.628e-3}). (b) Magnitude of the final-time displacement field $|\mathbf{u}|$ over the strip, reference (top row) against prediction (bottom row), for the validation cases at the $10$th, $50$th (median), and $90$th percentiles of the per-sample relative $L^2$ error ($n=\num{500}$); each column header gives the selected case's error, and the two panels of a column share one color scale, with the power-of-ten factor above the bar. \Cref{tab:elastodynamics-aggregate} reports the corresponding aggregate statistics and physics checks.}
\label{fig:elastodynamics-fields}
\end{figure}

\paragraph{Accuracy.} On the validation set, the audit-enabled champion attains a \rltwo{} of \num{2.629e-3}, about $3.5$\% lower than the audit-disabled champion's \num{2.724e-3}. The difference is larger at higher error percentiles: the audit-enabled champion's $95$th-percentile and worst-case errors are about $10$\% and $42$\% lower (\cref{tab:elastodynamics-aggregate}). Its worst-case error is about $2.1$ times the mean (\cref{fig:elastodynamics-dist} (a)), compared with about $3.5$ times for the audit-disabled champion (\cref{tab:elastodynamics-aggregate}). Because the reference field and this champion's prediction are both linear in the prescribed edge motion, a small motion gives essentially the same relative error as a larger motion of the same shape. For the audit-disabled champion, which is not linear by construction, this property is not guaranteed (\cref{tab:elastodynamics-contracts}). For the median-error case, the predicted and reference displacement histories at the node nearest the strip center agree closely, and both build up from rest (\cref{fig:elastodynamics-fields} (a)). The prescribed edge displacements are enforced by construction, so the differences lie inside the strip, most visibly in the finer spatial structure of the $90$th-percentile case (\cref{fig:elastodynamics-fields} (b)).

\begin{table}[htbp]
\centering
\caption{\textbf{Transient elastodynamics: accuracy and reference audit.} Validation displacement-field error of the audit-enabled and audit-disabled champions on the full validation set ($n=\num{500}$), and their post-run reference audit without retraining. The reference-audit status is determined by the sampled \violmax{} (at most $1$ pass); the Adversary \violmax{} is reported separately and does not affect this status. The audit-disabled champion fails through causality, a hard contract in the reference contract file. Entries printed as $0.00$ are below table precision; the audit-enabled champion's largest values are \num{4.5e-6} sampled and \num{6.7e-5} in the Adversary report.}
\label{tab:elastodynamics-aggregate}
\small
\begin{tabular}{lcccccc}
\toprule
Champion & \rltwo{} & val.\ p95 & val.\ max & sampled \violmax{} & Adversary \violmax{} & status \\
\midrule
Audit-enabled & \num{2.629e-3} & \num{4.42e-3} & \num{5.48e-3} & 0.00 & 0.00 & pass \\
Audit-disabled    & \num{2.724e-3} & \num{4.90e-3} & \num{9.51e-3} & \textbf{12.62} & \textbf{2750.62} & \textbf{fail} \\
\bottomrule
\end{tabular}
\end{table}

\paragraph{Candidate search.} The audit-enabled run generated \num{16} candidates. Two reached the training-time limit and produced no score, which shows that the compute budget remained an active constraint for this transient example (\cref{sec:execution-safeguards}); \cref{sec:conclusion} discusses this limitation. Of the \num{14} candidates scored, \num{12} passed the in-run sampled hard-contract gate. The two candidates that failed the in-run gate had scores of \num{2.83e-2} and \num{6.25e-3}; their largest sampled violations, \num{12.5} and \num{4.7}, came from the zero-input check. In physical terms, each predicted a nonzero displacement for a body that starts from rest and receives no boundary motion. Both also failed load superposition, amplitude scaling, and sign symmetry. A linear model must return zero response for zero input, a property that the load-independent output offset of the failed candidate in the first example also violates. The gate did not change the final selection: the lowest-score candidate in the audit-enabled search also passed. Among the gate-passing candidates, scores vary substantially while sampled violations remain more than four orders of magnitude below the gate threshold (\cref{fig:elastodynamics-dist} (b)).

\paragraph{Physics verification.} Under the reference contract file, the audit-disabled champion fails the causality check, consistent with its bidirectional branch (\cref{tab:elastodynamics-aggregate}). Across the sampled causality probes, altering only the later boundary motion changes its predicted displacement field up to the cut time by as much as $63$\% in relative $L^2$ terms, more than twelve times the $5$\% tolerance; the Adversary (\cref{fig:adversary}) finds changes exceeding $10^4$\%. Causality is the only hard check that the audit-disabled champion fails in the sampled audit; its sampled normalized violations on the other five hard checks are at most $0.29$ (\cref{tab:elastodynamics-contracts}). The Adversary, however, reports normalized values above $1$ on two of these checks, load superposition (\num{2.79}) and the prescribed right edge (\num{1.99}), without affecting the sampled audit's pass/fail status. Because none of these relations is built into this champion, its nonzero violations show how closely training reproduced them.

Dependence on future boundary motion makes the audit-disabled champion unsuitable for forward prediction when that motion is unknown, as in real-time estimation or feedback control. By contrast, the audit-enabled champion is causal by construction, and its violations are exactly zero except on load superposition and amplitude scaling, where they are negligible (at most \num{6.7e-5}). The post-run reference audit therefore separates two surrogates with similar mean validation error by a mechanics requirement that the predictive metric does not directly test.

\begin{table}[htbp]
\centering
\caption{\textbf{Transient elastodynamics: per-check violations.} Worst normalized violation for each check under the reference contract file (sampled audit / Adversary report) on the full validation set; a champion passes when every hard check has a sampled normalized violation at most $1$. The first six checks are hard; sign symmetry and the amplitude bound (daggered) are soft. Causality and the prescribed right edge are hard here but were soft in the in-run contract file. Zeros are exact. n.a.\ marks a check not included in the Adversary report.}
\label{tab:elastodynamics-contracts}
\small
\begin{tabular}{lcc}
\toprule
Requirement & audit-enabled champion & audit-disabled champion \\
\midrule
load superposition          & \num{4.5e-6} / \num{6.7e-5} & 0.25 / 2.79 \\
amplitude scaling           & \num{2.2e-6} / 0 & 0.09 / 0.55 \\
zero response to zero input & 0 / 0 & 0.02 / 0.03 \\
zero initial displacement   & 0 / 0 & 0.07 / 0.08 \\
causality & 0 / 0 & \textbf{12.62} / \textbf{2750.62} \\
prescribed right edge       & 0 / 0 & 0.29 / 1.99 \\
sign symmetry\textsuperscript{$\dagger$}               & 0 / n.a. & 0.15 / n.a. \\
amplitude bound\textsuperscript{$\dagger$}             & 0 / n.a. & 0 / n.a. \\
\bottomrule
\end{tabular}

\vspace{2pt} {\footnotesize \textsuperscript{$\dagger$}Soft diagnostic (does not gate).}
\end{table}

\paragraph{Post-discovery attribution.} The attribution tests (\cref{sec:attribution-method-card-saving}) examined the three improving steps on the path to the audit-enabled champion (\cref{tab:elastodynamics-attribution}). Unlike in the first example, the dominant edit at every step exceeded the attribution threshold and was saved as a method card. The first edit, which replaced the model with a linear-residual operator network whose output is linear in the load history, reproduces $129$\% of its step's score gain on its own. As in the first example, the largest share belongs to an architectural change that enforces physical structure. The second edit, which introduced the factorized space-time representation along with the causal load branch, reproduces $57$\% of its step's gain. The final edit, which combined separate temporal bases for the two displacement components with wider networks and additional spatial Fourier harmonics, reproduces $55$\% of its step's gain. All three edits bundle multiple components, so these tests do not separate the components' individual contributions to predictive accuracy, such as those of the lifting and the envelope; causality itself follows from the model construction, independently of the measured gain.

\begin{table}[htbp]
\centering
\caption{\textbf{Transient elastodynamics: saved method cards from post-discovery attribution.} The reported share of gain is the dominant single-edit share, the largest \(\phi_i\) among successfully scored variants of the corresponding step (\cref{eq:single-edit-share}), expressed as a percentage. A value above $100$\% means that the isolated variant scores better than the full child. Each listed edit was saved as a method card.}
\label{tab:elastodynamics-attribution}
\small
\begin{tabular}{llc}
\toprule
Improvement step & dominant single edit & share of gain \\
\midrule
first improvement & linear-residual operator-network model & 129\% \\
second improvement & factorized space-time representation with causal load branch & 57\% \\
final refinement & separate temporal bases, wider networks, more spatial Fourier harmonics & 55\% \\
\bottomrule
\end{tabular}
\end{table}

\paragraph{What this example shows.} Under the stated checks and tolerances, the answer to the first opening question is yes and to the second is no, and the example supports three conclusions. First, the two champions have almost the same mean validation error, but only the audit-enabled champion is causal. Second, the audit-enabled champion passes the stated hard checks and has lower $95$th-percentile and worst-case validation errors. Third, the reported pass/fail outcome depends on the contract file: causality is hard in the reference contract file but was soft in the in-run file (\cref{sec:verification}); \cref{sec:conclusion} discusses what this implies for reviewing proposed contracts. The two remaining examples increase the complexity of the material description, culminating in history-dependent plasticity.

\section{Progressive cantilever elasticity with increasing material complexity}
\label{sec:example-cantilever-progression}

The next static study keeps the geometry, loading, discretization, and output field fixed, while only increasing the material description provided to the surrogate. The body is a two-dimensional cantilever plate occupying $\Omega=[0,4]\times[0,1]$, discretized by $576$ four-node quadrilateral elements with \num{637} nodes. The left edge $x=0$ is fully fixed; the right edge $x=4$ carries a uniform downward traction $T_y$; the top and bottom edges are traction-free; and there is no body force. With $\mathbf{u}=(\ux,u_y)$, the three levels solve
\begin{align}
\nabla\!\cdot\boldsymbol{\sigma} &= \mathbf{0} && \text{in }\Omega,\\
\mathbf{u} &= \mathbf{0} && \text{on }x=0,\\
\boldsymbol{\sigma}\mathbf{n} &= (0,T_y) && \text{on }x=4,\\
\boldsymbol{\sigma}\mathbf{n} &= \mathbf{0} && \text{on }y=0,1,
\end{align}
under small-strain plane stress. The surrogate predicts both displacement components at all mesh nodes. The loading range is approximately $T_y\in[-1.5\times10^5,-2.5\times10^4]$~Pa in every level.

\begin{figure}[!htbp]
\centering
\includegraphics[width=0.96\linewidth]{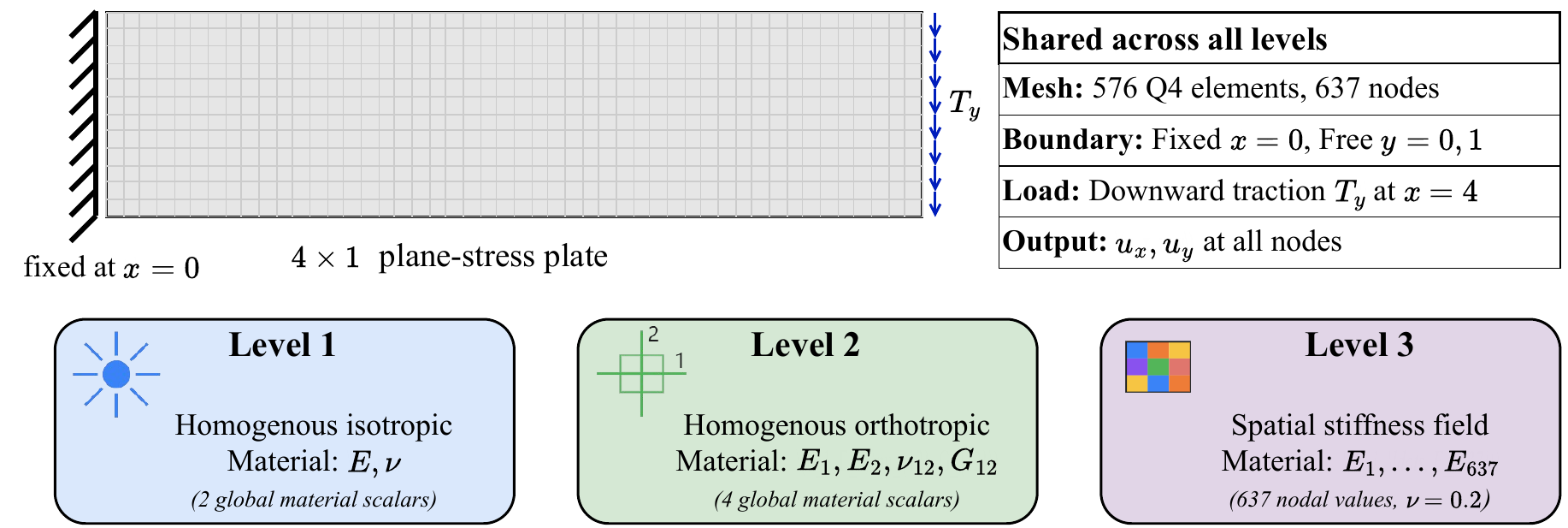}
\caption{\textbf{Progressive cantilever-plate benchmark.} All three levels use the same $4\times1$ plane-stress plate, \num{637}-node Q4 mesh, fully fixed left edge, downward right-edge traction, and traction-free horizontal edges. The material input increases from two homogeneous isotropic constants in Level~1, to four homogeneous orthotropic constants in Level~2, to the complete \num{637}-value Young's modulus field in Level~3. The surrogate predicts $(\ux,u_y)$ at every node.}
\label{fig:cantilever-material-levels}
\end{figure}

The progression changes the constitutive input. In Level~1, the material is homogeneous and isotropic, with one Young's modulus $E$ and Poisson's ratio $\nu$ per sample,
\begin{equation}
\boldsymbol{\sigma}=\mathbb{C}_{\mathrm{iso}}(E,\nu):\boldsymbol{\varepsilon}(\mathbf{u}),
\end{equation}
where the empirical dataset ranges are approximately $E\in[2.51\times10^{10},2.09\times10^{11}]$~Pa and $\nu\in[0.180,0.337]$. In Level~2, the material is homogeneous orthotropic with axes aligned with the global $x$--$y$ directions. The engineering constants are $E_1$, $E_2$, $\nu_{12}$, and $G_{12}$, with $\nu_{21}=\nu_{12}E_2/E_1$ and
\begin{equation}
\mathbb{C}_{\mathrm{ortho}}=
\begin{bmatrix}
E_1/d & \nu_{12}E_2/d & 0\\
\nu_{12}E_2/d & E_2/d & 0\\
0 & 0 & G_{12}
\end{bmatrix},
\qquad d=1-\nu_{12}\nu_{21}>0.
\label{eq:cantilever-orthotropic-C}
\end{equation}
The corresponding empirical ranges are $E_1\in[8.12\times10^9,3.49\times10^{10}]$~Pa, $E_2\in[4.01\times10^8,1.46\times10^{10}]$~Pa, $\nu_{12}\in[0.220,0.400]$, and $G_{12}\in[4.20\times10^8,5.89\times10^9]$~Pa. In Level~3 the material returns to isotropy but becomes spatially heterogeneous: each sample carries a complete nodal field $E(\mathbf{x})$, represented by \num{637} Young's modulus values in the same ordering as the mesh, with $E(\mathbf{x})\in[2.0\times10^{10},4.5\times10^{10}]$~Pa and fixed $\nu=0.20$. The three input--output maps are therefore
\begin{align}
(T_y,E,\nu) &\mapsto \mathbf{u}(\mathbf{x}) && \text{Level 1},\\
(T_y,E_1,E_2,\nu_{12},G_{12}) &\mapsto \mathbf{u}(\mathbf{x}) && \text{Level 2},\\
(T_y,E_1,\ldots,E_{637}) &\mapsto \mathbf{u}(\mathbf{x}) && \text{Level 3}.
\end{align}
The geometry and output dimension are unchanged, so the sequence isolates the increasing difficulty of representing homogeneous isotropy, orthotropy, and a spatial material field (\cref{fig:cantilever-material-levels}).

\paragraph{Reference data, split, and fixed evaluator.}
Each level contains \num{3000} finite element displacement fields, with \num{2400} used for training and \num{600} for validation. Level~1 uses a disjoint, shuffled, sample-wise split. For the material-state experiments in Levels~2 and~3, all traction cases belonging to one material state are kept in the same partition so that validation measures generalization to unseen constitutive states rather than merely to new traction magnitudes applied to a state already present in training. This distinction is particularly important in Level~3, where the \num{100} spatial material states each contain \num{30} traction cases and the default $80/20$ state split gives \num{2400} training and \num{600} validation samples. The common lower-is-better evaluator is the mean per-sample relative $L^2$ error of the full two-component displacement field,
\begin{equation}
\rltwo
=\frac{1}{N_{\mathrm{val}}}\sum_{i=1}^{N_{\mathrm{val}}}
\frac{\lVert\widehat{\mathbf{U}}_i-\mathbf{U}_i\rVert_F}
{\lVert\mathbf{U}_i\rVert_F},
\qquad \mathbf{U}_i\in\mathbb{R}^{637\times2}.
\label{eq:cantilever-rel-l2}
\end{equation}
The same evaluator is used for the audit-disabled and audit-enabled searches within each level. Median and maximum errors are reported to expose the clusters of high-error cases that appear once validation is grouped by material state.

\paragraph{Three search conditions.}
As in the other examples, the primary comparison is between the audit-disabled and audit-enabled searches (\cref{sec:design}). The audit-disabled search can still discover mechanics-compatible structure, such as analytical prefactors, boundary masks, or POD representations, as in the parametric linear elasticity example (\cref{sec:example-elasticity}).

Levels~2 and~3 add a second control that tests knowledge reuse independently of physics auditing. The \emph{knowledge-informed audit-disabled search} makes the learned knowledge base accumulated from preceding levels available to later proposals. The \emph{knowledge-free audit-disabled search} starts the same level without that previously learned knowledge. Both are still selected only by predictive error; this comparison therefore asks whether accumulated design knowledge changes what the agent discovers, rather than whether a physics gate changes selection. In Levels~2 and~3, the primary comparison uses the knowledge-informed audit-disabled search; the knowledge-free search provides the additional comparison of knowledge reuse.

\paragraph{Physics checks.}
The hard checks follow from the governing linear static problem but are
specialized to each material representation. Level~1 uses four hard
contracts: traction linearity, reciprocal homogeneous-stiffness scaling,
and the two fixed-edge displacement conditions, $u_x=u_y=0$ at $x=0$.
In particular, homogeneous stiffness scaling requires
$\mathbf{u}(T_y,cE,\nu)=c^{-1}\mathbf{u}(T_y,E,\nu)$.

Level~2 uses seven hard contracts. Traction dependence is checked through
two separate tests, traction linearity and traction homogeneity, together
with the two fixed-edge displacement conditions. The homogeneous
orthotropic representation additionally admits three material-structure
checks: common scaling of $(E_1,E_2,G_{12})$, joint load--stiffness
invariance, and the mid-height condition $\ux(x,0.5)=0$, which follows from
the homogeneous aligned orthotropic material, symmetric geometry, and loading.

Level~3 uses six hard contracts. As in Level~2, traction linearity and
traction homogeneity are tested separately, together with the two fixed-edge
displacement conditions. The homogeneous-material scaling statements are
replaced by common scaling of the complete Young's-modulus field,
$E(\mathbf{x})\mapsto cE(\mathbf{x})$, and by joint scaling of $T_y$ and
the complete field. No individual-node stiffness-scaling contract is used,
because rescaling one nodal modulus does not multiply the assembled global
stiffness matrix by a scalar. Likewise, the Level~2 mid-height condition is
not carried into Level~3 because a general heterogeneous field
$E(\mathbf{x})$ need not be symmetric about $y=0.5$.

\paragraph{Selected architectures.}
The primary champions use different branch--trunk representations as the material description becomes more complex (Appendix~\ref{app:arch-cantilever}). In Level~1, the audit-disabled champion is a conventional learned-trunk DeepONet. The branch sees only normalized $\nu$ and produces separate \num{160}-dimensional coefficient vectors for $\ux$ and $u_y$; a learned coordinate trunk with Fourier features provides the corresponding spatial bases. The final field is multiplied by the exact $T_y/E$ compliance prefactor and by an $x/4$ boundary factor. The audit-enabled search instead selects a POD-DeepONet in which four fixed POD modes are retained for each displacement component. A degree-five polynomial representation of normalized $\nu$, a linear path, and a small residual MLP predict eight coefficients in total. The spatial trunk is fixed.

In Level~2, the knowledge-informed audit-disabled search reuses the low-dimensional fixed-trunk idea and selects a compact POD-DeepONet. A shared branch maps nine engineered dimensionless material features to $12$ $\ux$ and $6$ $u_y$ POD coefficients. The exact scale $S=(E_1E_2G_{12})^{1/3}$ is removed analytically and $T_y/S$ is restored outside the learned branch. The audit-enabled champion retains fixed POD trunks but uses the invariant coordinates $[\log(E_2/E_1),\log(G_{12}/E_1),\nu_{12}]$ and a shared encoder followed by component-specific heads. Eight modes are retained for each component, and the $\ux$ basis is antisymmetrized so that the mid-height $\ux$ condition is exact.

Level~3 changes the problem qualitatively because the branch must now encode a field rather than a few scalars. The knowledge-informed audit-disabled champion uses the complete scale-free field $g=\log E-\operatorname{mean}(\log E)$ in a dense encoder with a skip projection, then predicts separate $40$-mode $\ux$ and $25$-mode $u_y$ POD coefficients. The geometric-mean scale $s=\exp[\operatorname{mean}(\log E)]$ is removed and $T_y/s$ is restored exactly after reconstruction. The audit-enabled champion instead processes every node with shared weights: a node encoder maps $(x/4,y,E/\bar E-1)$ to features whose weighted averages over five mechanically defined regions feed a head that predicts four coefficients of one fixed joint POD decoder. The exact $T_y/\bar E$ factor and a binary left-edge mask are applied outside the learned weights. This produces a smaller model with a four-mode reconstruction bottleneck.

\begin{figure}[!htbp]
\centering
\includegraphics[width=1\linewidth]{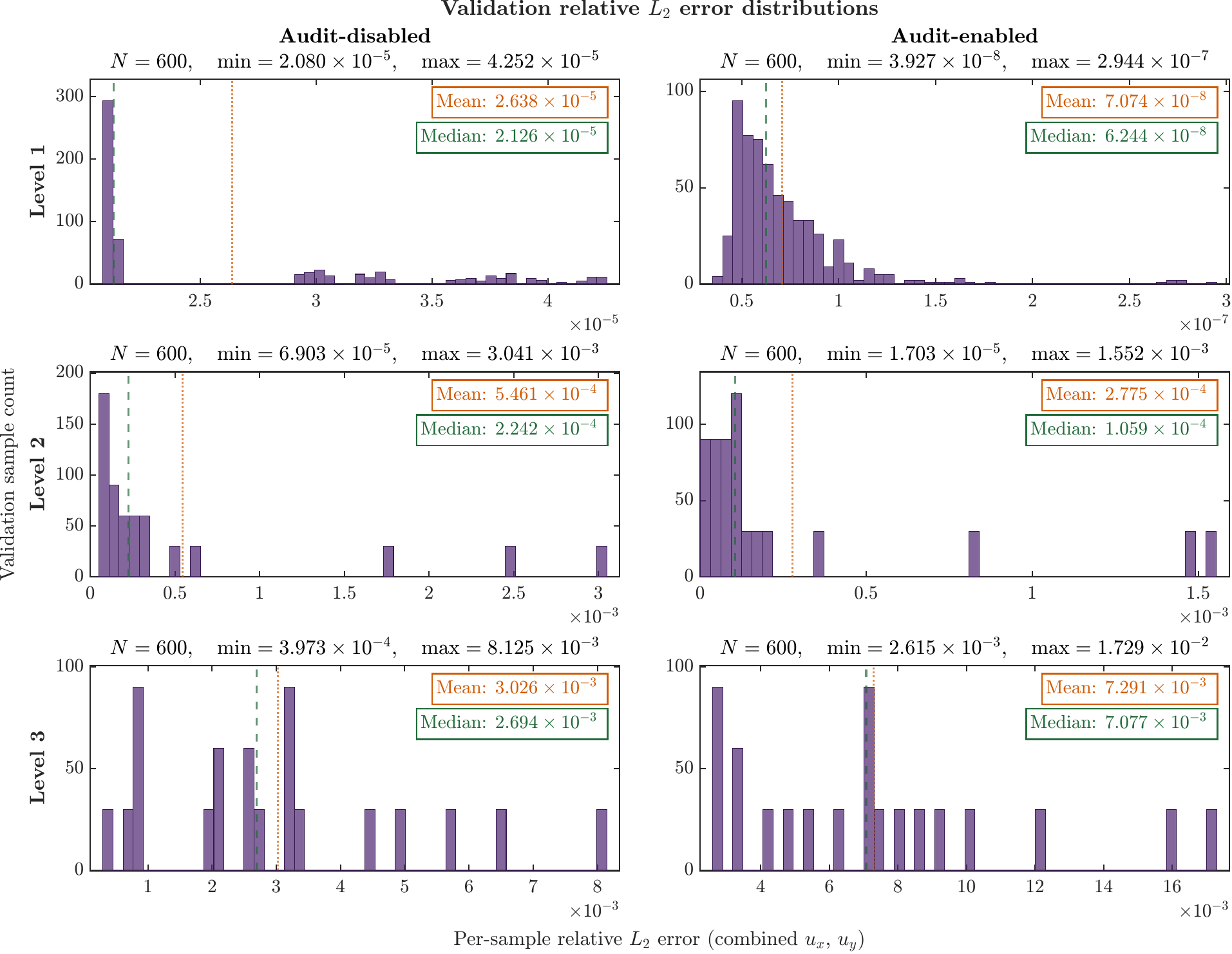}
\caption{\textbf{Validation-error distributions for the primary champions.} The left column is the audit-disabled champion and the right column is the audit-enabled champion for each level. The plots are the saved evaluator diagnostics for the \num{600} validation samples; dashed lines mark the mean and median. Level~1 benefits dramatically from the compact POD representation; Level~2 improves mainly for the material states with the largest errors; and Level~3 exposes the accuracy cost of the four-mode bottleneck of the audit-enabled champion.}
\label{fig:cantilever-error-distributions}
\end{figure}

\paragraph{Accuracy and model size.}
The resulting aggregate statistics are summarized in \cref{tab:cantilever-primary}, and the validation distributions are shown in \cref{fig:cantilever-error-distributions}. Level~1 gives the clearest case in which the audit-enabled search improves both accuracy and compactness: it reduces the mean error by approximately \num{373} times while producing a model that is about \num{512} times smaller. The fixed POD trunk and exact compliance factorization turn the problem into a small one-dimensional map from $\nu$ to displacement coefficients. At the same time, the audit-enabled champion passes all $4/4$ hard contracts, confirming traction linearity, reciprocal stiffness scaling, and the two fixed-edge displacement conditions.

Level~2 shows a different trade-off. The knowledge-informed audit-disabled champion reaches mean \num{5.461e-4}, while the audit-enabled champion reduces the mean to \num{2.775e-4}, corresponding to an improvement of approximately $50\%$. However, the more expressive split-head branch and symmetry-aware POD representation increase the parameter count from \num{10130} to \num{51088}. The validation histograms are strongly multimodal, with clusters containing multiples of \num{30} cases. This is consistent with the state-disjoint design: once the exact traction dependence has largely been factored out, errors organize primarily by material state rather than by individual load case. The hardest validation cases for the audit-enabled champion remain associated with extreme orthotropy, for which the retained eight-mode basis and coefficient map are least accurate. Despite these difficult cases, the audit-enabled champion passes all $7/7$ hard contracts, including the two traction checks, the fixed-edge conditions, the mid-height condition, common stiffness scaling, and joint load--stiffness invariance.

Level~3 provides the complementary result. The knowledge-informed audit-disabled champion is more accurate, with a mean of \num{3.026e-3}, whereas the audit-enabled champion is substantially smaller, but its mean error is \num{7.291e-3}, about \num{2.4} times that of the knowledge-informed audit-disabled champion. Its accuracy is limited by its representation: the four-mode joint POD decoder and regional weighted means discard fine spatial details needed for highly heterogeneous fields. Training also stopped at the wall-clock limit before completing the learning-rate schedule, so some of the accuracy gap may stem from incomplete training. Physics verification and representational capacity are distinct requirements: the audit-enabled Level~3 champion passes all hard checks, yet it is less accurate than both audit-disabled champions.

\begin{table}[htbp]
\centering
\caption{\textbf{Primary progressive-cantilever comparison.} Parameter counts are for the
active trained architecture after data-dependent POD truncation. For the audit-disabled searches, ``not audited'' describes the search workflow even when the exported network contains exact
analytical factors.}
\label{tab:cantilever-primary}
\small
\setlength{\tabcolsep}{4pt}
\begin{tabularx}{\linewidth}{c X r r r r c}
\toprule
Level & Search arm & Params. & Mean \rltwo{} & Median & Max & Sampled hard-contract gate \\
\midrule
1 & Audit-disabled
  & \num{2647680} & \num{2.638e-5} & \num{2.126e-5} & \num{4.252e-5}
  & not audited \\

1 & Audit-enabled
  & \num{5176} & \num{7.074e-8} & \num{6.244e-8} & \num{2.944e-7}
  & 4/4 pass \\

2 & Audit-disabled, with learned knowledge
  & \num{10130} & \num{5.461e-4} & \num{2.242e-4} & \num{3.041e-3}
  & not audited \\

2 & Audit-enabled
  & \num{51088} & \num{2.775e-4} & \num{1.059e-4} & \num{1.552e-3}
  & 7/7 pass \\

3 & Audit-disabled, with learned knowledge
  & \num{381825} & \num{3.026e-3} & \num{2.694e-3} & \num{8.125e-3}
  & not audited \\

3 & Audit-enabled
  & \num{167449} & \num{7.291e-3} & \num{7.077e-3} & \num{1.729e-2}
  & 6/6 pass \\
\bottomrule
\end{tabularx}
\end{table}

\paragraph{Effect of learned-knowledge reuse.}
The additional knowledge-free, audit-disabled searches provide an end-to-end ablation of the learned knowledge base, separate from the physics audit. In Level~2, removing prior learned knowledge changes the discovered model family itself. Instead of the \num{10130}-parameter fixed-trunk POD model, the independent run selects a component-separated DeepONet with four learned networks: separate material branches for $\ux$ and $u_y$ and separate learned coordinate trunks, resulting in an active architecture with \num{301440} trainable parameters. Making the previously learned knowledge available reduces the mean error by approximately $39\%$ while producing a model with nearly \num{30} times fewer trainable parameters. More importantly, the retrieved prior knowledge steers the later search toward a qualitatively different fixed-trunk POD representation rather than simply improving the same architecture.

\begin{figure}[htbp]
\centering
\includegraphics[width=1\linewidth]{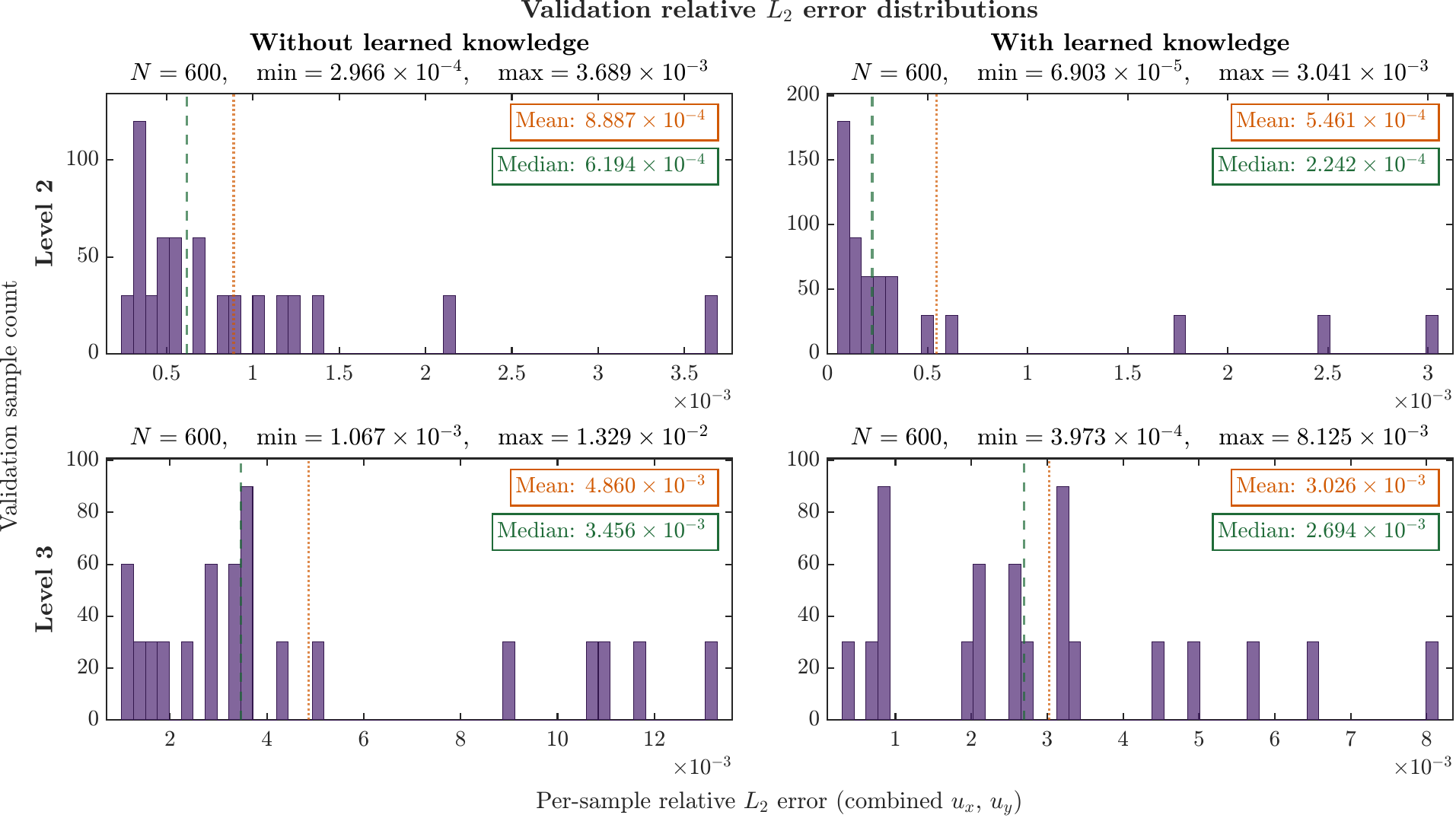}
\caption{\textbf{Learned-knowledge ablation in the audit-disabled search.} Validation-error distributions for Levels~2 and~3 when the level is started without previously learned knowledge (left) and when that knowledge is available to candidate proposals (right). In both levels, the knowledge-informed run lowers the mean error by approximately $38\%$. In Level~2, it also switches the selected architecture from a large, learned-trunk DeepONet to a compact, fixed-POD surrogate.}
\label{fig:cantilever-knowledge-transfer}
\end{figure}

The same test in Level~3 gives a similar reduction in mean error of approximately $38\%$. The knowledge-free champion is itself a physics-factored POD-DeepONet: shared scale-invariant per-node features are pooled globally and over $4\times2$ and $8\times4$ spatial bins, then mapped to $K=16$ joint POD coefficients. It uses exact $T_y/E_{\mathrm{ref}}$ factorization. Its active parameter count is \num{431440}. The knowledge-informed audit-disabled champion instead adopts the dense scale-free material encoder and the asymmetric component-specific POD representation described above, achieving lower error with \num{381825} parameters. Knowledge reuse therefore improves both later audit-disabled searches in these single runs, but it does so through different architectural changes in the two levels. This is evidence that the learned knowledge base changes the search trajectory; it is not an attribution of the improvement to any one stored method card. \Cref{tab:cantilever-knowledge-transfer,fig:cantilever-knowledge-transfer} summarize these results.

The Level~2 result is the clearest case: the knowledge-informed search recovers the fixed-trunk, factorized representation identified as reusable at Level~1 while adapting the coefficient map to orthotropy rather than copying the earlier branch unchanged. The two Level~3 audit-disabled champions also show that transferred knowledge is not simply architecture cloning. The transferable object is therefore better understood as search knowledge about useful mechanisms, structures, and representations, rather than as a fixed network template.

\begin{table}[htbp]
\centering
\caption{\textbf{Effect of learned-knowledge reuse within the audit-disabled search.} These comparisons do not use physics auditing for selection and therefore isolate knowledge availability at the workflow level.}
\label{tab:cantilever-knowledge-transfer}
\small
\begin{tabular}{llrrrr}
\toprule
Level & initial learned knowledge & trainable params. & mean \rltwo{} & median & max \\
\midrule
2 & unavailable & \num{301440} & \num{8.887e-4} & \num{6.194e-4} & \num{3.689e-3} \\
2 & available & \num{10130} & \num{5.461e-4} & \num{2.242e-4} & \num{3.041e-3} \\
3 & unavailable & \num{431440} & \num{4.860e-3} & \num{3.456e-3} & \num{1.329e-2} \\
3 & available & \num{381825} & \num{3.026e-3} & \num{2.694e-3} & \num{8.125e-3} \\
\bottomrule
\end{tabular}
\end{table}

\paragraph{Post-discovery attribution.}

Post-discovery attribution provides single-edit evidence for Levels~1 and~2, the only levels with method cards listed in \cref{tab:cantilever-method-cards}. In Level~1, replacing the learned spatial trunk with a fixed SVD/POD trunk with the exact physics prefactor recovers nearly the entire gain of its discovery step, and the polynomial-feature residual branch recovers more than the full gain of a later step. The extended AdamW schedule and component-normalized loss also survive isolation. In Level~2, the common-stiffness prefactor, together with dimensionless stiffness ratios, accounts for the full first improvement; an energy-weighted POD-coefficient loss yields a gain larger than the net step because simultaneous edits were harmful; and the run finds that a plain MLP branch outperforms the polynomial-residual form that was successful in Level~1. The higher initial learning rate is also retained.

\paragraph{What the progression shows.}
Taken together, the progression shows that mechanics structure can reduce both approximation error and model size. The appropriate learned representation changes with constitutive complexity, and the encoding of the spatial material field becomes central as heterogeneity is introduced. Physics verification and predictive accuracy remain distinct outcomes. A physics pass confirms the tested properties at the sampled inputs; it does not show that the surrogate has enough representational capacity for every difficult material state.

The knowledge-reuse comparisons also provide evidence that accumulated design knowledge can influence later discovery. These are single-run observations: repeated searches with independent seeds are required before interpreting the improvements as a stable effect.

Finally, the progression clarifies the boundary of the present linear-elastic contracts. The load and common-stiffness factorization used here is exact only because the governing problem is linear and the constitutive response is elastic. The nonlinear-material example that follows requires a different verification set: superposition and reciprocal stiffness scaling are no longer valid global requirements and must be replaced by checks appropriate to the chosen nonlinear constitutive model.

\begin{table}[htbp]
\centering
\caption{\textbf{Saved isolated-edit lessons for the progressive cantilever study.} The share is the fraction of the corresponding multi-edit step's score improvement recovered by the isolated edit; values above $100\%$ mean that other simultaneous edits were harmful.}
\label{tab:cantilever-method-cards}
\small
\begin{tabular}{llc}
\toprule
Level & saved isolated edit & share of step gain \\
\midrule
1 & component-normalized loss with gradient clipping & 64\% \\
1 & AdamW with extended training budget & 91\% \\
1 & fixed POD trunk with physics-prefactor factorization & 98\% \\
1 & polynomial-feature branch with residual MLP & 129\% \\
2 & common-stiffness prefactor and dimensionless ratio inputs & 100\% \\
2 & energy-weighted POD-mode auxiliary loss & 621\% \\
2 & plain MLP branch replacing polynomial-residual branch & 103\% \\
2 & higher initial learning rate & 107\% \\
\bottomrule
\end{tabular}
\end{table}

\section{Plate with a hole under cyclic loading: from perfect plasticity to combined hardening}
\label{sec:example-plate-with-a-hole}

\paragraph{Reference data, design space and metrics.}

\begin{figure}[!htbp]
\centering
\includegraphics[width=1\linewidth]{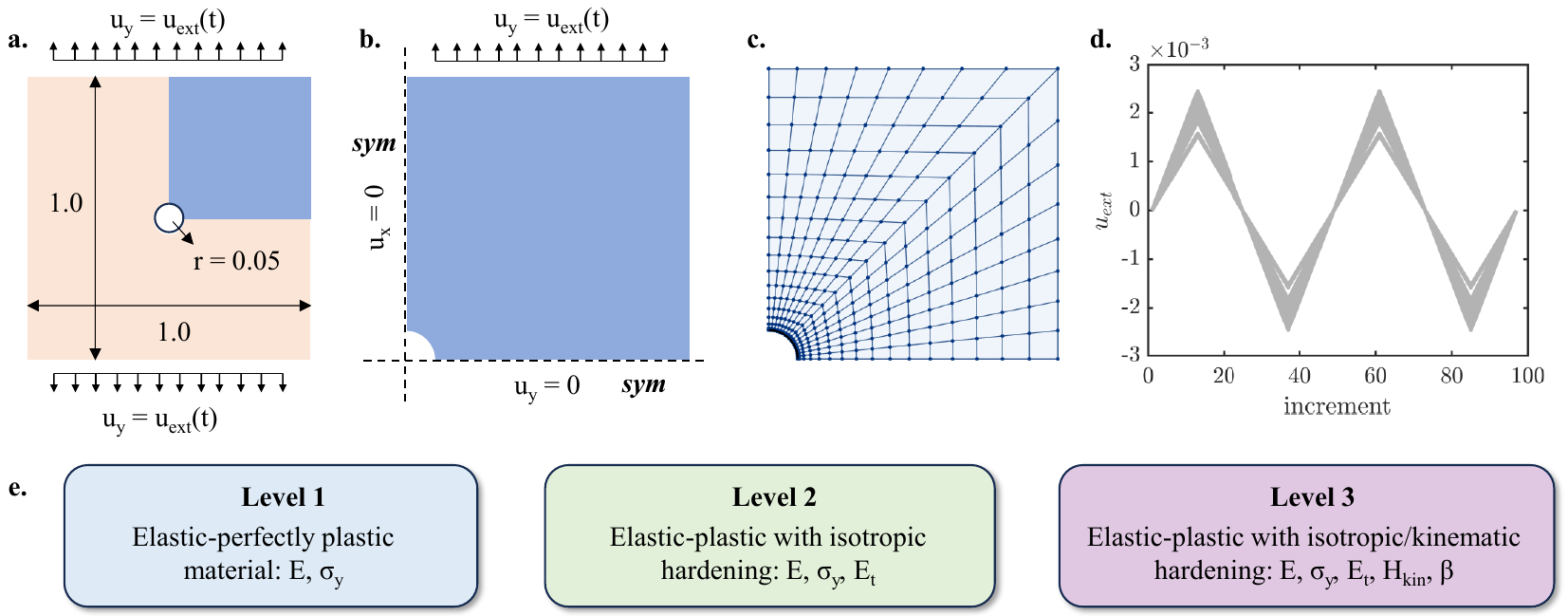}
\caption{\textbf{Plate with a hole: problem setup.} (a) Full domain, (b) symmetric top right quadrant, (c) finite element mesh, (d) externally applied displacement for the validation samples, and (e) description of the three levels. Units are in mm. }
\label{fig:Figure_DomainBCs}
\end{figure}

The final study focuses on discovering neural operators that capture nonlinear material behavior across increasing levels of constitutive complexity. The domain under investigation is a square plate occupying $\Omega = [0, 1] \times [0, 1]$~mm$^{2}$, with a central circular hole of radius $r = 0.05$~mm. A schematic of the domain is shown in \cref{fig:Figure_DomainBCs}a. The plate is subjected to cyclic loading via equal and opposite displacements applied on the top and bottom edges, while the vertical edges remain traction-free. Because of the problem's symmetry, we simulate only the top-right quadrant of the domain, using appropriate symmetry boundary conditions, as shown in \cref{fig:Figure_DomainBCs}b. In the absence of body forces, the problem is defined by the following governing equations: 

\begin{align}
\nabla\!\cdot\boldsymbol{\sigma} &= \mathbf{0}  && \text{in }\Omega,\\
\mathbf{u_{y}}                   &= \mathbf{u_{ext}(t)}  && \text{on }y=1.0,\\
\mathbf{u_{x}}                   &= \mathbf{0}  && \text{on }x=0,\\
\mathbf{u_{y}}                   &= \mathbf{0}  && \text{on }y=0.5,\\
\end{align}

The domain is discretized using an unstructured mesh of four-node quadrilateral elements, with refinement near the circular cutout to better capture stress concentrations. The resulting finite element model consists of 289 nodes, as shown in \cref{fig:Figure_DomainBCs}c.

The material of this study is taken as structural steel, for which three levels of material complexity are considered:

\begin{itemize}
    \item \textbf{Level~1 (elastic--perfectly plastic)}. Parameterized by two independent properties: Young's modulus $E$ and yield stress $\sigma_y$.
    \item \textbf{Level~2 (elastic with isotropic hardening)}. Parameterized by three properties: $E$, $\sigma_y$, and the tangent modulus $E_t$.
    \item \textbf{Level~3 (elastic with combined isotropic--kinematic hardening)}. Parameterized by five properties: $E$, $\sigma_y$, $E_t$, the kinematic hardening modulus $H_{\text{kin}}$, and the hardening ratio $\beta$.
\end{itemize}

For each sample, the material parameters are sampled independently from uniform distributions within the following bounds: $E \in[190000, 215000]$~MPa, $\sigma_{y} \in[350, 450]$~MPa, $E_{t} \in[1000, 3000]$~MPa, $H_{\text{kin}} \in[800, 1800]$~MPa, and $\beta \in[0.3, 0.7]$.

A dataset of 100 independent samples per level is generated, with an 80/20 split between training and validation sets. Each sample undergoes two complete loading cycles, where the applied displacement varies as $0 \to u_{\text{ext,max}} \to -u_{\text{ext,max}} \to 0$ twice, with peak amplitudes drawn uniformly from $|u_{\text{ext,max}}| \in [0.0015, 0.0025]$~mm. The temporal response is recorded across 97 load increments (48 per cycle plus the initial state). The sequence length is held constant across all samples. \Cref{fig:Figure_DomainBCs}d depicts the loading series for the validation samples.

Unlike the previous examples, where the output quantity was restricted to the 2D displacement field, the neural operators in this study are tasked with predicting three target fields: the displacements $u_{x}$ and $u_{y}$, and the nodal stress $\sigma$. Each sample's input representation includes both the material parameters and the history of the applied displacement. The learning task for each level is thus defined as:

\begin{align}
[u_{\text{ext}}(t), E, \sigma_{y}]  &\mapsto \mathbf{u}(\mathbf{x}, t), \boldsymbol{\sigma}(\mathbf{x}, t) && \text{Level~1} \\
[u_{\text{ext}}(t), E, \sigma_{y}, E_{t}]  &\mapsto \mathbf{u}(\mathbf{x}, t), \boldsymbol{\sigma}(\mathbf{x}, t) && \text{Level~2} \\ 
[u_{\text{ext}}(t), E, \sigma_{y}, E_{t}, H_{\text{kin}}, \beta]  &\mapsto \mathbf{u}(\mathbf{x}, t), \boldsymbol{\sigma}(\mathbf{x}, t) && \text{Level~3}
\end{align}

In this study, we limit both the audit-enabled and audit-disabled searches to 100,000 trainable parameters. This limit lets us examine the agentic framework's ability to discover operators that are both accurate and lightweight. The evaluator expression for this problem is the mean relative Frobenius norm error, averaged over all validation samples and across all three target fields. For each output variable $v \in \{u_{x}, u_{y}, \sigma_{\text{vM}}\}$, the relative error $\rltwo(v)$ is:

\begin{equation}
\rltwo(v) = \frac{1}{N_{\mathrm{val}}}\sum_{i=1}^{N_{\mathrm{val}}} \frac{\lVert{v_{\text{pred},i}} - {v_{\text{true},i}} \rVert_F}{\lVert {v_{\text{true},i}}\rVert_F + \varepsilon}
\end{equation}

\noindent where $\lVert\cdot\rVert_{F}$ is the matrix Frobenius norm, and $\varepsilon$ is a numerical stability constant set to $10^{-8}$. The final score is the unweighted average across all three variables:

\begin{equation}
    \mathrm{Score} = \frac{\rltwo(u_{x}) + \rltwo(u_{y}) + \rltwo(\sigma_{\text{vM}})}{3}
\end{equation}

The same score metric is utilized for both the audit-enabled and the audit-disabled searches. 

\paragraph{Physics checks.}

For the audit-enabled searches, the Mechanician drafted a contract file with the following checks; all are hard at every level unless noted:

\begin{itemize}
    
    \item \textbf{Symmetry condition on the left edge}. By symmetry about the vertical centerline of the plate, the predicted field must satisfy $u_{x}=0$ on $x=0$. 
    
    \item \textbf{Symmetry condition on the bottom edge}. By symmetry about the horizontal centerline, the predicted field must satisfy $u_{y}=0$ on $y=0.5$. 
    
    \item \textbf{Prescribed displacement on the top edge}. The predicted field must satisfy the loading condition $u_{y}(t)=u_{\text{ext}}(t)$ on $y=1.0$.
    
    \item \textbf{Zero-input response}. Provided with zero input, the operator must predict zero displacements and stresses. 
    
    \item \textbf{Causality}. Plastic response is history-dependent, and information from future load increments cannot leak to earlier ones. Therefore, if the operator is provided with two time series that are identical up to load increment $t_{\text{cut}}$ and different afterward, it must predict identical responses until $t_{\text{cut}}$.
    
    \item \textbf{Displacement bounds}. Both displacement components must satisfy $|u_{x}|, |u_{y}| \le 0.005$~mm, twice the largest applied displacement amplitude of 0.0025~mm.
    
    \item \textbf{Von Mises stress bounds}. By definition, the von Mises stress is a non-negative scalar quantity. The operator must always predict non-negative values for this component. Additionally, for Level~1, which corresponds to the elastic--perfectly plastic response, the maximum predicted von Mises stress must not exceed the maximum yield strength in the dataset ($\sigma_{\text{vM}} \le 450$~MPa).
    
    \item \textbf{Poisson's effect}. The right edge of the domain is traction-free, and due to Poisson's effect it must contract horizontally ($\Delta u_{x} \le 0$) under vertical stretching ($\Delta u_{\text{ext}} \ge 0$). Similarly, it must expand horizontally under vertical compression ($\Delta u_{x} \ge 0$ when $\Delta u_{\text{ext}} \le 0$).  
    
    \item \textbf{Displacement monotonicity}. Between consecutive load increments, the vertical displacement increment $\Delta u_{y} = u_{y,t} - u_{y,t-1}$ must have the same sign as the applied displacement increment $\Delta u_{\text{ext}} = u_{\text{ext},t} - u_{\text{ext},t-1}$.
    
    \item \textbf{Stiffness monotonicity}. Under identical imposed displacement loads and during the early linear elastic load increments, stiffer materials (higher Young's modulus \textit{E}) must produce smaller displacement magnitudes ($u_{x}$, $u_{y}$).
    
\end{itemize}

\paragraph{Results.}

\Cref{tab:pwh_physics_audit_results} summarizes the evaluation metrics for all champions, including per-field \rltwo{} errors, overall scores, number of trainable parameters, and physics-contract compliance. At Level~1, the audit-enabled and audit-disabled champions perform similarly from a purely numerical perspective, with overall scores of about $4.66\%$. At Level~2, however, the audit-enabled champion substantially outperforms the audit-disabled champion, achieving a score of approximately $4.57\%$, compared with $13.9\%$ for the audit-disabled champion. Notably, the Level~2 audit-disabled search arrived at a considerably smaller number of parameters (approximately 70,000).

In the more challenging Level~3, the agentic framework could not identify an acceptable root solution for the audit-disabled search within 8 attempts, and only by removing the parameter limit did the audit-disabled search arrive at a viable champion. However, this came at the cost of drastically increasing model size, surpassing 10,000,000 trainable parameters, two orders of magnitude above the initial limit. By contrast, the audit-enabled search found a champion within the original 100,000-parameter limit, achieving a score of approximately $25\%$. This investigation showed that imposing physics contracts can effectively guide the discovery of neural operators toward viable candidates without resorting to brute-force attempts, such as a blatant increase in trainable parameters.

All audit-enabled champions satisfied every imposed physics contract, whereas every audit-disabled champion violated at least one. Altogether, these results demonstrate that the audit-enabled workflow not only maintains competitive accuracy at lower difficulty levels but also substantially improves robustness and physical consistency under increasingly complex problems and stringent model-size constraints.

\begin{table}[htbp]
\centering
\caption{\textbf{Plate with a hole: accuracy, model size, and physics audit status for Levels~1--3.} Pass/fail labels for the audit-disabled champions were obtained from post-run reference audits after selection.}
\label{tab:pwh_physics_audit_results}
\renewcommand{\arraystretch}{1.2}
\small
\setlength{\tabcolsep}{4pt}
\begin{tabular}{ccccccl}
\toprule
Case & Parameters & $\rltwo(u_x)$ & $\rltwo(u_y)$ & $\rltwo(\sigma_{\text{vM}})$ & Score & Sampled hard-contract gate \\
\midrule
L1 audit-disabled & 99,653 & 0.041825 & 0.036128 & 0.061656 & 0.046536 & fail \\
L1 audit-enabled & 96,624 & 0.065874 & 0.036589 & 0.037531 & 0.046664 & pass (0.599) \\
L2 audit-disabled & 68,800 & 0.136944 & 0.094928 & 0.179913 & 0.138385 & fail \\
L2 audit-enabled & 95,907 & 0.049814 & 0.032334 & 0.054931 & 0.045676 & pass (0.118) \\
L3 audit-enabled & 93,362     & 0.223455 & 0.184535 & 0.339896 & 0.249295 & pass (0.903) \\
L3 audit-disabled (100k) & 122,155    & N/A      & N/A      & N/A      & N/A      & N/A \\
L3 audit-disabled (free) & 10,687,011 & 0.011204 & 0.009283 & 0.009445 & 0.009855 & fail \\
\bottomrule
\end{tabular}
\end{table}

\paragraph{Selected architectures.}

\begin{figure}[H]
\centering
\includegraphics[width=1\linewidth]{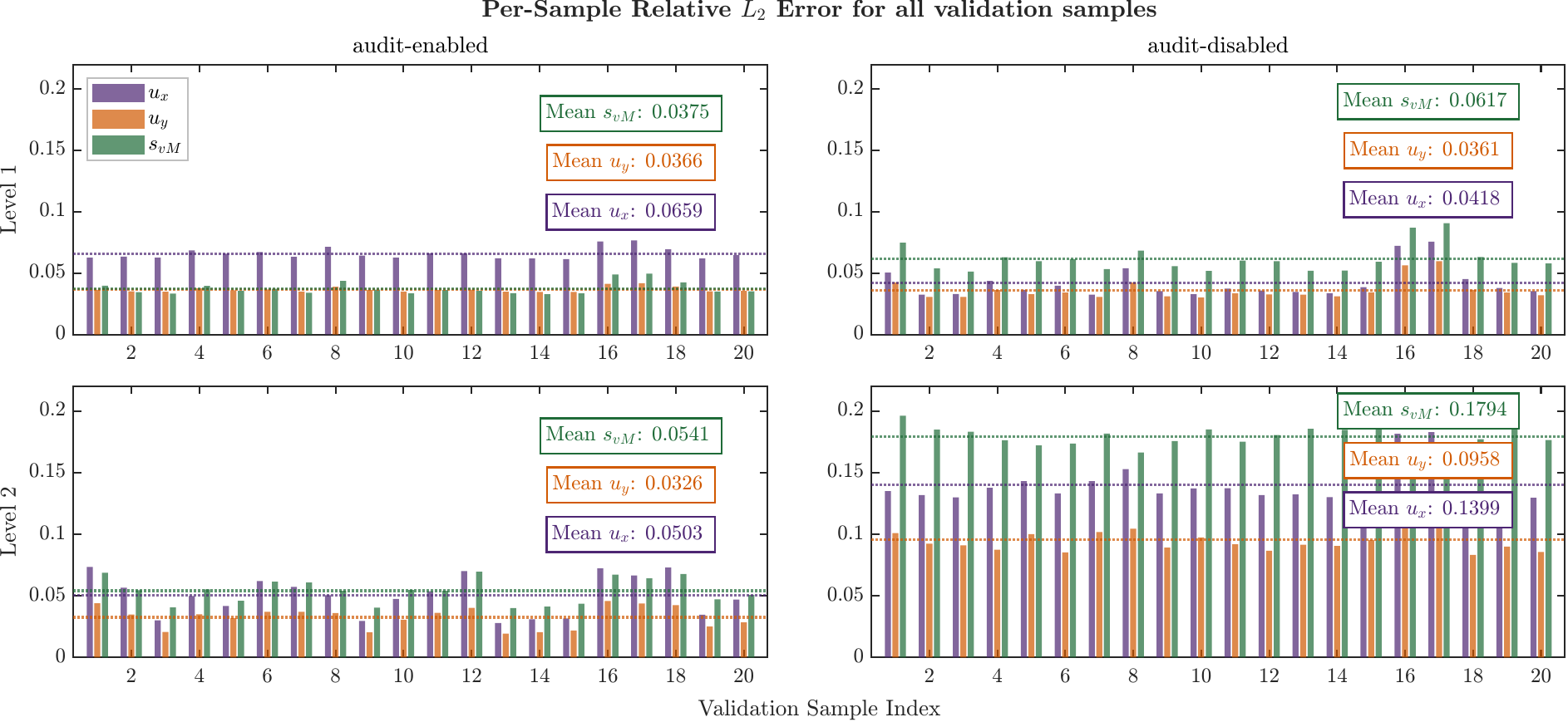}
\caption{\textbf{Plate with a hole: per-sample \rltwo{} error across all validation cases.} Comparisons are shown for displacement fields ($u_x$, $u_y$) and von Mises stress ($\sigma_{\text{vM}}$), across Level~1 and Level~2, for both the audit-enabled and the audit-disabled champions. The comparison for Level~3 is omitted because the audit-disabled search failed to find any candidate solution within the parameter limit.}
\label{fig:pwh_persampleL2}
\end{figure}

Appendix~\ref{app:arch-plate} shows the champion architectures for Levels~1--3. Several notable architectural trends directly influence the champions' contract compliance and are discussed in detail in the following paragraphs. The Level~2 audit-disabled champion uses a dense MLP architecture. This design choice lacks several specialized inductive biases that are required to optimize performance under a strict parameter limit. Consequently, this champion yields higher errors, as the detailed \rltwo{} error distribution in \cref{fig:pwh_persampleL2} shows. The plot shows the degraded performance of the L2 audit-disabled champion across all queried fields, and particularly for the displacement component ($u_{x}$) and the von Mises stress ($\sigma_{\text{vM}}$). At Level~3, once the parameter limit was removed for the audit-disabled search, its first candidate used more than 10 million trainable parameters. This observation highlights the importance of imposing external efficiency constraints, as the agents evidently did not prioritize them.

\paragraph{Field predictions.}

\begin{figure}[!htbp]
\centering
\includegraphics[width=1\linewidth]{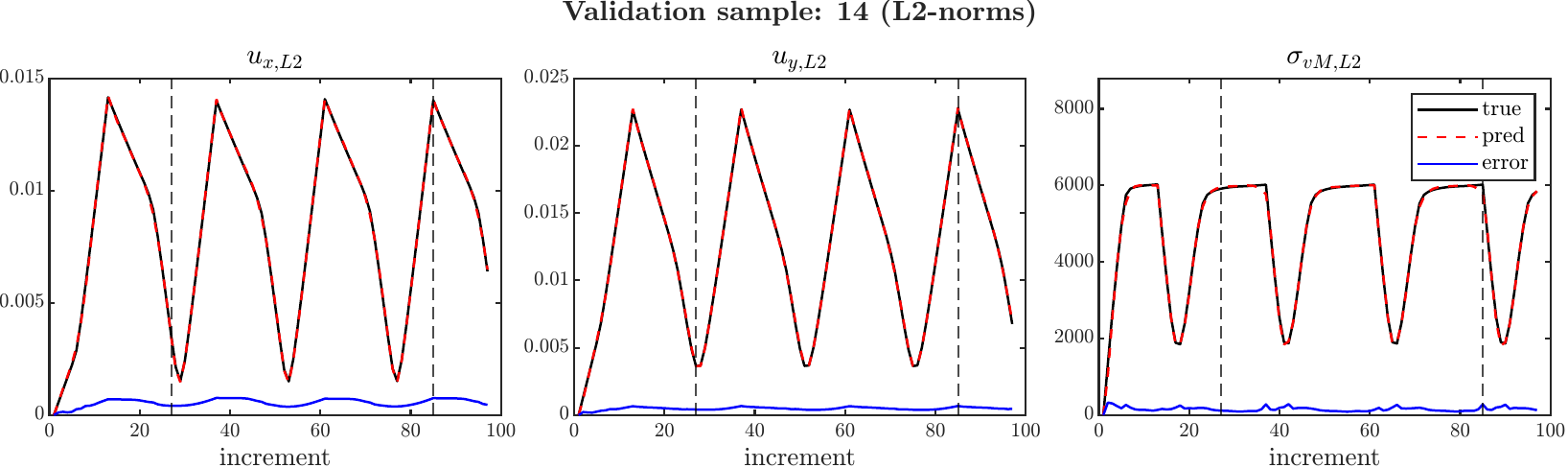}
\caption{\textbf{Plate with a hole: field norms for validation sample 14.} Evolution of $L^2$ norms for the displacement field ($u_x$, $u_y$) and von Mises stress ($\sigma_{\text{vM}}$) across all load increments. Comparisons are shown between the ground truth, predictions, and absolute error profiles using the Level~1 audit-enabled champion.}
\label{fig:pwh_valsample14_l1audit_100kparams_errors}
\end{figure}

\begin{figure}[!htbp]
\centering
\includegraphics[width=0.85\linewidth]{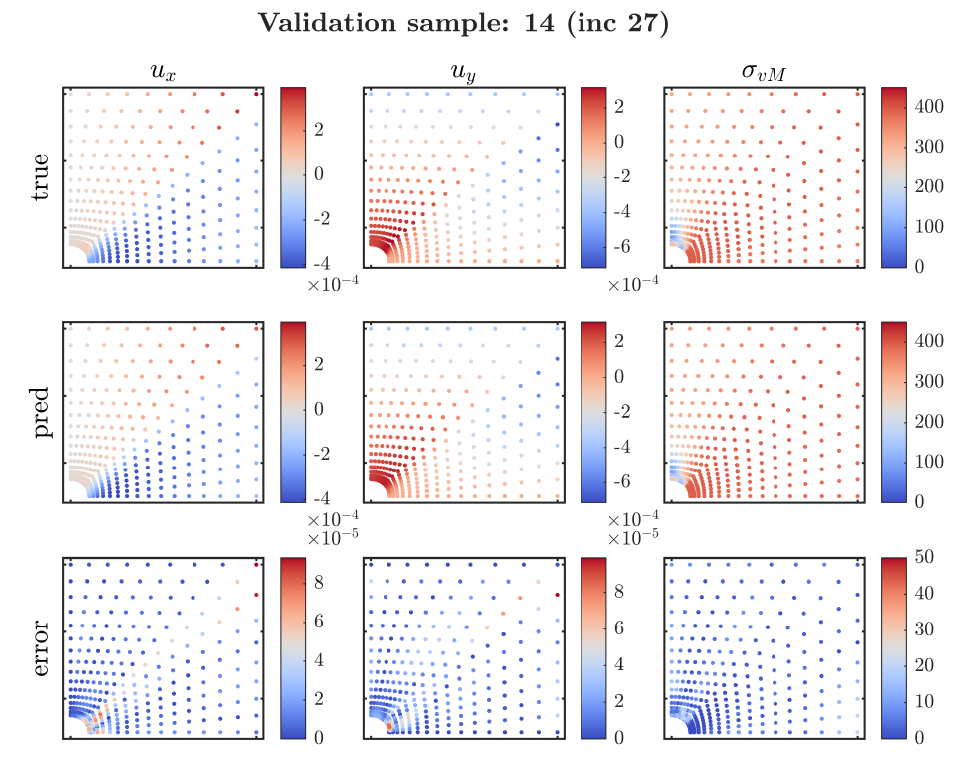}
\caption{\textbf{Plate with a hole: field contours for validation sample 14 ($27^{th}$ increment).} True (top), predicted (middle), and absolute error (bottom) maps for the displacement ($u_x$, $u_y$) and von Mises stress ($\sigma_{\text{vM}}$) fields, using the Level~1 audit-enabled champion.}
\label{fig:pwh_valsample14_l1audit_100kparams_contours_inc27}
\end{figure}

\begin{figure}[!htbp]
\centering
\includegraphics[width=0.85\linewidth]{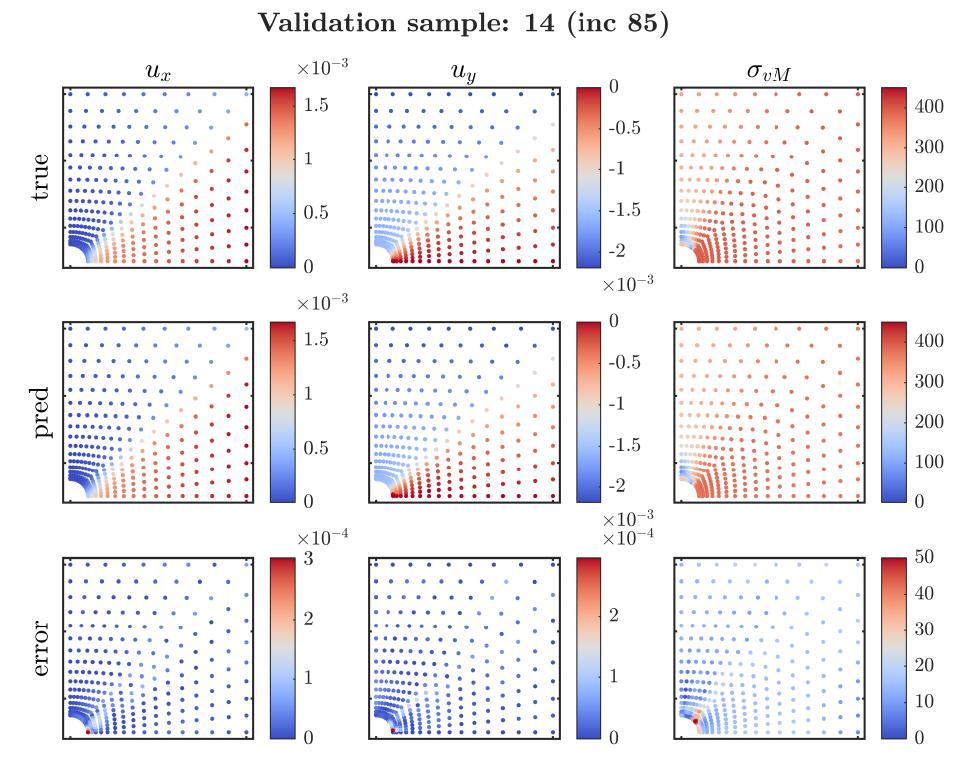}
\caption{\textbf{Plate with a hole: field contours for validation sample 14 ($85^{th}$ increment).} True (top), predicted (middle), and absolute error (bottom) maps for the displacement ($u_x$, $u_y$) and von Mises stress ($\sigma_{\text{vM}}$) fields, using the Level~1 audit-enabled champion.}
\label{fig:pwh_valsample14_l1audit_100kparams_contours_inc85}
\end{figure}

\Cref{fig:pwh_valsample14_l1audit_100kparams_errors} shows the evolution of the $L^2$ norms of $u_x$, $u_y$, and $\sigma_{\text{vM}}$ across all load increments for validation sample 14, using the Level~1 audit-enabled champion. The predicted fields closely follow the reference throughout the entire load history. The per-sample errors for all validation cases are given in \cref{fig:pwh_persampleL2}. For the same sample, \cref{fig:pwh_valsample14_l1audit_100kparams_contours_inc27,fig:pwh_valsample14_l1audit_100kparams_contours_inc85} show the fields at two representative load increments ($t=27$ and $t=85$ respectively).

\paragraph{Physics verification.}

\Cref{tab:pwh_physics_audit_results} demonstrates that the Level~1 and Level~2 audit-enabled champions strictly adhere to all physics contracts, achieving maximum violation magnitudes of $0.599 \le 1.0$ and $0.118 \le 1.0$, respectively. In contrast, the audit-disabled champions satisfy only a few checks by construction, such as the boundary conditions, and fail several others. A detailed analysis of contract compliance for each champion is provided below.

\begin{figure}[!htbp]
\centering
\includegraphics[width=0.95\linewidth]{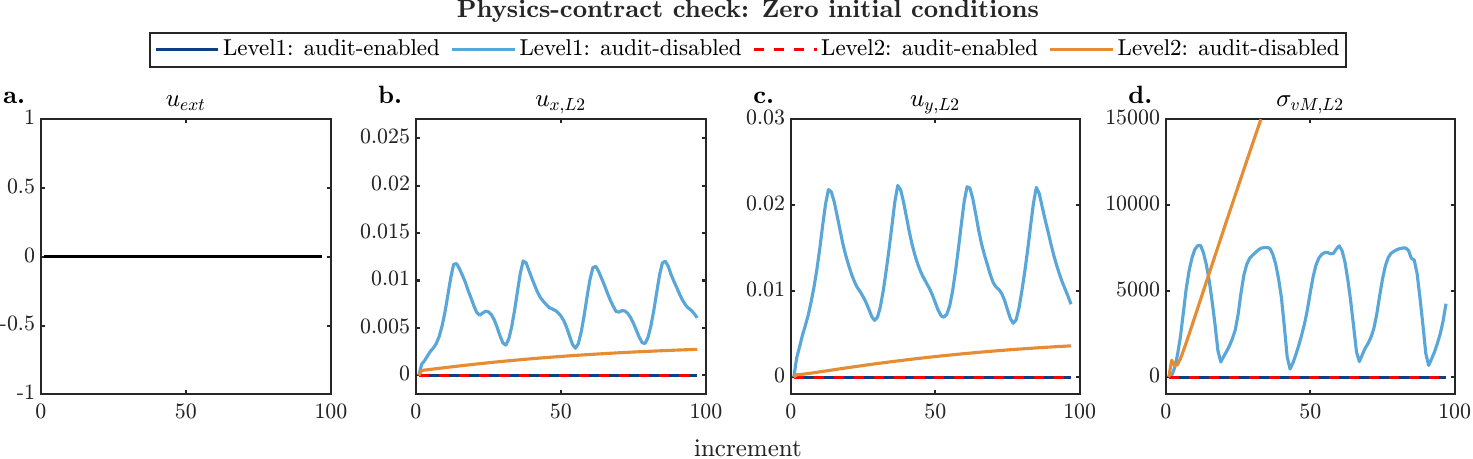}
\caption{\textbf{Plate with a hole: zero-input check.} Given zero applied displacement (a), the audit-enabled champion correctly predicts a zero response across all fields. In contrast, the audit-disabled champion generates a nonphysical, non-zero temporal evolution for all fields. The curves illustrate the $L^2$ norms for the three queried field variables: (b) $u_x$, (c) $u_y$, and (d) $\sigma_{\text{vM}}$.}
\label{fig:pwh_zeroconditions}
\end{figure}

\Cref{fig:pwh_zeroconditions} evaluates the champions' compliance with the zero-input contract. Because the plate starts stress-free, a physically admissible operator must predict a zero response in every field when it is provided with zero input ($u_{\text{ext}} = 0$). Both audit-enabled champions strictly satisfy this constraint, yielding zero displacement and stress fields throughout the time sequence. In contrast, the L1 audit-disabled champion predicts a non-physical, cyclic-like response, while the L2 audit-disabled champion also predicts non-zero field evolutions. Although the L2 audit-disabled champion's response is relatively smooth and tends toward zero, it remains non-zero, at least for the displacement field, and is thus physically inadmissible.

\begin{figure}[!htbp]
\centering
\includegraphics[width=1\linewidth]{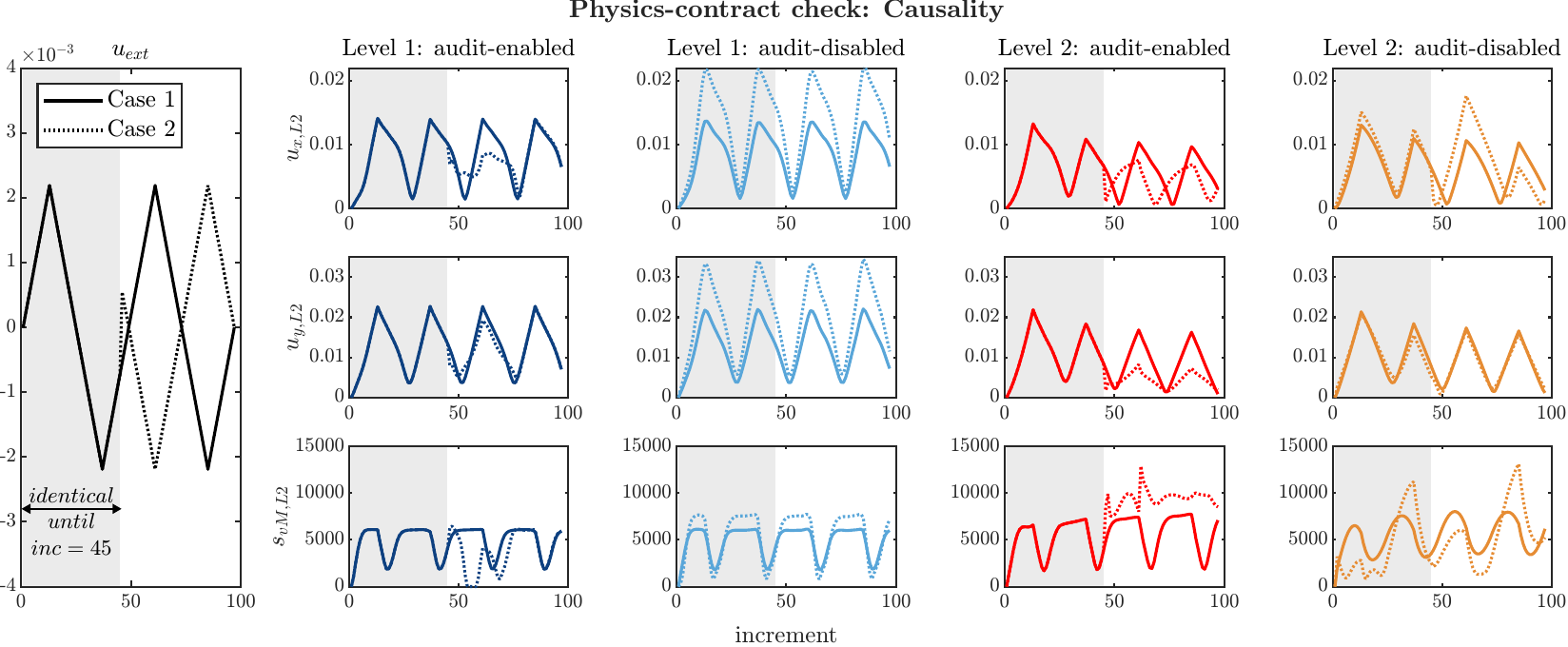}
\caption{\textbf{Plate with a hole: causality check.} Given two time series that are identical up to load increment $t_{\text{cut}}$ and diverge thereafter, both audit-enabled champions (second and fourth columns) respect causality by outputting identical predictions up to $t_{\text{cut}}$. In contrast, both audit-disabled champions (third and fifth columns) output divergent responses prior to $t_{\text{cut}}$ across all variables ($u_{x}$, $u_{y}$, and $\sigma_{\text{vM}}$), revealing a non-physical leakage of future information into past load increments.}
\label{fig:pwh_causality}
\end{figure}

\Cref{fig:pwh_causality} shows the two input time series and the $L^2$ norms for all fields for the Level~1 and~2 champions. The audit-enabled champions (columns 2 and 4) yield identical predictions for all variables until $t_{\text{cut}} = 45$. In contrast, the audit-disabled champions violate this fundamental property, with the responses before $t_{\text{cut}} = 45$ being clearly divergent. 

Further inspection of the network architectures reveals the structural reason behind this behavior. The L1 audit-enabled champion uses a 2-layer recurrent network (GRU) to process the load history, allowing only forward propagation in time. The L2 audit-enabled champion uses strictly causal mathematical prefix operations and a pointwise temporal MLP applied independently at each load increment. In both cases, the result is the strict enforcement of forward-only predictions. Such architectural features are not present in the audit-disabled champions. The L1 audit-disabled champion applies a 1D CNN but with adaptive average pooling, which leaks future load steps into the entire series, while the L2 audit-disabled champion simply implements a fully dense MLP that severely violates causality.   

\begin{figure}[!htbp]
\centering
\includegraphics[width=0.9\linewidth]{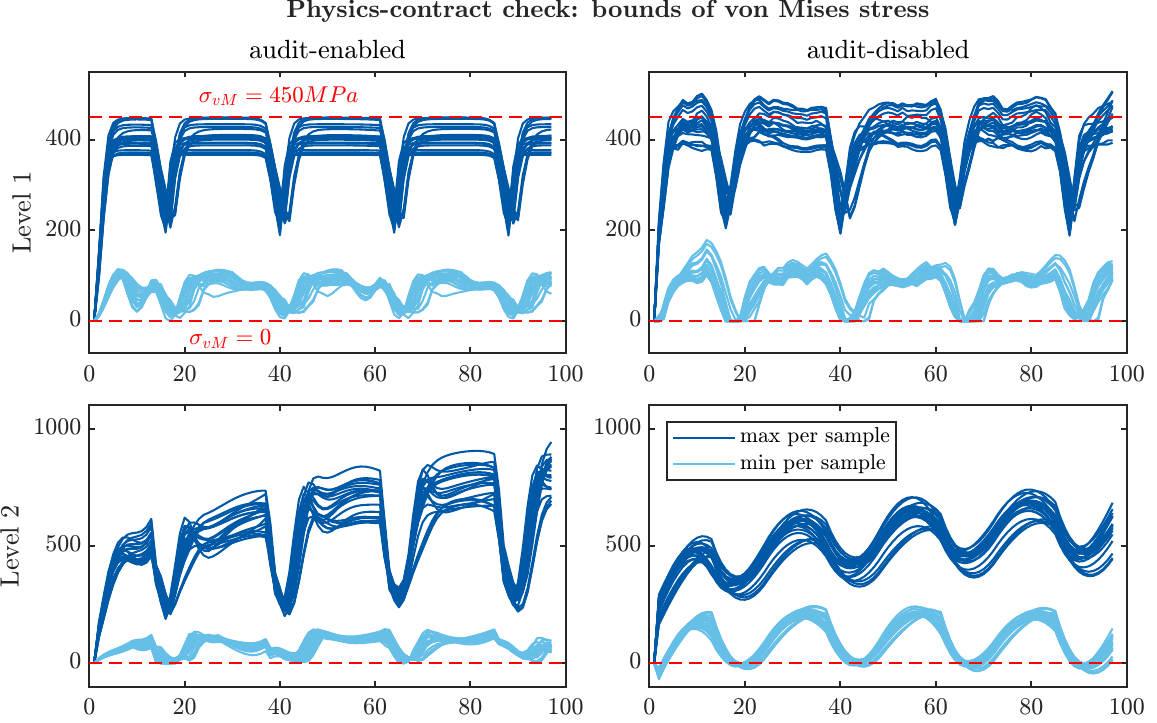}
\caption{\textbf{Plate with a hole: von Mises stress-bound check.} Evolution of the maximum (dark blue) and minimum (light blue) von Mises stress across all validation samples for both levels and approaches. At Level~1 (elastic--perfectly plastic), the audit-disabled champion violates the upper stress limit ($\sigma_{\text{vM}} > 450$~MPa). At Level~2, it violates the lower bound ($\sigma_{\text{vM}} < 0$~MPa). In contrast, the audit-enabled champions strictly enforce any applicable stress bounds at both levels.}
\label{fig:pwh_svMbounds}
\end{figure}

\Cref{fig:pwh_svMbounds} depicts the temporal evolution of the maximum (dark blue) and minimum (light blue) von Mises stress at Level~1 (top) and Level~2 (bottom) for the audit-enabled (left) and audit-disabled (right) champions. For elastic--perfectly plastic materials (Level~1), the von Mises stress must not exceed the yield strength. The audit-enabled champions satisfy the non-negativity requirement across both levels, and the Level~1 champion is strictly upper-bounded by the yield strength. This physical compliance stems from the architectural design choices. In the L1 audit-enabled champion (Appendix~\ref{app:arch-plate}), the predicted stress is expressed as the product of the yield strength $\sigma_{y}$ and a bounded activation term $\text{ReLU}(\tanh(\sigma))$, which enforces $\sigma_{\text{vM}} \in [0, \sigma_y]$. At Level~2, a \textit{softplus} activation function enforces non-negativity, without imposing an upper bound. In contrast, the L1 audit-disabled champion only applies a \textit{clamp.min($\sigma$,0)} operation, resulting in multiple upper-bound violations, while the L2 audit-disabled champion uses no bound constraints, yielding non-physical negative stress values.

\section{Conclusion}
\label{sec:conclusion}

This work demonstrates the value of assessing automatically discovered mechanics surrogates through both validation error and physics checks on their predicted fields. A low validation error does not, by itself, establish that a surrogate respects the relationships governing its response to changes in loads, material properties, or loading history. Across four examples of increasing complexity, from parametric linear elasticity to cyclic plasticity, every audit-enabled champion passed the required physics checks at sampled inputs, whereas audit-disabled champions violated causality, zero-input response, or stress bounds in the transient and plasticity examples. The audit-enabled champions also matched or exceeded the accuracy of the audit-disabled champions in six of eight comparisons. Physics auditing examines these relations directly and provides evidence that remains useful even when it does not change the final model selection. In \sysname{}, this evidence also informs later proposals, allowing discovery to respond to physics-check violations alongside validation error. The searches produced representations that enforce mechanical relations by construction, while single-edit tests showed how much of each score gain individual design changes reproduce, with qualifying changes saved as method cards for reuse. Satisfying the tested relations does not, by itself, make a surrogate accurate, as the heterogeneous cantilever shows; therefore, validation error and physics-check results must remain separately interpretable. These findings support physics auditing as an integral component of physically consistent agentic discovery.

The present results are limited by the coverage of the physics checks and the extent of the numerical study. Passing a finite set of checks does not establish that a surrogate satisfies the full governing problem throughout the admissible input domain. The primary comparisons use one run per configuration and assess complete workflows, so repeated searches and controlled studies of individual components are needed to establish the consistency of the observed differences and how physics-check results influence later proposals. Because the requirements encode judgment about mechanics, a further direction is to assess more systematically whether the chosen requirements, their tolerances, and their pass/fail roles represent the mechanics of the intended application, and to direct additional testing toward inputs where violations may have been missed. Extending the framework to three-dimensional, coupled mechanics problems would further test its applicability as solution fields and input spaces become more complex.

\section*{Acknowledgements}
This work was partially supported by the Sand Hazards and Opportunities for Resilience, Energy, and Sustainability (SHORES) Center, funded by Tamkeen under the NYUAD Research Institute Award CG013. The authors also acknowledge the NYUAD Center for Research Computing for providing resources, services, and skilled personnel.

\bibliographystyle{unsrtnat}
\bibliography{references}

\appendix
\section{Reproducibility}
\label{app:repro}

\subsection{Data availability}
\label{app:data-availability}
The code and data supporting this study will be made publicly available upon publication of the paper.

\subsection{Workflow roles and fixed routines}
\label{app:roles}
\Cref{tab:llm-roles,tab:fixed-components} summarize the LLM-mediated roles and fixed routines described in \cref{sec:framework}.

\begin{table}[H]
\centering
\small
\renewcommand{\arraystretch}{0.98}
\caption{\textbf{LLM-mediated roles in \sysname{}.}}
\label{tab:llm-roles}
\begin{tabular}{@{}>{\raggedright\arraybackslash}p{0.15\linewidth}
>{\raggedright\arraybackslash}p{0.30\linewidth}
>{\raggedright\arraybackslash}p{0.15\linewidth}
>{\raggedright\arraybackslash}p{0.30\linewidth}@{}}
\toprule
Role & Purpose & Role & Purpose \\
\midrule
Data analyst &
Summarizes the available data and problem context for setup. &
Proposer &
Suggests candidate model changes during candidate search. \\
\addlinespace[1.4pt]
Evaluator &
Drafts the predictive scoring routine that becomes the fixed evaluator for comparing candidates. &
Critic &
Reviews proposed changes before implementation. \\
\addlinespace[1.4pt]
Mechanician &
Selects supported physics-check types, marks each hard or soft, and proposes their parameters and tolerances from the supplied
problem statement and calibration results obtained during setup. &
Engineer &
Implements proposed candidates and repair edits. \\
\addlinespace[1.4pt]
Root engineer &
Builds the first candidate from the problem statement and data summary. &
Debugger &
Diagnoses code-validation failures to guide repairs within the retry limit. \\
\addlinespace[1.4pt]
Retriever &
Selects at most one saved method card for each selected parent. &
Selector &
When eligible parents exceed the per-iteration limit, combines LLM-ensemble votes (\cref{tab:lm-assignments}) with deterministic selection rules. The lowest-score candidate found so far is always included while it remains eligible. \\
\addlinespace[1.4pt]
Result analyst &
Summarizes candidate predictions, scores, and physics-check evidence for later proposals. &
Ablator &
Splits an improving parent--child change into isolated edits for retesting. \\
\addlinespace[1.4pt]
&
&
Distiller &
Drafts a method card after the required saving checks pass. \\
\bottomrule
\end{tabular}
\end{table}

\begin{table}[H]
\centering
\small
\renewcommand{\arraystretch}{0.92}
\caption{\textbf{Fixed routines in \sysname{}.}}
\label{tab:fixed-components}
\begin{tabular}{>{\raggedright\arraybackslash}p{0.30\linewidth}
>{\raggedright\arraybackslash}p{0.62\linewidth}}
\toprule
Routine & Purpose \\
\midrule
Fixed evaluator &
Evaluates candidate predictions using the same lower-is-better predictive metric within each run. \\
\addlinespace[1.4pt]
Calibration &
Runs predefined diagnostics on the reference training data before contract drafting and records evidence that
informs the Mechanician's contract selections and proposed tolerances. \\
\addlinespace[1.4pt]
Code validation and scoring routine &
Runs code validation and training, then scores the candidate with the fixed evaluator; records execution errors and timeouts,
and marks candidates for which no score is obtained as failed attempts. \\
\addlinespace[1.4pt]
Physics audit &
Applies the configured physics checks to predicted fields and records, for each audited candidate, the per-check
violations and the sampled \(\violmax{}\), the maximum tolerance-normalized violation over active hard checks. \\
\addlinespace[1.4pt]
Sampled hard-contract gate &
At final reporting, selects the lowest-score audited surrogate with sampled \(\violmax{}\le\gamma\); if none pass, it reports the
lowest-score surrogate among all scored candidates as unverified. \\
\addlinespace[1.4pt]
Adversary &
High-violation input search run after the physics audit when enabled; it keeps a trained surrogate
unchanged and varies admissible inputs to seek large physics-check violations. \\
\addlinespace[1.4pt]
Numericist &
Runs advisory numerical diagnostics during the proposal--critique exchange; these diagnostics inform proposals
only and do not enter the physics audit or the gate. \\
\addlinespace[1.4pt]
Method-card saving rule &
Applies the saving checks of \cref{sec:attribution-method-card-saving} before the Distiller drafts a card,
then saves the card with its supporting evidence. \\
\bottomrule
\end{tabular}
\end{table}

\subsection{Language-model assignments}
\label{app:lm-assignments}
For reproducibility, \cref{tab:lm-assignments} records the language models assigned to the LLM-mediated workflow roles in the reported audit-enabled runs. These assignments apply to proposal, critique, code generation and repair, setup of contract files, and post-run analysis. They do not apply to fixed routines: scoring by the fixed evaluator, the physics audit, the Adversary search, and the sampled hard-contract gate are executed by the workflow code.

\begin{table}[H]
\centering
\caption{\textbf{Language-model assignment used by the reported audit-enabled workflow.} Claude denotes Claude Sonnet 4.6; GPT denotes GPT-5.4. Full provider-specific model identifiers will be included with the released configuration files.}
\label{tab:lm-assignments}
\footnotesize
\setlength{\tabcolsep}{3pt}
\begin{tabular}{>{\raggedright\arraybackslash}p{0.58\linewidth}
>{\raggedright\arraybackslash}p{0.34\linewidth}}
\toprule
Workflow role group & Language model \\
\midrule
Problem analysis, fixed evaluator construction, retrieval, proposal, implementation, and result summary & Claude \\
Physics-check setup, post-discovery attribution, and method-card saving & Claude \\
Root engineer, Critic, and Debugger & GPT \\
Selector ensemble & One GPT and two Claude ensemble members \\
\bottomrule
\end{tabular}
\end{table}

\section{Physics-check types used in the numerical examples}
\label{app:contract-schemas}

This appendix lists the physics-check types used by the in-run and post-run reference contract files in the four numerical examples (\cref{tab:contract-schemas}). The list excludes check types supported by the code but unused in these examples, and it is not a list of universal mechanics assumptions. Which checks are hard or soft in each example is stated in the section for that example; \cref{sec:verification} describes how contract files are drafted and applied.

The physics audit constructs probe inputs in one of two ways. In \emph{parameter-vector mode} (the implementation's \texttt{box} mode), the surrogate input is a bounded vector of scalars, such as loads or material parameters, and probes change selected components within prescribed bounds. In \emph{function-valued input mode} (the implementation's \texttt{dataset} mode), the surrogate input is a whole function or time history, such as a boundary-loading history, and probes are drawn from a prescribed collection of inputs or formed as bounded combinations and transformations of those inputs. A probe is one audit input or a coordinated set of inputs, for example a base input and its scaled copy.

For each check \(c\), the physics audit computes a raw violation \(v_c\), which measures the discrepancy before division by the tolerance \(\tau_c\). The linearity check, for example, uses a relative \(L^2\) difference. The normalized violation is \(v_c/\tau_c\). Full definitions will be provided with the code release (\cref{app:data-availability}).

\begin{table}[H]
\centering
\footnotesize
\setlength{\tabcolsep}{3pt}
\caption{\textbf{Physics-check types used in the numerical examples.} Each row states what the physics audit does for the check. The last column lists the examples that use each check: E, parametric linear elasticity (\cref{sec:example-elasticity}); T, transient elastodynamics (\cref{sec:example-elastodynamics}); C, progressive cantilever (\cref{sec:example-cantilever-progression}); P, plate with a hole (\cref{sec:example-plate-with-a-hole}). Use of a check in these examples does not imply that the corresponding property is valid for every mechanics problem.}
\label{tab:contract-schemas}
\begin{tabular}{>{\raggedright\arraybackslash}p{0.19\linewidth}
>{\raggedright\arraybackslash}p{0.66\linewidth}
>{\raggedright\arraybackslash}p{0.08\linewidth}}
\toprule
\textbf{Check type} & \textbf{What the check does} & \textbf{Used in} \\
\midrule
\multicolumn{3}{l}{\textbf{\textit{Linearity, scaling, and symmetry}}} \\
\addlinespace[1pt]
\texttt{linearity} &
Uses bounded weights to combine selected parameter components or whole input histories, keeping other parameters fixed and all inputs admissible. Compares the resulting prediction with the same weighted sum of the separate predictions. &
E, T, C \\
\addlinespace[3pt]
\texttt{scaling} &
Multiplies specified parameter components or the entire input history by \(\lambda\) and compares the prediction with \(\lambda^{\alpha}\) times the base prediction. For parameter scaling, \(\alpha\) is the configured homogeneity exponent, for example \(\alpha=-1\) for reciprocal-modulus scaling; for input histories, the check uses \(\alpha=1\). &
E, T, C \\
\addlinespace[3pt]
\texttt{symmetry} &
Negates listed parameter coordinates or the whole input history and compares the prediction with the base prediction, either unchanged or with its sign flipped, as the contract specifies. &
E, T \\
\addlinespace[2pt]
\cmidrule{1-3}
\multicolumn{3}{l}{\textbf{\textit{Boundary and initial conditions}}} \\
\addlinespace[1pt]
\texttt{bc\_exactness} &
Compares the specified field component with its prescribed value at selected nodes, including boundary nodes and the cantilever's mid-height symmetry line. &
E, C, P \\
\addlinespace[3pt]
\begin{tabular}[t]{@{}l@{}}\texttt{prescribed\_bc\_}\\\texttt{consistency}\end{tabular} &
Compares the predicted displacement components at selected boundary nodes with the prescribed displacement history supplied to the surrogate. &
T, P \\
\addlinespace[3pt]
\texttt{ic\_exactness} &
Compares the predicted displacement at a specified time with its prescribed value; the transient example checks zero initial displacement. &
T \\
\addlinespace[2pt]
\cmidrule{1-3}
\multicolumn{3}{l}{\textbf{\textit{Output bounds and zero-input response}}} \\
\addlinespace[1pt]
\texttt{bound} &
Samples the configured inputs and checks that the predicted outputs stay inside the configured interval. &
E, T, P \\
\addlinespace[3pt]
\texttt{zero\_input} &
Applies a zero load or a zero input history and checks that the predicted response is zero. &
T, P \\
\addlinespace[2pt]
\cmidrule{1-3}
\multicolumn{3}{l}{\textbf{\textit{Causality}}} \\
\addlinespace[1pt]
\texttt{causality} &
Builds two input histories that agree up to a cut time and differ only afterward, and checks that the two predicted responses agree up to the cut time. &
T, P \\
\addlinespace[2pt]
\cmidrule{1-3}
\multicolumn{3}{l}{\textbf{\textit{Sign and monotonicity conditions}}} \\
\addlinespace[1pt]
Poisson's effect &
Checks that the horizontal displacement increment on the traction-free right edge is non-positive when the applied vertical displacement increases and non-negative when it decreases. &
P \\
\addlinespace[3pt]
Displacement monotonicity &
Checks that each displacement increment in the loading direction has the same sign as the applied displacement increment. &
P \\
\addlinespace[3pt]
Stiffness monotonicity &
Increases \(E\) with the other inputs fixed; over the early elastic increments, the displacement magnitude must decrease. &
P \\
\bottomrule
\end{tabular}
\end{table}

\clearpage
\section{Champion architectures}
\label{app:architectures}

This appendix collects the architecture schematics and forward-structure listings of the champions in \cref{sec:example-elasticity,sec:example-elastodynamics,sec:example-cantilever-progression,sec:example-plate-with-a-hole}. For the first two examples, each schematic shows both champions.

\subsection{Parametric linear elasticity}
\label{app:arch-elasticity}

\begin{figure}[p]
\centering
\includegraphics[width=\linewidth]{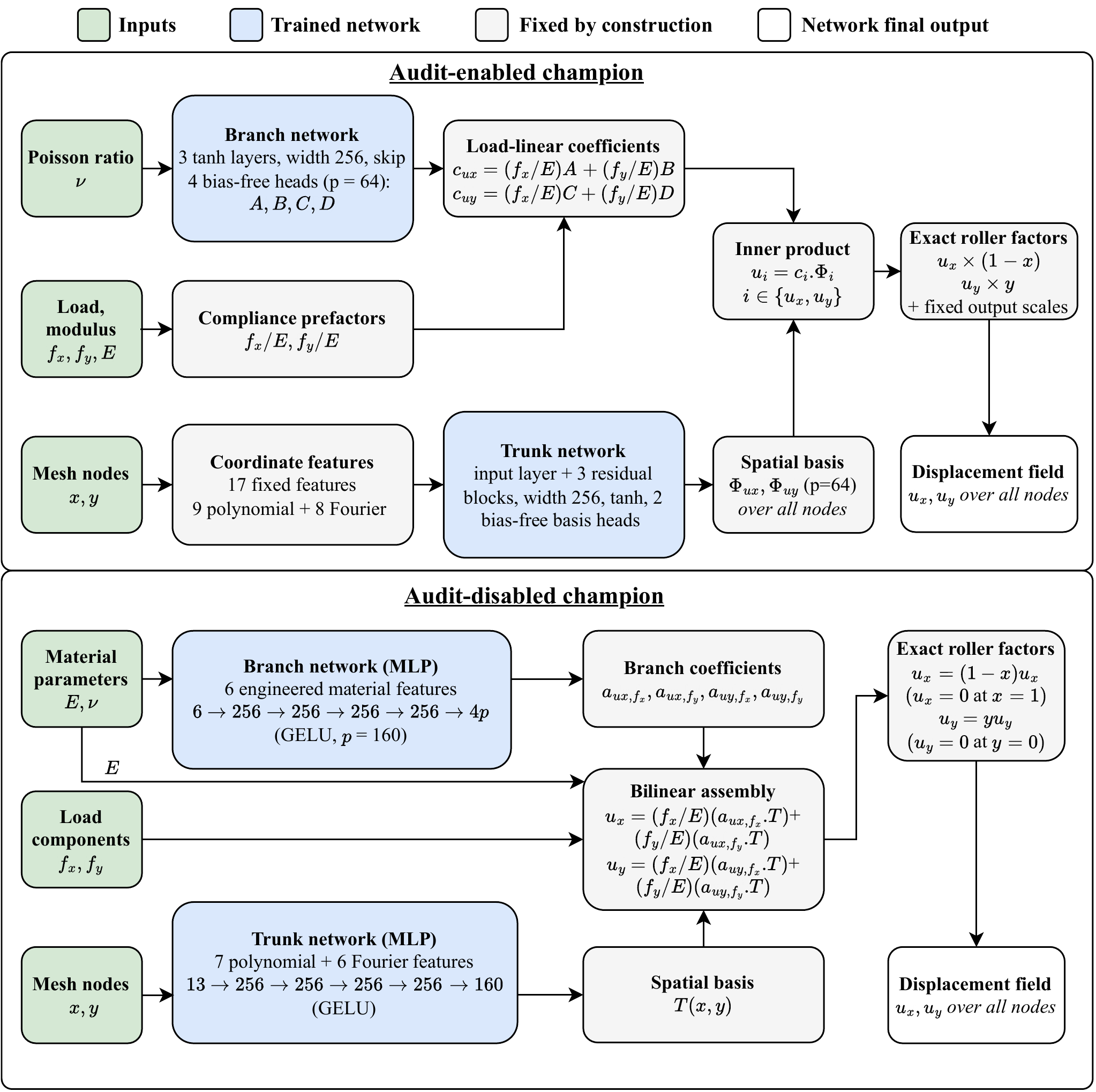}
\caption{\textbf{Champions for the parametric linear elasticity example: audit-enabled (top) and audit-disabled (bottom).} Both are DeepONets in which the fixed compliance prefactors $f_x/\Emod$ and $f_y/\Emod$ combine the learned branch coefficients with a learned spatial basis, so load superposition and load-sign reversal hold for any trained weights; the roller factors, $(1-x)$ on \ux{} and $y$ on $u_y$, enforce the two normal-displacement conditions identically. In the audit-enabled champion, the branch receives only Poisson's ratio, so reciprocal-\Emod{} scaling also holds by construction. Its trunk maps $17$ fixed coordinate features (nine polynomial terms and eight Fourier features \citep{tancik2020fourier,wang2021eigenvector}) through shared hidden layers and two output heads to separate bases of $64$ functions for \ux{} and $u_y$, a fixed per-component scale sets the output amplitude, and the assembled model has the form of \cref{eq:elasticity-superposition}, with about $6.3\times10^{5}$ trainable parameters. In the audit-disabled champion, the branch receives both \Emod{} and \poisson{}, so reciprocal-\Emod{} scaling is not enforced exactly. Its branch produces four coefficient vectors of length $160$, one for each traction--displacement pairing, and its trunk produces a single basis of $160$ functions that both components share before the roller factors are applied.}
\label{fig:elasticity-surrogate}
\end{figure}

\Cref{fig:elasticity-surrogate} shows the audit-enabled and audit-disabled champions of \cref{sec:example-elasticity}, and \cref{lst:elasticity-champion} gives the forward structure of the audit-enabled champion.

\begin{lstlisting}[style=surrogatepy,
caption={\textbf{Forward structure of the audit-enabled champion for the parametric linear elasticity example.} The listing shows the load and material factorization; input normalization, device handling, and output scaling are omitted.},
label={lst:elasticity-champion}]
fx, fy = params[:, 0:1], params[:, 1:2]
E,  nu = params[:, 2:3], params[:, 3:4]
fx_over_E, fy_over_E = fx / E, fy / E      # loads enter only as fx/E, fy/E
A, B, C, D = self.branch(nu_norm)          # branch input: Poisson ratio only
coeff_ux = fx_over_E * A + fy_over_E * B   # linear in (fx, fy) by construction
coeff_uy = fx_over_E * C + fy_over_E * D   # reverses sign when loads reverse
ux = einsum('bp,mp->bm', coeff_ux, trunk_ux_basis)  # DeepONet inner product
uy = einsum('bp,mp->bm', coeff_uy, trunk_uy_basis)
ux = ux * (1.0 - x)                        # right-edge roller: ux = 0 at x = 1
uy = uy * y                                # bottom-edge roller: uy = 0 at y = 0
\end{lstlisting}

\clearpage
\subsection{Transient elastodynamics}
\label{app:arch-elastodynamics}

\begin{figure}[p]
\centering
\includegraphics[width=\linewidth]{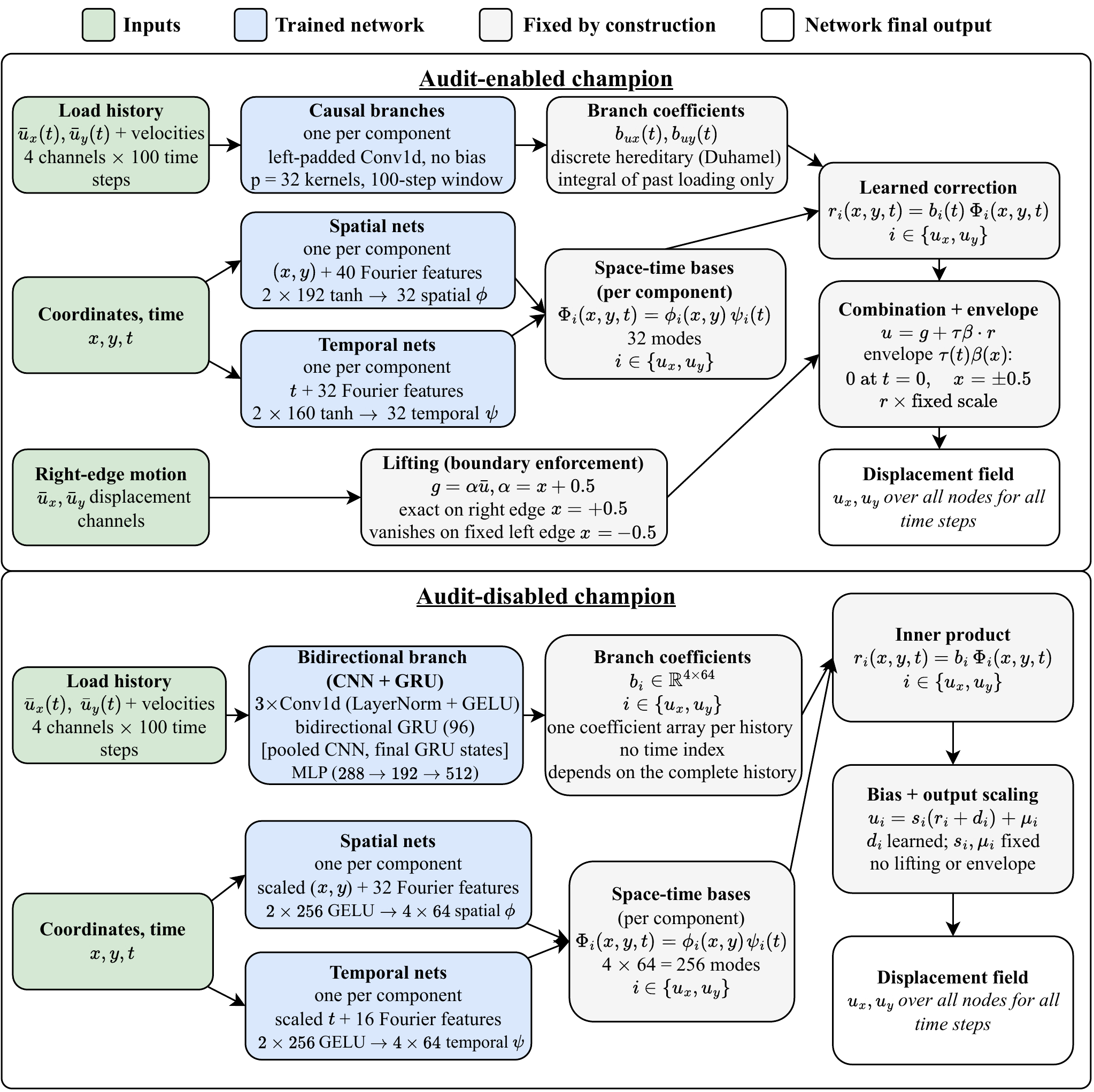}
\caption{\textbf{Champions for the transient elastodynamics example: audit-enabled (top) and audit-disabled (bottom).} In the audit-enabled champion, one causal branch per displacement component reads the right-edge motion history through left-padded, bias-free convolutions, so its coefficients depend only on present and past loading and act as discrete hereditary (Duhamel) integrals; a factorized space-time basis $\Phi_i(x,y,t)=\varphi_i(x,y)\,\psi_i(t)$ of $32$ functions gives the learned correction. A lifting $\mathbf{g}=(x+0.5)\,\bar{\mathbf{u}}$ carries the prescribed right-edge motion, and an envelope that vanishes at $t{=}0$ and at both ends gates the correction. Causality, load superposition, amplitude scaling, sign symmetry, zero response to zero input, zero initial displacement, and both displacement boundary conditions therefore hold for any trained weights; the model has about $2.0\times10^5$ trainable parameters. The audit-disabled champion combines convolutional features averaged over the complete history with the final forward and backward states of a bidirectional GRU, so its coefficients can depend on later boundary motion. Its trunk is also factorized, with separate spatial and temporal networks for each displacement component, giving $256$ basis functions per component. With GELU activations, an output bias, and no lifting or envelope, none of these relations holds by construction.}
\label{fig:elastodynamics-surrogate}
\end{figure}

\Cref{fig:elastodynamics-surrogate} shows the audit-enabled and audit-disabled champions of \cref{sec:example-elastodynamics}, and \cref{lst:elastodynamics-champion} gives the forward structure of the audit-enabled champion.

\begin{lstlisting}[style=surrogatepy,
caption={\textbf{Forward structure of the audit-enabled champion for the transient example.} The listing shows the causal branch and boundary
structure. Normalization and output scaling are omitted. Left padding makes the branch
causal; the lifting carries the prescribed right-edge motion and vanishes at the fixed left edge; the
envelope vanishes at $t=0$ and on the left and right edges.},
label={lst:elastodynamics-champion}]
x_pad = F.pad(load, (T - 1, 0))            # left-pad: only present and past load values in view
b_ux  = self.branch_ux(x_pad)              # causal Conv1d, no bias: linear and causal
trunk_ux = phi_ux * psi_ux                 # factorized spatial x temporal basis
r_ux  = einsum('btp,tnp->btn', b_ux, trunk_ux)  # learned correction field
alpha = x + 0.5                            # 0 at fixed left edge, 1 at moving right edge
g_ux  = alpha * u_bar_x                    # lifting: prescribed motion, exact at x = +0.5
envelope = tau * beta                      # zero at t = 0 and at x = -0.5, +0.5
ux = g_ux + envelope * r_ux
\end{lstlisting}

\clearpage
\subsection{Progressive cantilever}
\label{app:arch-cantilever}

\begin{figure}[H]
\centering
\includegraphics[width=0.96\linewidth]{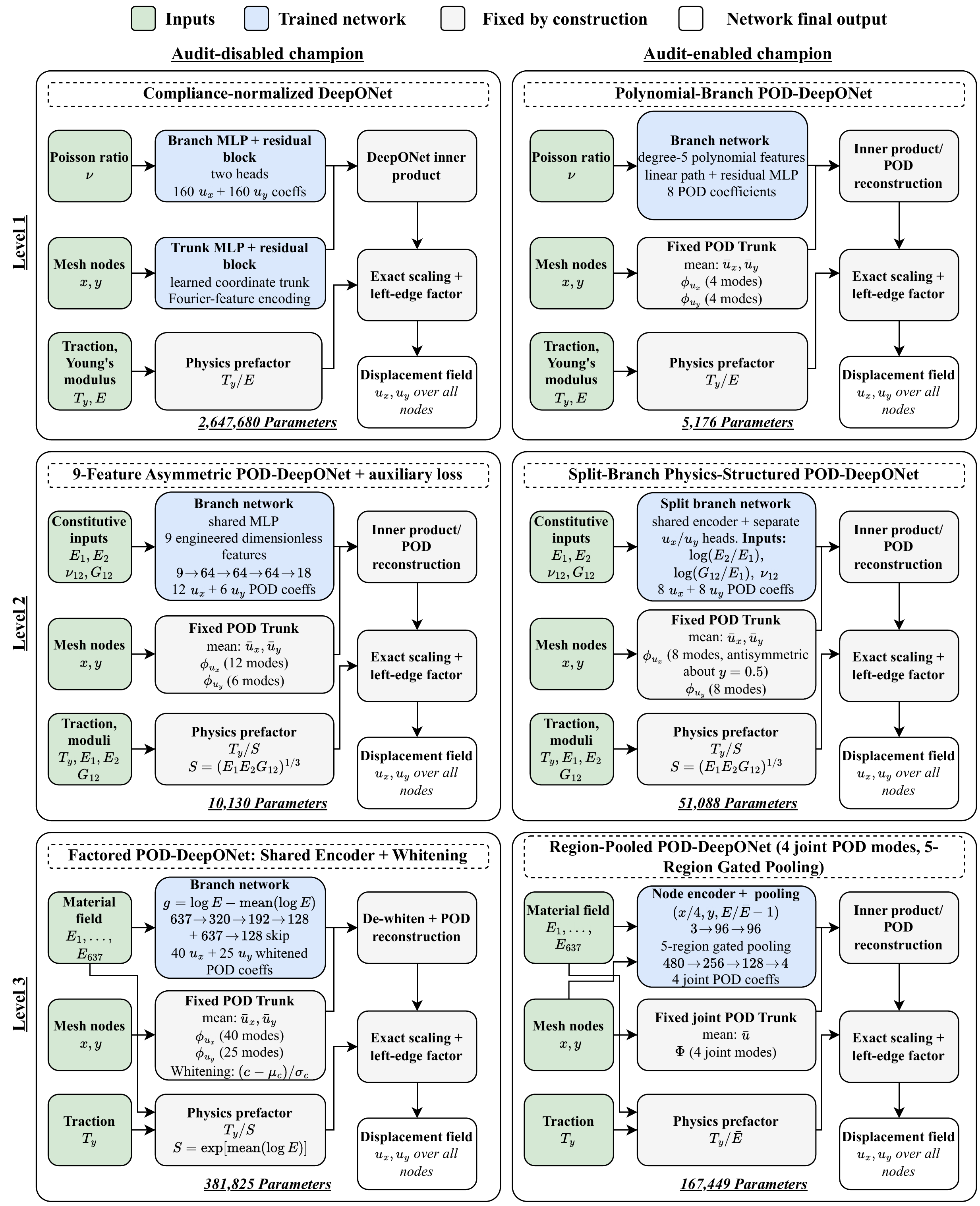}
\caption{\textbf{Selected DeepONet architectures for the primary comparison.} Each row compares the knowledge-informed audit-disabled champion (left; Level~1 has no prior knowledge to reuse) with the audit-enabled champion (right). The progression moves from a learned spatial trunk in Level~1, to compact POD coefficient maps in Level~2, and finally to field encoders in Level~3. In Level~2, the knowledge-informed audit-disabled branch has dimensions $9\!\to\!64\!\to\!64\!\to\!64\!\to\!18$. In Level~3, the knowledge-informed audit-disabled encoder has dimensions $637\!\to\!320\!\to\!192\!\to\!128$, with a $637\!\to\!128$ skip projection. The audit-enabled node encoder has dimensions $3\!\to\!96\!\to\!96$; its pooled representation has \num{480} components and feeds a $480\!\to\!256\!\to\!128\!\to\!4$ head.}
\label{fig:cantilever-champion-architectures}
\end{figure}

\Cref{fig:cantilever-champion-architectures} compares the audit-disabled and audit-enabled champions of \cref{sec:example-cantilever-progression} at each level, and \cref{lst:cantilever-champions} gives the forward structure of the three audit-enabled champions.

\begin{lstlisting}[style=surrogatepy,
caption={\textbf{Schematic forward structures of the three audit-enabled cantilever champions.} Normalization statistics, coefficient de-standardization, and device handling are omitted. The fixed POD arrays act as data-derived trunks; the trainable branch predicts coefficients, and the external prefactors and masks impose the named linear-elastic structure.},
label={lst:cantilever-champions}]
# Level 1: homogeneous isotropic
coeff_ux, coeff_uy = branch(polynomial_features(nu))
w_ux = mean_ux + coeff_ux @ phi_ux
w_uy = mean_uy + coeff_uy @ phi_uy
u_pred = (Ty / E) * left_edge_mask * stack(w_ux, w_uy)

# Level 2: homogeneous orthotropic
S = (E1 * E2 * G12) ** (1.0 / 3.0)
z = stack(log(E2 / E1), log(G12 / E1), nu12)
coeff_ux, coeff_uy = split_branch(z)
w_ux = mean_ux + coeff_ux @ antisymmetric_phi_ux
w_uy = mean_uy + coeff_uy @ phi_uy
u_pred = (Ty / S) * left_edge_mask * stack(w_ux, w_uy)

# Level 3: spatial E field
Ebar = mean(E, dim=nodes)
e_norm = E / Ebar - 1.0
node_features = node_mlp(stack(x / 4.0, y, e_norm))
regional = [gated_mean(node_features, region) for region in regions]
coeff = head(concat(regional))
u_norm = pod_mean + coeff @ pod_basis.T
u_pred = (Ty / Ebar) * left_edge_mask * reshape(u_norm, [637, 2])
\end{lstlisting}

\clearpage
\subsection{Plate with a hole}
\label{app:arch-plate}

\begin{figure}[H]
\centering
\includegraphics[width=1\linewidth]{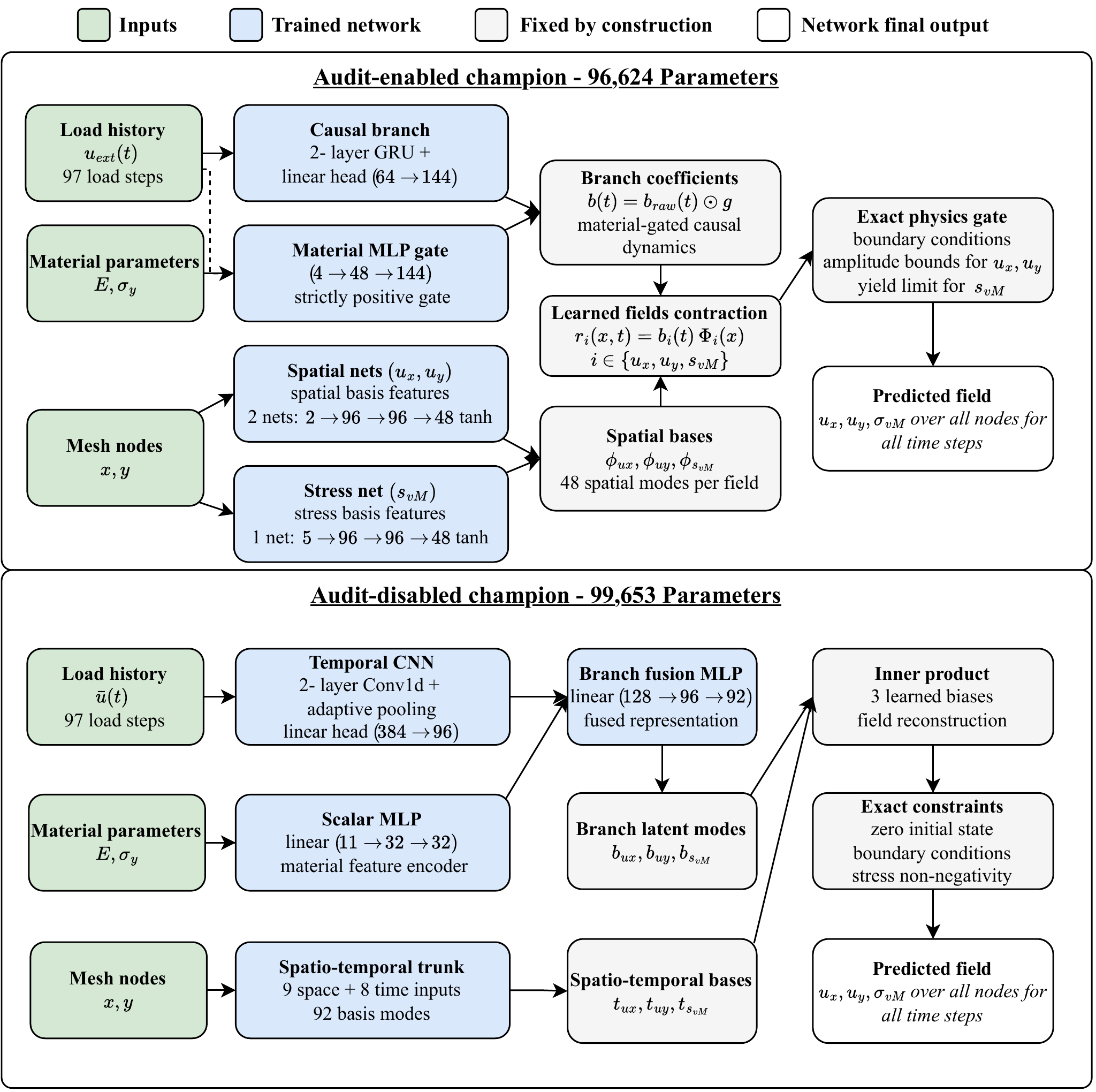}
\caption{\textbf{Plate with a hole: Level~1 champion architectures.} Audit-enabled (top) and audit-disabled (bottom).}
\label{fig:Figure_architectures_l1}
\end{figure}

\begin{figure}[H]
\centering
\includegraphics[width=1\linewidth]{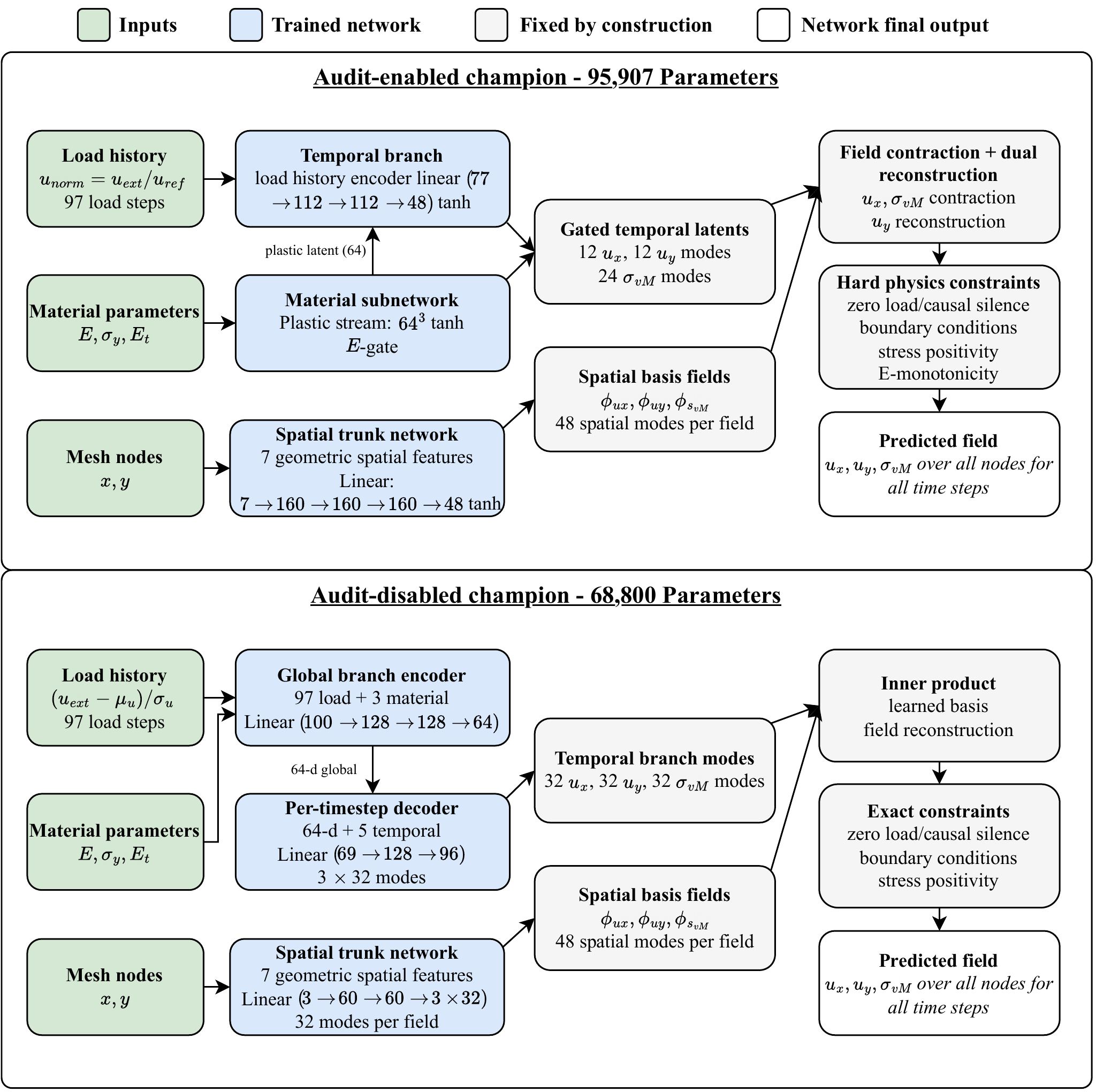}
\caption{\textbf{Plate with a hole: Level~2 champion architectures.} Audit-enabled (top) and audit-disabled (bottom).}
\label{fig:Figure_architectures_l2}
\end{figure}

\begin{figure}[H]
\centering
\includegraphics[width=1\linewidth]{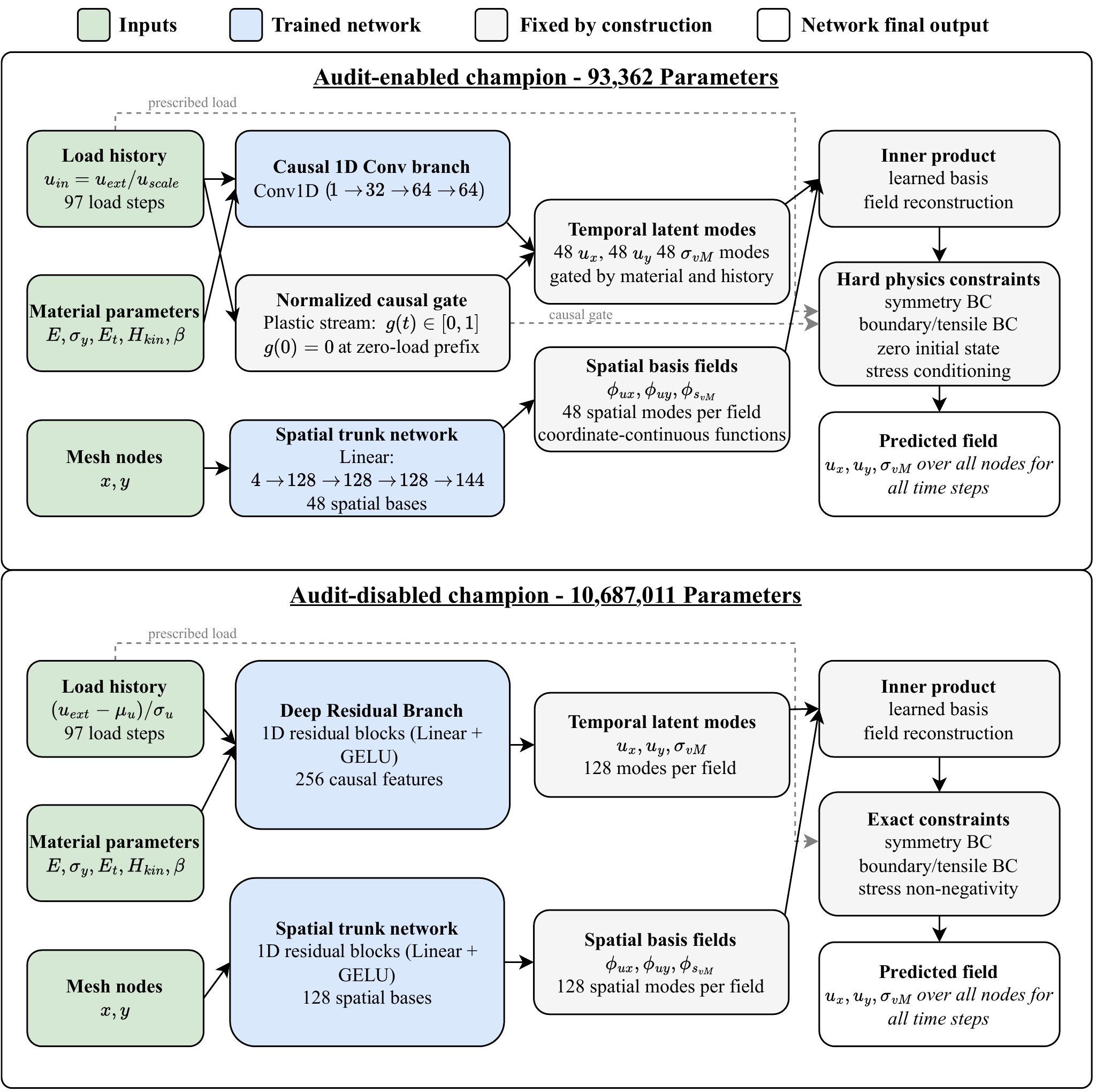}
\caption{\textbf{Plate with a hole: Level~3 champion architectures.} Audit-enabled (top) and audit-disabled (bottom). The limitation on the maximum allowable number of parameters is removed for the audit-disabled search.}
\label{fig:Figure_architectures_l3}
\end{figure}

\Cref{fig:Figure_architectures_l1,fig:Figure_architectures_l2,fig:Figure_architectures_l3} show the audit-enabled and audit-disabled champions of \cref{sec:example-plate-with-a-hole} for Levels~1--3.

\end{document}